\documentclass[10pt, twocolumn, letterpaper]{article}

\usepackage{subfigure}
\usepackage[pagenumbers]{cvpr}      

\usepackage{graphicx}
\usepackage{amsmath}
\usepackage{amssymb}
\usepackage{booktabs}

\usepackage[utf8]{inputenc} 
\usepackage[T1]{fontenc}    
\usepackage{amsfonts}       
\usepackage{nicefrac}       
\usepackage{microtype}      
\usepackage{xcolor}         
\usepackage{kotex} 
\usepackage{adjustbox}
\usepackage{comment}
\usepackage[ruled]{algorithm2e}
\usepackage{mathtools}
\usepackage{multirow}
\usepackage{bm}
\usepackage{dsfont}
\usepackage{textcomp}

\usepackage[pagebackref,breaklinks,colorlinks]{hyperref}

\usepackage[capitalize]{cleveref}
\crefname{section}{Sec.}{Secs.}
\Crefname{section}{Section}{Sections}
\Crefname{table}{Table}{Tables}
\crefname{table}{Tab.}{Tabs.}

\DeclareMathOperator*{\argmax}{argmax} 
\newcommand\naeun[1]{{\textcolor{black}{#1}}}
\newcommand\yj[1]{{\textcolor{black}{#1}}}

\def\cvprPaperID{8404} 
\def\confName{CVPR}
\def\confYear{2022}

\usepackage{amssymb}
\usepackage{pifont}

\newcommand{\cmark}{\ding{51}}%
\newcommand{\xmark}{\ding{55}}%

\begin{document}
\title{NCIS: Neural Contextual Iterative Smoothing for \\ Purifying Adversarial Perturbations}

\author{Sungmin Cha\textsuperscript{\rm 1}, Naeun Ko\textsuperscript{\rm 2}, Youngjoon Yoo\textsuperscript{\rm 2, \rm 3}, and Taesup Moon\textsuperscript{\rm 1}\thanks{Corresponding author (E-mail: \texttt{tsmoon@snu.ac.kr})} \\\

\textsuperscript{\rm 1}Department of Electrical and Computer Engineering, Seoul National University\\
\textsuperscript{\rm 2}Face, NAVER Clova 
\textsuperscript{\rm 3}NAVER AI Lab\\
{\tt\small sungmin.cha@snu.ac.kr, \tt\small naeun.ko@navercorp.com, \tt\small youngjoon.yoo@navercorp.com, \tt\small tsmoon@snu.ac.kr}
}
\maketitle

\begin{abstract}

We propose a novel and effective purification based adversarial defense method against pre-processor blind white- and black-box attacks. Our method is computationally efficient and trained only with self-supervised learning on  general images, without requiring any adversarial training or retraining of the classification model. 
We first show an empirical analysis on the adversarial noise, defined to be the residual between an original image and its adversarial example, has almost zero mean, symmetric distribution. Based on this observation, we propose a very simple iterative Gaussian Smoothing (GS) which can effectively smooth out adversarial noise and achieve substantially high robust accuracy. To further improve it, we propose Neural Contextual Iterative Smoothing (NCIS), which trains a blind-spot network (BSN) in a self-supervised manner to reconstruct the discriminative features of the original image that is also smoothed out by GS.
From our extensive experiments on the large-scale ImageNet using four classification models, we show that our method achieves both competitive standard accuracy and state-of-the-art robust accuracy against most strong purifier-blind white- and black-box attacks. Also, we propose a new benchmark for evaluating a purification method based on commercial image classification APIs, such as AWS, Azure, Clarifai and Google. We generate adversarial examples by ensemble transfer-based black-box attack, which can induce complete misclassification of APIs, and demonstrate that our method can be used to increase adversarial robustness of APIs.



\end{abstract}

\begin{figure*}[t]
\centering 
{\includegraphics[width=0.9\linewidth]{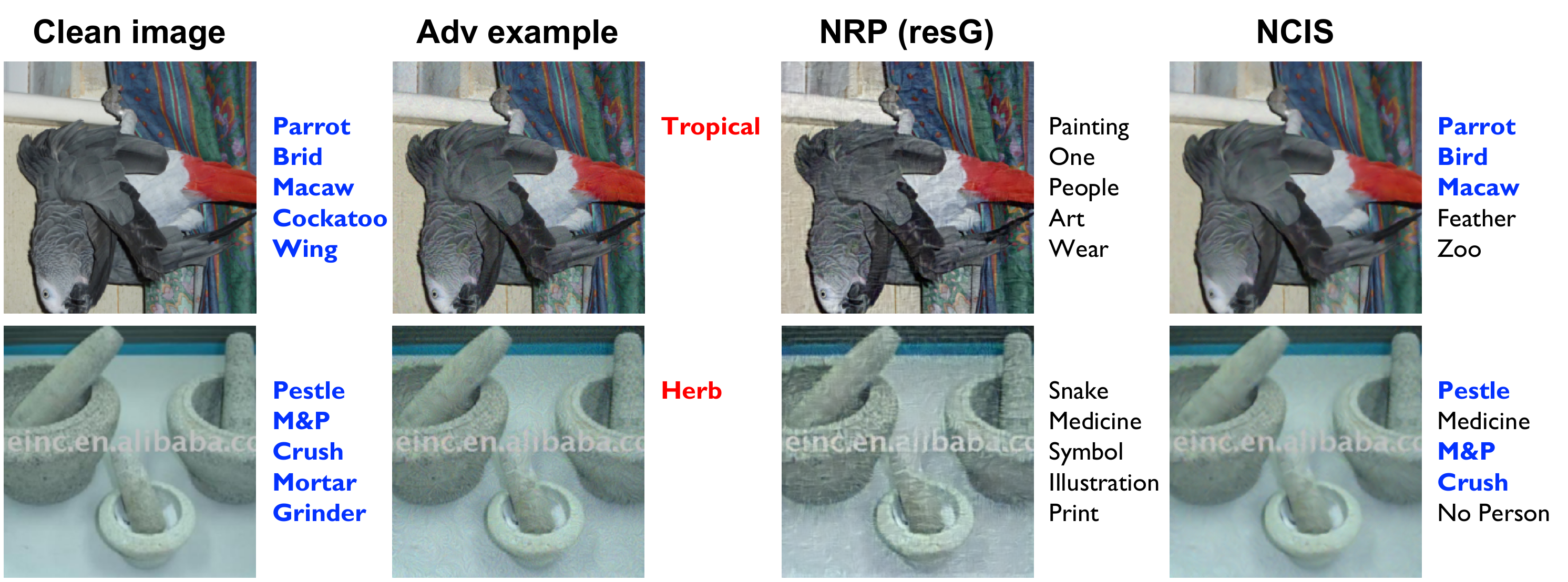}}
\vspace{-2mm}
\caption{Visualization examples for defending the commercial vision API (Clarifai). The API predicts the correct top-5 predictions for the original clean images (first column), while it gets completely fooled by the adversarial examples (second column). The right two columns show the prediction results when two purifiers, NRP (resG) \cite{(NRP)naseer2020self} and our NCIS, are applied to the adversarial examples, and we clearly observe the superior performance of NCIS. 
For more thorough and quantitative results, see Table \ref{table:api_table}.  }\vspace{-.2in}\label{figure:api_example}
 \end{figure*}

\section{Introduction}

Despite the great success of deep learning-based image classification models, it is well known that they are vulnerable to the \textit{adversarial attacks} \cite{(FGSM)goodfellow2014explaining}. Among various attempts for defending such attacks, the adversarial training (AT) \cite{(PGD)madry2017towards} is regarded as one of the most robust defense methods. 
However, AT also possesses several limitations~\cite{(advtr_survey)bai2021recent}; (a) the computational cost is very high \cite{(advtrfree)shafahi2019adversarial} particularly for the large-scale datasets like ImageNet \cite{(imagenet)deng2009imagenet}, (b) the generalization capability is weak as the model needs to be re-trained for every different task and adversary~\cite{(PGD)madry2017towards}, and (c) the standard accuracy deteriorates significantly even for a small perturbation budget for the attack~\cite{(PGD)madry2017towards}.

Regarding above limitations, the input transformation methods \cite{guo2017countering, (denoiser)liao2018defense, (FS)xu2017feature}, which attempt to remove the adversarial noise in the images before feeding them to classifiers, can be thought of as alternatives for AT since they do not require the re-training of the classifier. While such methods are widely regarded as broken by the strong \cite{guo2017countering} and sophisticated adaptive attacks, e.g., \cite{athalye2018obfuscated}, they have been reconsidered recently due to their practical value, often under the term of \textit{purification} of adversarial attacks \cite{(NRP)naseer2020self,shi2021online,(adp)yoon2021adversarial}.

In this paper, we also focus on the adversarial purification setting. In other words, we assume the adversary may (or may not) have a full access to the classifier subject to attack, but has \textit{no} access to the purification model. Such setting may seem relatively weak compared to other stronger attack scenarios, in which the adversary has additional access to the purification model or its output~\cite{(obfuscated)athalye2018obfuscated, (adaptive_attack)tramer2020adaptive}. However, we argue that such purifier-blind attack is what we may encounter the most in practice. For a more concrete argument, consider the situation in which a provider of large-scale image classification API would want to make the classifier ``robust'' to the potential adversarial attacks~\cite{(denoisedsmoothing)salman2020denoised}, while maintaining the standard accuracy as high as possible. In such case, allowing the strong adaptive attack would be unrealistic, since it basically means the entire API service system has been breached by the adversary to access the classifier and purifier, in which concerning the adversarial robustness of the classifier becomes a secondary issue. A more realistic scenario would be the one where the adversary aims to attack the classifier for the API (with or without the knowledge on the classifier)~\cite{(ensemble_attack)liu2016delving}, and the API provider can devise additional guard, i.e., the purifier, for defending the pre-trained classifier.

To that end, we devise a novel smoothing-based purification scheme that can significantly enhance the robustness of the classifier while maintaining the standard accuracy in the above mentioned setting. Our contributions are threefold. First, we make a novel observation on the distribution of the adversarial noise that it is more or less zero mean and symmetric at the patch level. Second, from above observation, we show that a very simple Gaussian Smoothing (GS), which essentially employs a non-negative, symmetric convolution kernel, can achieve surprisingly high robust accuracy when it is iteratively applied as a purifier. Third, in order to compensate the loss of the standard accuracy of the iterative GS, we employ a novel and efficient Blind-Spot Network (BSN), extended from a recent self-supervised learning based denoiser  \cite{(fbi)byun2021fbi}, to reconstruct the discriminative features in the original image that are smoothed out by the iterative GS. 

In order to validate our method, which is dubbed as Neural Contextual Iterative Smoothing (NCIS), we carry out extensive evaluation of our method on the large-scale ImageNet \cite{(imagenet)deng2009imagenet} dataset for the various  purifier-blind attack settings. More concretely, we follow the standard experimental protocol proposed in \cite{(benchmarking)dong2020benchmarking} and compare NCIS with other strong recent baselines for the following variations: four different classifier models, white-box/black-box attacks, $L_2$/$L_{\infty}$ targeted and untargeted attacks, and varying perturbation budget $\epsilon$ and attack iterations. Furthermore, to the best of our knowledge, we propose a new evaluation benchmark for the four commercial vision APIs, \textit{i.e.}, Amazon AWS, Microsoft Azure, Google, and Clarifai, for the first time and evaluate the purifier performance for the strong transfer-based black-box attacks \cite{(ensemble_attack)liu2016delving} (See Figure \ref{figure:api_example}). In results, we convincingly demonstrate that our NCIS achieves very robust performance across all the tested attack settings and significantly outperforms the state-of-the-art purifier \cite{(NRP)naseer2020self}, with $14\% $ fast inference time and $\times 14.7$ lower GPU memory requirement. Moreover, for white-box $L_{\infty}$ PGD attack, we show our NCIS even outperforms the strong AT-based defense \cite{(featuredenoising)xie2019feature}, without any re-training of the classifier.

\section{Related Work}
\vspace{-0.05in}
\paragraph{White-box adversarial attack}

White-box attacks generate adversarial examples based on the input gradient, having \yj{network} information of target models. FGSM~\cite{(FGSM)goodfellow2014explaining} generates adversarial examples with an optimization-based method by a single-step update. In follow-up \yj{studies}, multi-step methods were proposed which strengthen the level of attacks by taking multiple gradient steps~\cite{(PGD)madry2017towards, (BIM)kurakin2016adversarial} \yj{in iterative manner}. 
\yj{Also, other attack approaches including} new loss functions~\cite{(C&W)carlini2017towards, (DeepFool)moosavi2016deepfool}, momentum-based iterative attack~\cite{(MIFGSM)dong2018boosting}, and creating diverse input patterns~\cite{(DIFGSM)xie2019improving} \yj{have been} introduced.

\vspace{-.2in}
\paragraph{Black-box adversarial attack}
\yj{Black-box attack deals with the case} an attacker cannot access the model gradient, and hence much difficult than white-box attack scenario. Transfer-based attacks~\cite{(ensemble_attack)liu2016delving} generate adversarial examples against a substitute model, and it is known that attacks are more successful when the architecture of the substitute model and the target model is similar~\cite{(benchmarking)dong2020benchmarking}. 
Query-based attacks~\cite{(zoo)chen2017zoo, (SPSA)uesato2018adversarial, (NES)ilyas2018black, (square)andriushchenko2020square, (square)andriushchenko2020square} estimate the gradient through queries, but require much amounts for adversarial examples generation.

\vspace{-.2in}
\paragraph{Adversarial defense} 
Adversarial training (AT) \yj{methods}~\cite{(FGSM)goodfellow2014explaining, (PGD)madry2017towards} train the classifier with adversarial examples by min-max optimization, \yj{and shows} stable robustness against \yj{various} adversarial attacks. 
However, \yj{they require excessive training cost and have trade-off~\cite{(benchmarking)dong2020benchmarking, (featuredenoising)xie2019feature} between performance and attack-robustness.} 
Certified defense methods \yj{aim to} guarantee the model to be robust against adversarial perturbations \yj{in theoretical bound}. \cite{(certified)cohen2019certified} proved a tight robustness guatantee in $L_2$ norm for randomized smoothing with Gaussian noise. 
\cite{(denoisedsmoothing)salman2020denoised} proposed to apply a pre-trained Gaussian denoiser \yj{to any classifier, and guaranteed $L_p$-robust to adversarial examples.}

Input transformation processes the input images to achieve robustness against adversarial attacks~\cite{(JPEG)dziugaite2016study, (FS)xu2017feature, guo2017countering, (denoiser)liao2018defense}. These methods are simple and cheap to apply but are broken easily by strong adversarial attacks~\cite{(benchmarking)dong2020benchmarking}.
Recently, purification methods which shift the adversarial examples back to the clean data representation have been \yj{actively} proposed~\cite{(ape-gan)shen2017ape, (defense-gan)samangouei2018defense, (adp)yoon2021adversarial, (SOAP)shi2020online, (NRP)naseer2020self}. 
However, most of \yj{the methods} conducted experiments only on small \yj{datasets} (\textit{e.g.}, MNIST and CIFAR-10/100). Among the methods, \cite{(NRP)naseer2020self} is the only one who showed experimental results on ImageNet.
Also, in \cite{(SOAP)shi2020online}, \yj{the proposed auxiliary task should be trained together with the classifier.} 
\vspace{-.2in}
\paragraph{Blind-spot networks (BSN) for image denoising} 
Recent blind image denoising, which is to train a neural network based denoiser only using noisy images, has made \yj{significant improvements}.
One of the main research directions for this is to use blind-spot network (BSN). 
Several types of BSN are proposed for their own proposed method~\cite{(N2V)krull2018noise2void,(NAIDE)cha2018neural,(HQDenoising)laine2019high,(FCAIDE)cha2019fully,(DBSN)wu2020unpaired,(fbi)byun2021fbi}.
Among them, FBI-Net~\cite{(fbi)byun2021fbi} achieves the superior denoising performance with \yj{small-sized network add-on}.

\section{Motivation}
\subsection{Preliminary and Notations}\label{subsec:prelim}
\paragraph{Adversarial attack} 
In general image classification task, we denote the original input image by
$\bm x\in \mathcal{X}\subset \mathbb{R}^{D}$, in  which $D$ is the number of pixels, and the ground-truth label of $\bm x$ by $\bm{y}\in \mathcal{Y} \subset \mathbb{R}^{L}$,
in which $L$ is the number of classes.
A classification model parameterized by $\bm\phi$ is denoted by $g_{\bm\phi}:\mathcal{X}\rightarrow \mathcal{Y}$. and we assume $g_{\bm\phi}$ is pre-trained with external training data. 
Now, an adversary attempts to attack $g_{\bm\phi}$ by generating an adversarial example $\bm x'$ for $\bm x$ \yj{cause} the mis-classification of $\bm x$. The attack is called \textit{untargeted} if $g_{\bm\phi}(\bm x')\neq\bm y$ and  \textit{targeted} if $g_{\bm\phi}(\bm x')=\bm y^{*}$ where $\bm y^{*}$ is a target class different from $\bm y$. The adversarial example $\bm x'$ is typically generated by solving the optimization, \textit{e.g.}, 
\begin{equation}\label{eq:adversarial_attack}
    \bm x' = \argmax_{\bm z:\|\bm z - \bm x\|_{p} \le \epsilon} \mathcal{L}(g_{\bm\phi}(\bm z), \bm y ; \bm \phi),
\end{equation}
for the untargeted attack, \yj{where} $\mathcal{L}$ denotes a classification loss function, $\|\bm z- \bm x\|_{p} \le \epsilon$ denotes the $L_p$-norm constraint of the optimization, \yj{and}  $\epsilon$ is the perturbation budget. Directly solving of (\ref{eq:adversarial_attack}) is intractable and various approximation methods have been proposed \cite{(FGSM)goodfellow2014explaining, (C&W)carlini2017towards, (PGD)madry2017towards, (BIM)kurakin2016adversarial, (MIFGSM)dong2018boosting, (DIFGSM)xie2019improving}. 
We refer to \textit{white-box} attack when the model architecture and weight of $g_{\bm\phi}$ are completely known, and \textit{black-box} attack \textit{vice versa}.
\vspace{-.2in}
\paragraph{Purifier} 
Purification is a pre-processing method which purifies a given input image before it is fed to a classification model. 
Let the goal of the purifier $T_{\bm \theta}:\mathcal{X}\rightarrow \mathcal{X}$ is to reconstruct the purified image which allows $g_{\bm\phi}$ to predict an original prediction of $\bm x$ for a given input. Then, given the adversarial example $\bm x'$, the optimal purifier $T_{\bm \theta^{*}}$ will be  
\begin{equation}\label{eq:purification_goal}
g_{\bm\phi}(T_{\bm \theta^{*}}(\bm x'))=\hat{\bm y},
\end{equation}
where $\hat{\bm y} = g_{\bm\phi}(\bm x)$.
However, note that the purifier should randomly take adversarial example $\bm x'$ and original image $\bm x$ as the input image in the general image classification situation for defense, \yj{making difficult to be trained in a naive supervised manner.}

\vspace{-.2in}
\paragraph{Blind-spot network (BSN)} 

The BSN is a special form of neural network that tries to reconstruct a pixel in the middle based on its surrounding context pixels. Namely, for a given input $\bm x$, the output of the BSN for pixel $i$ is denoted by
\begin{eqnarray}
f_{\bm\theta,i}(\bm x) = f(\bm \theta, \bm C_{k\times k}^{-i}),\label{eq:bsn}
\end{eqnarray}
in which $\bm C_{k\times k}^{-i}$ is the $k\times k$ patch of $\bm x$ surrounding $i$, that does \textit{not} include the pixel $i$. The BSN is typically used as a denoiser to estimate the clean $\bm x$ based on the noisy $\bm z$, of which the model parameter $\bm\theta$ is trained by minimizing  $\|\bm z- f_{\bm\theta}(\bm z)\|_2^2$, \textit{i.e.}, in a self-supervised manner. In this paper, we utilize BSN as a smoothing function that is trained with the original \textit{clean} $\bm x$ and show that it can work as an effective purifier that can both remove the adversarial noise and maintain the original discriminative feature of $\bm x$. Furthermore, for a particular BSN architecture, we extend the recent state-of-the-art network called FBI-Net in \cite{(fbi)byun2021fbi}.


\begin{figure*}[h]
\centering 
\subfigure[Distribution of mean]
{\includegraphics[width=0.230\linewidth]{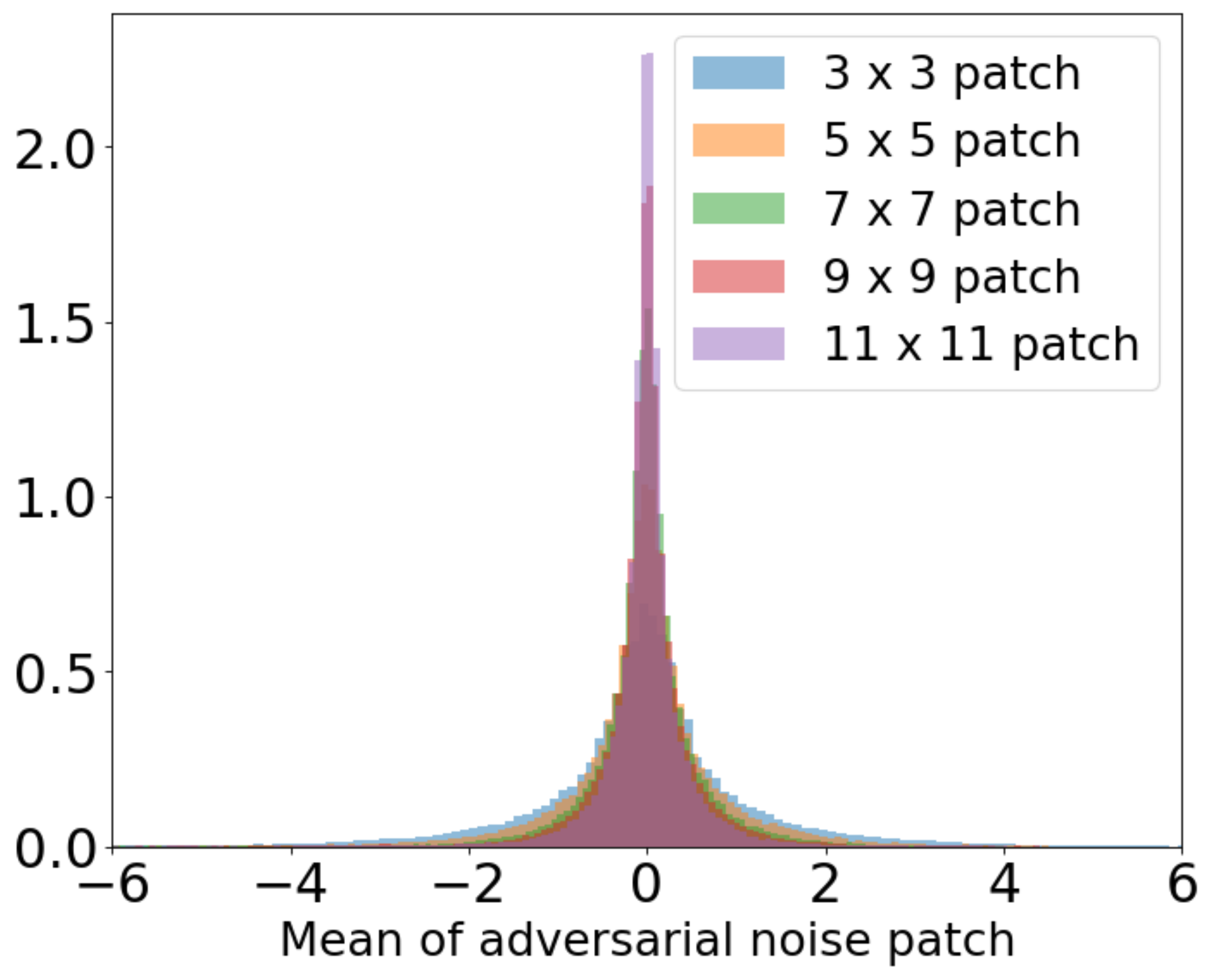}
\label{figure:motivation_distribution}}
\subfigure[Skewness]
{\includegraphics[width=0.230\linewidth]{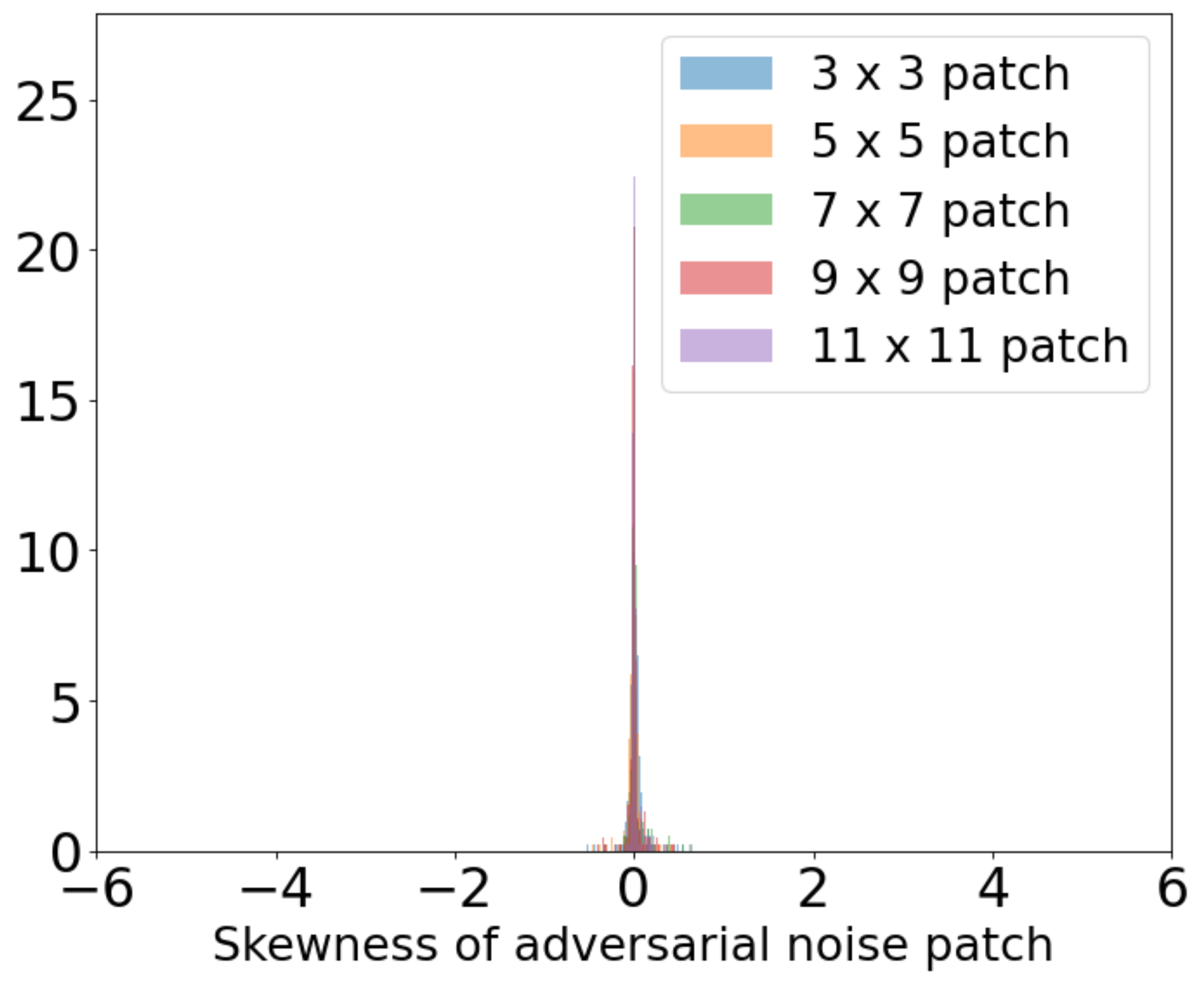}
\label{figure:motivation_skewness}}
\subfigure[MSE]
{\includegraphics[width=0.252\linewidth]{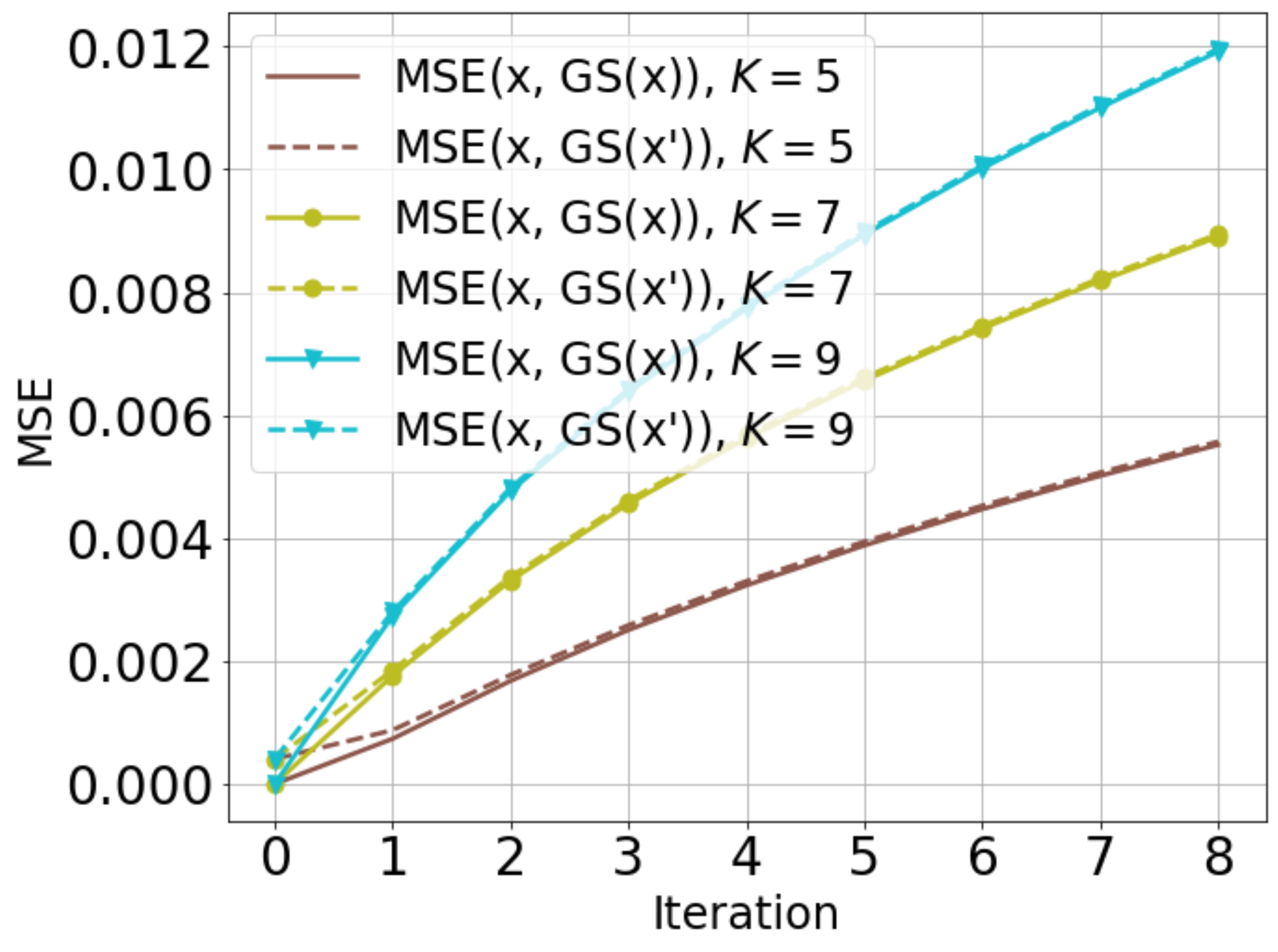}
\label{figure:motivation_mse}}
\subfigure[Accuracy]
{\includegraphics[width=0.248\linewidth]{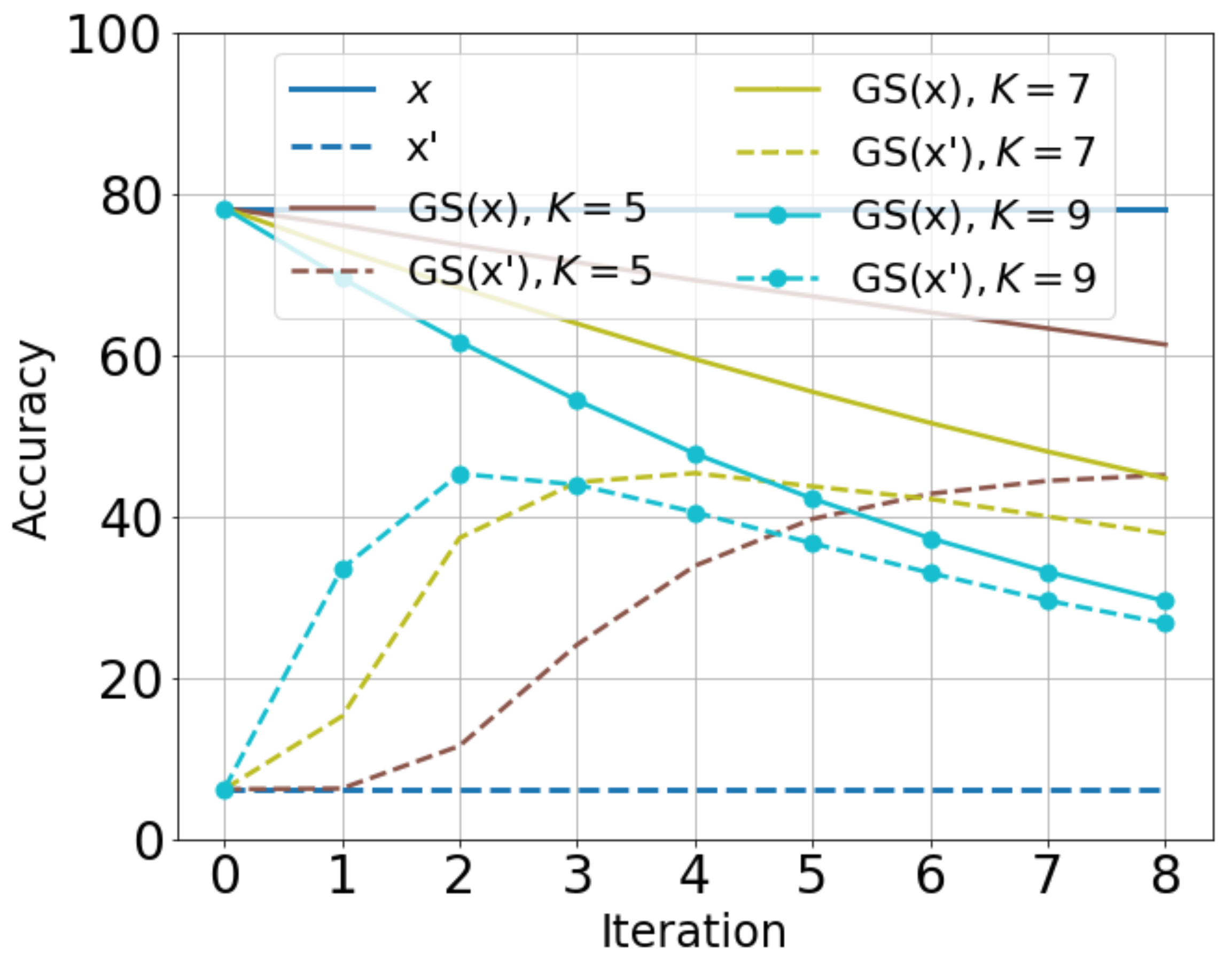}
\label{figure:motivation_acc}
}\vspace{-.15in}
\caption{Experimental analysis for adversarial examples generated from untargeted $L_{\infty}$ PGD ($\alpha$ = 1.6 / 255, where $\alpha$ is a step size) attack with 10 attack iterations, and iterative Gaussian smoothing ($\sigma = (K-1)/6$ where $K$ is the size of a kernel).}\vspace{-.25in}
 \end{figure*}
 
\subsection{Analysis on Adversarial Noise}

Suppose we have $n$ clean images $\{\bm x_i\}_{i=1}^n$ and generated adversarial examples $\{\bm x'_i\}_{i=1}^n$ for each clean image by the process (\ref{eq:adversarial_attack}). Assuming that the adversarial noise is additive, we denote the $i$-th adversarial \textit{noise} image as 
\begin{equation}
    \bm N'_{i} = \bm x'_{i} - \bm x_{i}, \ \  i = 1, \dots, n,\label{eq:adv_noise}
\end{equation}
and analyze the empirical distribution of $\bm N'_{i}$. 
More concretely, we randomly selected 1,000 images from the ImageNet training set and generated adversarial examples with \yj{the most generally used}, untargeted $L_\infty$ PGD white-box attack \cite{(PGD)madry2017towards}. 
We collected  5,000 adversarial noise images in total by generating 5 attacked images for each $\bm x_i$ using the perturbation budgets $\epsilon = \{1/255, 2/255, 4/255, 8/255, 16/255\}$, respectively. Then, we randomly cropped 100 $K\times K$ patches from each noise image (thus, obtained $5\times 10^5$ patches) and computed the empirical mean and skewness~\cite{(skewness)groeneveld1984measuring} of those patches. 

Figure \ref{figure:motivation_distribution} shows the empirical mean of the adversarial noise in the patches (for varying patch sizes), and Figure \ref{figure:motivation_skewness} shows the empirical skewness. From the figures, we clearly observe that the adversarial noise in a patch is more or less zero mean and has a symmetric distribution. This somewhat surprising regularity of adversarial noise, although generated from the complex iterative optimization process, motivates us to use a very simple Gaussian Smoothing (GS) based purifier in the next section.

\subsection{Iterative Gaussian Smoothing (GS)}
\label{sec:IGS}

Gaussian Smoothing (GS) 
\yj{is one of the  widely used low-pass filters} for image processing, and is used to smooth an image with a Gaussian kernel. 
The mechanism can be represented by the convolution operation between a Gaussian kernel $\mathbf{k}$ of size $K \times K$ 
and an input image. 
Namely, when an adversarial example $\bm x'$ is given as the input to the Gaussian smoothing function $\mathbf{G}(\cdot)$, we denote the GS process as 
\begin{equation}\label{eqn:gaussian_smoothing}
    \mathbf{G}(\bm x') = \bm x' \otimes \mathbf{k}, 
\end{equation}
in which $\otimes$ is the convolution operation.
Since $\mathbf{G}(\cdot)$ is a linear operation, and from (\ref{eq:adv_noise}), we have
\begin{eqnarray}
\begin{aligned}
\label{eqn:gaussian_smoothing_linear}
    \mathbf{G}(\bm x') &= \mathbf{G}(\bm x + \bm{N'}) = \mathbf{G}(\bm x) + \mathbf{G}(\bm{N'}) \\
                       &= \mathbf{G}(\bm x) + \bm{N'} \otimes \mathbf{k} \approx \mathbf{G}(\bm x).
\end{aligned}
\end{eqnarray}
The last approximation follows from $\bm{N'} \otimes \mathbf{k}\approx 0$, which is from the observation in Figures \ref{figure:motivation_distribution} and \ref{figure:motivation_skewness} that the pixels in $K\times K$ patches in $\bm N'$ have symmetric, zero mean distribution and $\mathbf{k}$ is a non-negative, symmetric kernel. From (\ref{eqn:gaussian_smoothing_linear}), therefore, we can deduce that GS can mostly wash out the adversarial noise in $\bm x'$ and will make the smoothed version of $\bm x'$ become almost identical to that of $\bm x$.



Furthermore, by denoting $\mathbf{G}^i = \mathbf{G}\circ \mathbf{G}^{i-1}$ as applying GS  $i$ times iteratively, we can apply the similar logic as in equation (\ref{eqn:gaussian_smoothing_linear}) and deduce $\mathbf{G}^i(\bm x')\approx \mathbf{G}^i(\bm x)$ as well, but with smaller approximation error. The reason is because the patches in $\mathbf{N}_i\otimes \mathbf{k}$ will still have zero mean, symmetric distributions, and the adversarial noise will keep getting washed out as we iteratively apply the Gaussian kernel $\mathbf{k}$. On the other hand, as GS continues, we can also expect that $\mathbf{G}^i(\bm x')$ and $\mathbf{G}^i(\bm x)$  will get farther from $\bm x$, since GS will also wash out detailed and discriminative features in $\bm x$. Thus, as the iteration continues, we can conjecture that iterative GS may encounter a trade-off of increasing the robust accuracy, by removing adversarial noise, while hurting the standard accuracy, by also removing the discriminative features in $\bm x$.



Figure \ref{figure:motivation_mse} and \ref{figure:motivation_acc} experimentally validate above conjecture for the iterative GS. 
Namely, for an ImageNet pre-trained ResNet-152 classifier, we tested with 1,000 clean images, $\bm x$, randomly subsampled from the ImageNet training dataset, and their adversarial examples, $\bm x'$, generated by $L_\infty$ PGD $(\epsilon = 16/255)$ attack. 
Figure \ref{figure:motivation_mse} shows the MSE of $\|\bm x-\mathbf{G}^i(\bm x)\|_2^2$ and $\|\bm x-\mathbf{G}^i(\bm x')\|_2^2$ as $i$ increases, for different patch size $K$. From the figure, we can clearly observe that the two MSEs become almost identical, but increase (\textit{i.e.}, both $\mathbf{G}^i(\bm x)$ and $\mathbf{G}^i(\bm x')$ get farther from $\bm x$), as $i$ increases, corroborating our above conjecture. 
Figure \ref{figure:motivation_acc} reports both standard and robust accuracy of ResNet-152 when the iterative GS is used as a purifier, for the same $L_\infty$ PGD attack on the whole ImageNet validation set. 
We observe that the very simple iterative GS is surprisingly effective in purifying the adversarial noise as the robust accuracy of $44.92\%$ is achieved when $K=5$ and $i=7$. However, we also note that the standard accuracy decreases as $i$ gets larger due to the removed discriminative features. Motivated by this result, we propose our NCIS, which utilizes the iterative GS to maintain high robust accuracy, but also employs BSN that can reconstruct the discriminative features washed out by GS to also achieve high standard accuracy.

\section{Neural Contextual Iterative Smoothing}


Motivated by the strong performance of iterative GS in the previous section, we propose to use a learnable neural network-based smoothing function as an iterative smoothing based purifier. More concretely, we first present the improved performance of using self-supervised trained \textbf{FBI-Net} \cite{(fbi)byun2021fbi} for iterative smoothing, devise a more efficient version of FBI-Net dubbed as \textbf{FBI-E}, and present our \textbf{NCIS}, combining FBI-E with GS for a more stable and superior purifier. \vspace{.1in}

\begin{figure}[h]
\centering 
\vspace{-.2in}
\subfigure[Reconstruction process of FBI]
{\includegraphics[width=0.47\linewidth]{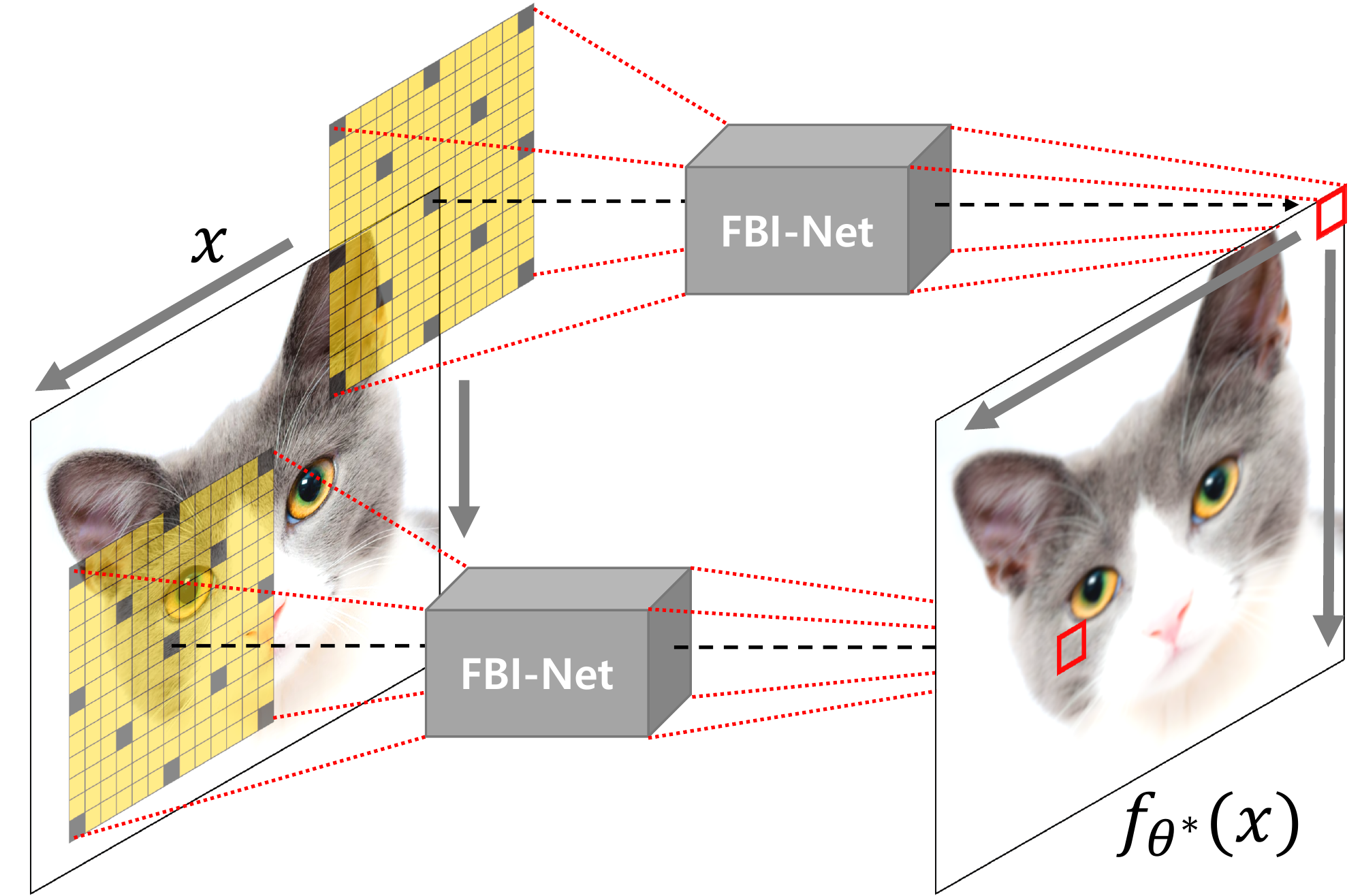}
\label{figure:illustration}}
\subfigure[Example of FBI-E for $m=2$.]
{\includegraphics[width=0.48\linewidth]{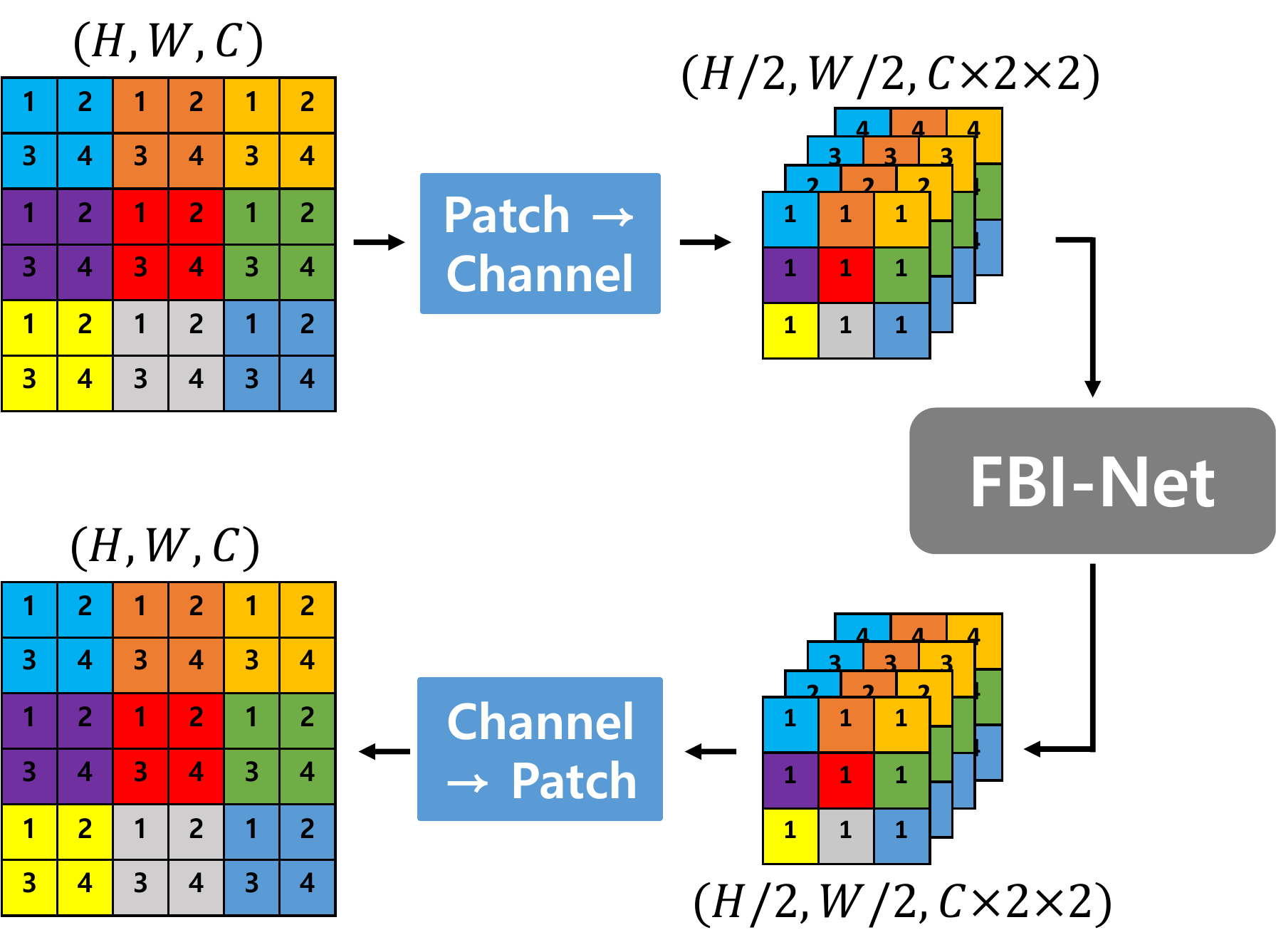}
\label{figure:operation}}
\vspace{-.11in}
\caption{Illustration of FBI and FBI-E. (a): The black points are a masked pixel in a receptive field of FBI-Net.}
\vspace{-.1in}
 \end{figure}
\noindent\textbf{Iterative smoothing with FBI-Net} FBI-Net is a fully convolutional network that utilizes a special class of masked convolution filters as shown in Figure \ref{figure:illustration} such that the BSN condition (\ref{eq:bsn}) can be satisfied. Now, for given $n$ clean images $\{\bm x_i\}_{i=1}^n$, we can train the network parameters, $\bm\theta$, of the FBI-Net by minimizing 
\begin{eqnarray}
\sum_{i=1}^n\|\bm x_i- f_{\bm\theta}(\bm x_i)\|_2^2,\label{eq:ssl_fbi}
\end{eqnarray}
namely, in a self-supervised manner. Denoting the learned parameter by $\bm\theta^*$, we expect that $f_{\bm\theta^*}(\bm x)$ can be an effective smoothing-based purifier. The reasoning is, since the adversarial noise in $\bm x'$ is very small as shown in Figure \ref{figure:motivation_distribution}, and it is generated independent of $f_{\bm \theta^*}$, the trained FBI-Net would have the similar outputs for both $\bm x$ and $\bm x'$, and they will be a smoothed reconstruction of $\bm x$. Namely, we expect 
\begin{eqnarray}
\begin{aligned}
\label{eqn:fbi_linear}
    \| f_{\bm \theta^{*}}(\bm x') - \bm x \|_2^2 \approx \| f_{\bm \theta^{*}}(\bm x) - \bm x \|_2^2 \\
    < \|\mathbf{G}(\bm x') - \bm x \|_2^2 \approx \|\mathbf{G}(\bm x) - \bm x \|_2^2 \label{eq:approx}
\end{aligned}
\end{eqnarray}
Note the reasoning for (\ref{eq:approx}) is possible since we are using BSN; when an ordinary fully convolutional network based denoiser, \textit{e.g.}, DnCNN \cite{(dncnn)zhang2017beyond}, is used for $f_{\bm\theta}$ in (\ref{eq:ssl_fbi}), then it will end up learning an identity map, hence,  
$\| f_{\bm \theta^{*}}(\bm x') - \bm x' \|_2^2 \approx 0$. Thus, the adversarial example will be preserved. 

\begin{figure}[h]
\centering 
 \vspace{-.2in}
\subfigure[MSE for original $\bm x$]
{\includegraphics[width=0.49\linewidth]{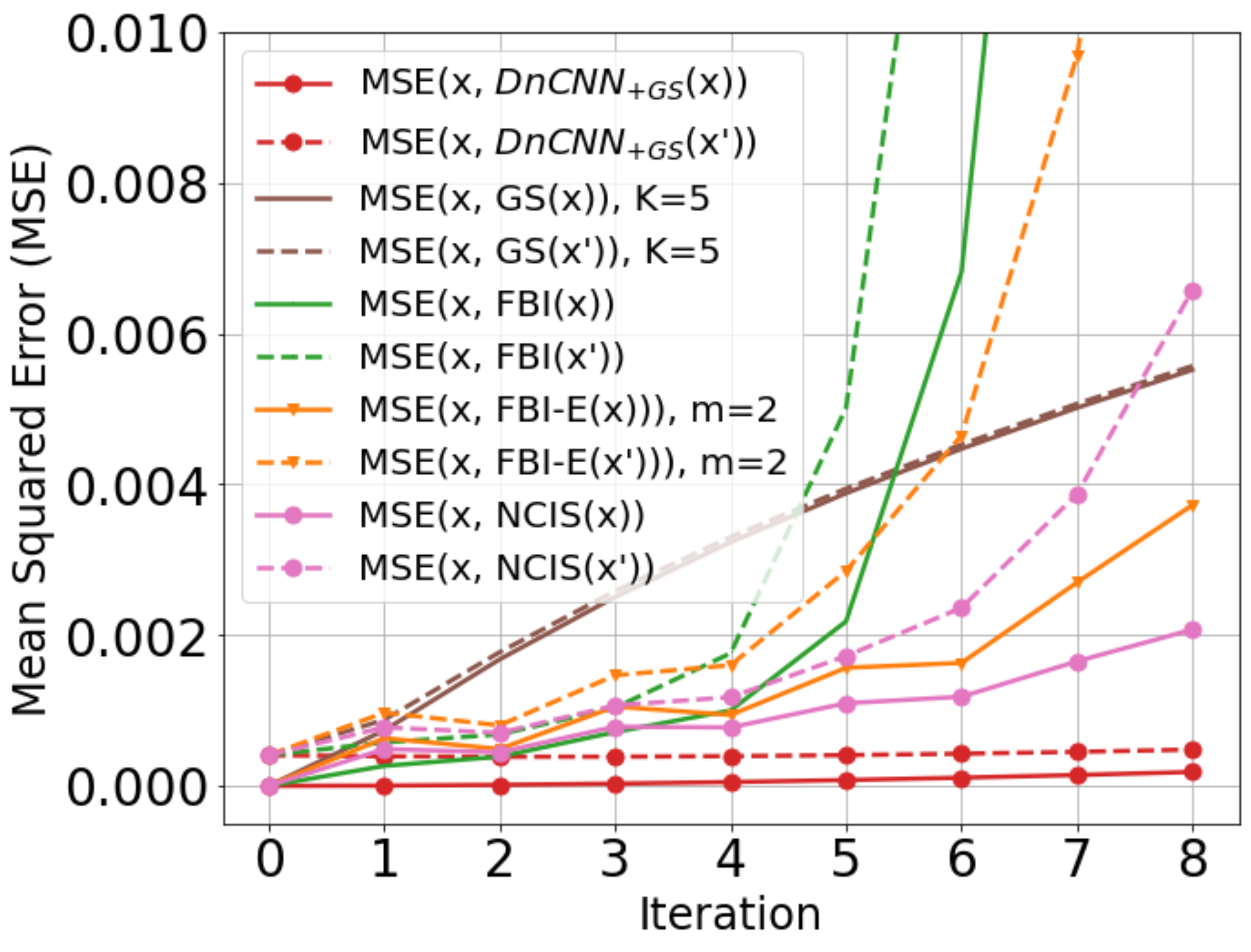}
\label{figure:verification_mse}}
\subfigure[Purification Success Rate (PSR)]
{\includegraphics[width=0.47\linewidth]{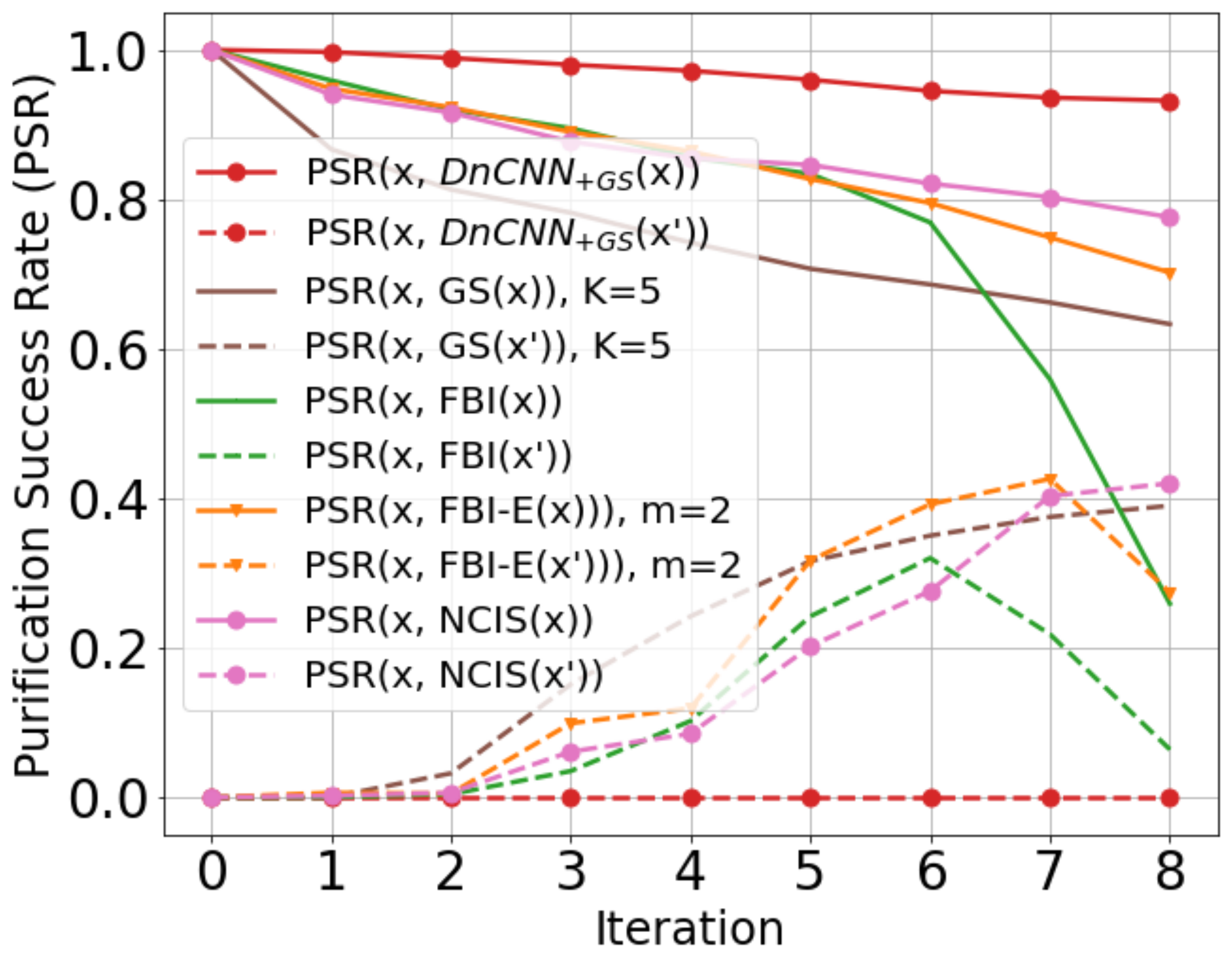}
\label{figure:verification_psr}}
\vspace{-.15in}
\caption{Verification using randomly selected 1,000 images from ImageNet training dataset and adversarial examples of generated from untargeted $L_{\infty}$ PGD ($\epsilon = 16/255$, $\alpha = 1.6/255$) attack with 10 attack iterations}
\vspace{-.2in}
\label{figure:verification}
 \end{figure}

Figure \ref{figure:verification_mse}, which is for the same setting as Figure \ref{figure:motivation_mse},  experimentally verifies above intuition for the iterative smoothing with the FBI-Net. Namely, denoting $f^i_{\bm\theta^*}(\bm x)=f_{\bm\theta^*}\circ f_{\bm\theta^*}^{i-1}(\bm x)$, the figure shows $\|\bm x-f^i_{\bm\theta^*}(\bm x)\|_2^2$ and $\|\bm x-f^i_{\bm\theta^*}(\bm x')\|_2^2$ as the iteration number $i$ increases (the green solid and dashed line, respectively). Compared to the best iterative GS with $K=5$ (brown solid and dashed lines), we observe that FBI-Net based smoothing does a much better job than  GS in reconstructing the original image $\bm x$ for both input cases ($\bm x$ and $\bm x'$), until $i\leq 4$. Moreover, in Figure \ref{figure:verification_psr}, we show the Purification Success Rate (PSR), which is defined as
$$
\text{PSR}(\bm x, \bm z) = \frac{1}{n}\sum_{i=1}^n\mathds{1}\{g_{\phi}(\bm x)=g_{\phi}(\bm z)\}
$$
in which $\bm z$ is the output of a purifier that takes either $\bm x$ or $\bm x'$ as input. The PSR is a metric for how much the PSR can recover the original classification result for $\bm x$. From the figure, we again observe that the iterative smoothing with FBI-Net does a much better job than iterative GS, in line with Figure \ref{figure:verification_mse}, in preserving the original classification for $\bm x$ (green solid), until iteration $i\leq 6$. We also observe that PSR for $\bm x'$ (green dashed) increases as well until $i\leq 6$.

\vspace{-.1in}
\paragraph{FBI-E: Extending FBI-Net}
While promising, we also observe that iterative smoothing with FBI-Net shows a few limitations. 
First, PSR for the given $\bm x'$ is still low\yj{er} than iterative GS. 
Second, MSE and PSR diverge significantly, causing unstable results after a certain iteration.
Finally, FBI-Net requires large GPU memory cost for the reconstruction and has a slow inference time to be used as a purifier.

To overcome these limitations and improve efficiency, we introduce two tensor operations for FBI-Net, \textit{patch$\rightarrow$channel} and \textit{channel$\rightarrow$patch}, to expand the reconstruction process from \textit{context$\rightarrow$pixel} to \textit{context$\rightarrow$patch}. Hence, the FBI-E mapping, $F_{\bm \theta}(\bm x)$, can be denoted as
\begin{equation}
    F_{\bm \theta}(\bm x) = \mathcal{O}_{C\rightarrow P}(f_{\bm \theta}(\mathcal{O}_{P\rightarrow C}(\bm x))),
\end{equation}\label{eqn:extension}
in which $\mathcal{O}_{C\rightarrow P}(\cdot)$ and $\mathcal{O}_{P\rightarrow C}(\cdot)$ denotes \textit{patch$\rightarrow$channel} and \textit{channel$\rightarrow$patch} operation, respectively.
 Figure \ref{figure:operation} illustrates the two operations being applied as a pre- and post-operation for FBI-Net;
$\mathcal{O}_{C\rightarrow P}(\cdot)$ transfers pixels of each $m \times m$ patch in an input image (color-coded) to channel-wise pixels and $\mathcal{O}_{P\rightarrow C}(\cdot)$ exactly does the reverse operation.
By these operations, FBI-E is also a BSN, but the reconstruction is now done in a patch level based on the context around the patch, and we expect (\ref{eq:approx}) would also hold for $F_{\bm\theta}$.


\yj{The advantage of the two proposed operations} is quite huge since it reduces the spatial resolution of both the input image and all feature maps of FBI-Net by $m \times m$ times.
As shown in Figure \ref{figure:verification_mse} and \ref{figure:verification_psr}, we experimentally checked that FBI-E with $m = 2$ achieves superior MSE and PSR for both input cases (orange solid and dashed lines) than GS and FBI-Net. Furthermore, the inference time and memory improvement of FBI-E over FBI-Net is given in the ablation study given in the later section. 
\begin{figure*}[t]
\centering 
 \vspace{-.1in}
{\includegraphics[width=0.98\linewidth]{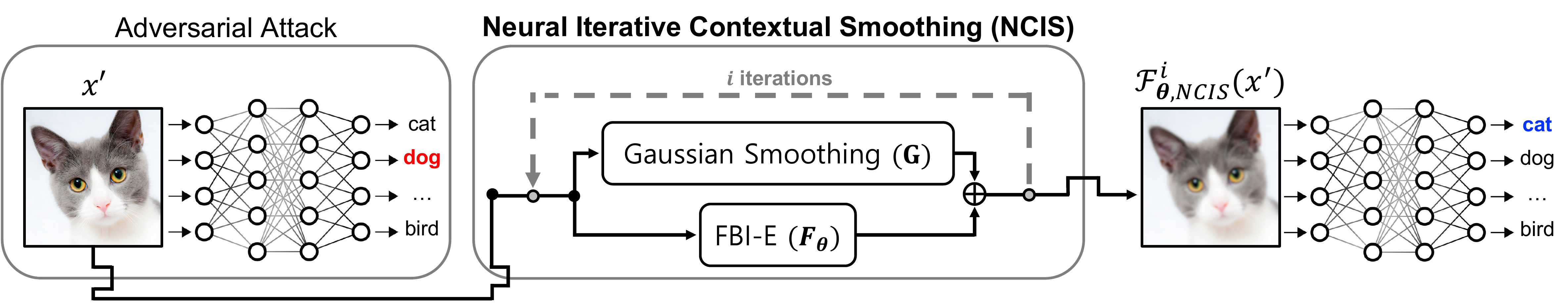}}
 \vspace{-.1in}
\caption{Overall procedure of NCIS for adversarial purification.}\vspace{-.25in}\label{figure:ncis_overall}
 \end{figure*}
\vspace{-.2in}
\paragraph{Neural contextual iterative smoothing (NCIS)}
After applying the proposed two operations, the level of difficulty for the reconstruction increases since FBI-E has to reconstruct the entire patch of size $m \times m$, not just a single pixel.
\yj{To address this problem, we propose to combine FBI-Net with GS and denote the new smoothing function as}
\begin{equation}
    \mathcal{F}_{\bm \theta, \text{NCIS}}(\bm x) = F_{\bm \theta}(\bm x) +\mathbf{G}(\bm x).
\end{equation}\label{eqn:SSICS}
The network parameter $\bm\theta^*$ is obtained by minimizing (\ref{eq:ssl_fbi}), which results in FBI-E, $F_{\bm \theta}(\bm x)$, learning to reconstruct the residual $\bm x - \mathbf{G}(\bm x)$. The intuition is, when an adversarial example $\bm x'$ is given as input, we expect $\mathbf{G}(\bm x')$ to wash out the adversarial noise such that $\mathbf{G}(\bm x')\approx \mathbf{G}(\bm x)$, and then we let $F_{\bm \theta^*}(\bm x')$ to reconstruct the discriminative features of the original $\bm x$ that is also smoothed out by $\mathbf{G}(\bm x)$. 
Hence, 
we expect $F_{\bm \theta^*}(\bm x')\approx F_{\bm \theta^*, \text{NCIS}}(\bm x)\approx \bm x-\mathbf{G}(\bm x)$.

 
Finally, our proposed Neural Contextual Iterative Smoothing (NCIS) is obtained by iteratively applying  $\mathcal{F}_{\bm \theta, \text{NCIS}}(\bm x)$ denoted as $\mathcal{F}^i_{\bm \theta, \text{NCIS}}(\bm x)=\mathcal{F}_{\bm \theta, \text{NCIS}}\circ \mathcal{F}^{i-1}_{\bm \theta, \text{NCIS}}(\bm x)$.
The overall procedure of our NCIS is depicted in Figure \ref{figure:ncis_overall}.

In Figure \ref{figure:verification_mse} and \ref{figure:verification_psr}, we verify the promising results of our NCIS. First, NCIS ($m=2$ for FBI-E and $K=11$ for GS) achieves the lowest MSE for both input cases until the $i=7$
compared to other methods, which shows that NCIS can successfully estimate the original image $\bm x$ for both inputs ($\bm x$ and $\bm x'$). Second, consequently, NCIS with seven iterations attains the highest PSR for $\bm x$ maintaining the competitive PSNR for $\bm x'$ compared to FBI-E.
\yj{The results show }that iterative GS and NCIS, which are trained based solely on self-supervised learning using only original images (\textit{i.e.}, without any adversarial training), have a potential to become a \yj{powerful} purifier for adversarial defense.
Finally, we add the DnCNN + GS ($K=11$), using DnCNN in place of $F_{\bm\theta}(\bm x)$ in (\ref{eqn:SSICS}), and we observe that the general CNN-based denoiser model cannot be used as the smoothing function for the iterative smoothing since it almost perfectly reconstructs the input image, even the adversarial noise in the input.

\section{Experimental Results}

\subsection{Experimental settings}
In this section, we validate our NCIS against various types of attacks on ImageNet validation dataset.
We designed our experimental setting following  \cite{(evaluating)carlini2019evaluating, (benchmarking)dong2020benchmarking} for rigorous verification of our proposed method. Also, we used  attacks and defenses implemented on public packages~\cite{(advertorch)ding2019advertorch, (advertorch)kim2020torchattacks, (ART)art2018, (benchmarking)dong2020benchmarking}.
\vspace{-.2in}
\paragraph{Adversarial attack} 
For \textbf{white-box} attack experiments, we consider that attackers can only access to weights of a classification model, not a purifier. We selected four ImageNet pretrained models, ResNet-152~\cite{(resnet)he2016deep}, WideResNet-101~\cite{(WResNet)zagoruyko2016wide}, ResNeXT-101~\cite{(resnext)xie2017aggregated} and RegNet-32G~\cite{(regnet)radosavovic2020designing}, 
and then we generated adversarial examples of the ImageNet validation dataset using four \textit{untargeted} gradient-based iterative attacks, PGD\cite{(PGD)madry2017towards}, CW\cite{(C&W)carlini2017towards}, MIFGSM\cite{(MIFGSM)dong2018boosting} and DIFGSM\cite{(DIFGSM)xie2019improving}.
Although our paper focuses on \yj{white-box attack} for a classification model, we also conducted experiments on \textit{full} white-box attacks where attacker can both access and are aware of the purifier, and the experimental results are proposed in S.M. 
For \textbf{black-box} attack, we mainly experimented with \textit{transfer-based} attacks which still achieve a strong and efficient attack success rate compared to other types of black-box attack.
For transfer-based attack, we generated adversarial examples by attacking a substitute model with $L_\infty$ PGD attack, and the list of the substitute model 
are listed in S.M.
\vspace{-.2in}
\paragraph{Adversarial defense}
We selected four input transformation based defense methods abbreviated as FS  (depth$=4$)~\cite{(input_transform_bitred)xu2017feature}, JPEG (quality$=90$)~\cite{(JPEG)dziugaite2016study}, TVM ($p=0.3, \lambda=0.5,$ max iteration$ =10$)~\cite{(tvm)rudin1992nonlinear, guo2017countering} and SR  ($\sigma=0.04$)~\cite{(SR)mustafa2019image}. 
Moreover, we implemented the current state-of-the-art purifier NRP and NRP(resG) with their official code. 
Note that NRP (resG) is the lightweight version of NRP and newly proposed by their official code. 
Lastly, we added FD~\cite{(featuredenoising)xie2019feature}, the current state-of-the-art method for adversarial training using ResNet-152, by implementing the code of~\cite{(benchmarking)dong2020benchmarking}. Additionally, we had considered about adding recently proposed purification methods such as SOAP~\cite{(SOAP)shi2020online}, ADP~\cite{(adp)yoon2021adversarial} but neither of them publicized their code nor experimented on ImageNet.
For Gaussian smoothing (GS) used in all experiments including GS in NCIS, we set $\sigma=(K-1)/6$ and only controls the kernel size $K$.
If there are no additional notation, we set $K$ for GS to 5 and only used NCIS consisting of FBI-E ($m=2$) and GS ($K=11$) for all experiments.
We conducted experiments using NCIS trained by three different seeds and report the average result.
The detailed description on the experimental settings, the architecture of FBI-Net and hyperparameters are in S.M.

\begin{table}[t]
\caption{Experimental results of \textit{untargeted} white-box adversarial attack. For $L_\infty$ attacks, we set $\epsilon = 16 / 255$, $\alpha = 1.6 / 255$ and attack iterations = 10. For $L_2$ PGD  attack, we used $\epsilon = 5$ and $\alpha = 0.1$. For $L_2$ CW attack, other than setting attack iterations as 10, we applied default hyperparameters proposed in \cite{(advertorch)kim2020torchattacks}. \textbf{Boldface} denotes our proposed methods, and \textcolor{red}{red} and \textcolor{blue}{blue} denotes the highest and second highest results respectively.}
\vspace{-.1in}
\centering
\smallskip\noindent
\resizebox{\linewidth}{!}{
\begin{tabular}{|cc||c||c|c|c|c|c|}
\hline
\multicolumn{2}{|c|}{Model / Defense}                                                                          & \begin{tabular}[c]{@{}c@{}}Standard\\ Accuracy\end{tabular} & \begin{tabular}[c]{@{}c@{}}CW \\ ($L_2$)\end{tabular} & \begin{tabular}[c]{@{}c@{}}MIFGSM \\ ($L_\infty$)\end{tabular} & \begin{tabular}[c]{@{}c@{}}DIFGSM \\ ($L_\infty$)\end{tabular} & \begin{tabular}[c]{@{}c@{}}PGD \\ ($L_\infty$)\end{tabular} & \begin{tabular}[c]{@{}c@{}}PGD \\ ($L_2$)\end{tabular} \\ \hline \hline
\multicolumn{1}{|c|}{\multirow{9}{*}{\rotatebox[origin=c]{90}{ResNet-152}}}     & \begin{tabular}[c]{@{}c@{}}W/o defense\end{tabular} & 78.25                                                       & 9.37                                                  & 6.34                                                           & 0.43                                                           & 6.20                                                        & 10.66                                                  \\ \cline{2-8} 
\multicolumn{1}{|c|}{}                                & JPEG                                                   & \textcolor{red}{78.13}                                                       & 26.78                                                 & 6.36                                                           & 0.61                                                           & 6.25                                                        & 29.01                                                  \\ \cline{2-8} 
\multicolumn{1}{|c|}{}                                & FS                                                     & \textcolor{blue}{76.66}                                                       & 46.37                                                 & 6.35                                                           & 0.46                                                           & 6.22                                                        & 41.08                                                  \\ \cline{2-8} 
\multicolumn{1}{|c|}{}                                & TVM                                                    & 69.84                                                       & 59.37                                                 & 9.32                                                           & 5.02                                                           & 17.41                                                       & 59.15                                                  \\ \cline{2-8} 
\multicolumn{1}{|c|}{}                                & SR                                                     & 77.24                                                       & 40.87                                                 & 6.36                                                           & 0.05                                                           & 6.22                                                        & 31.06                                                  \\ \cline{2-8} 
\multicolumn{1}{|c|}{}                                & NRP (resG)                                              & 72.24                                                       & 58.05                                                 & 16.39                                                          & 2.40                                                           & 9.60                                                        & 55.84                                                  \\ \cline{2-8} 
\multicolumn{1}{|c|}{}                                & NRP                                                    & 74.04                                                       & 58.16                                                 & 12.35                                                          & 2.59                                                           & 10.71                                                       & 55.58                                                  \\ \cline{2-8} 
\multicolumn{1}{|c|}{}                                & \textbf{GS ($i=7$)}                                            & \textbf{63.32}                                              & \textbf{\textcolor{blue}{60.36}}                                        & \textbf{\textcolor{blue}{24.28}}                                                 & \textbf{\textcolor{blue}{22.28}}                                                 & \textbf{\textcolor{blue}{44.92}}                                              & \textbf{\textcolor{blue}{60.65}}                                         \\ \cline{2-8} 
\multicolumn{1}{|c|}{}                                & \textbf{NCIS ($i=7$)}                                          & \textbf{68.93}                                              & \textbf{\textcolor{red}{64.32}}                                        & \textbf{\textcolor{red}{39.05}}                                                 & \textbf{\textcolor{red}{33.29}}                                                 & \textbf{\textcolor{red}{48.06}}                                              & \textbf{\textcolor{red}{64.28}}                                         \\ \hline \hline
\multicolumn{1}{|c|}{\multirow{9}{*}{\rotatebox[origin=c]{90}{WideResNet-101}}} & \begin{tabular}[c]{@{}c@{}}W/o defense\end{tabular} & 78.91                                                       & 9.42                                                  & 6.96                                                           & 0.31                                                           & 6.60                                                        & 11.97                                                  \\ \cline{2-8} 
\multicolumn{1}{|c|}{}                                & JPEG                                                   & \textcolor{red}{78.22}                                                       & 41.55                                                 & 6.95                                                           & 0.10                                                           & 6.91                                                        & 33.19                                                  \\ \cline{2-8} 
\multicolumn{1}{|c|}{}                                & FS                                                     & 77.03                                                       & 48.82                                                 & 6.95                                                           & 0.10                                                           & 6.91                                                        & 46.41                                                  \\ \cline{2-8} 
\multicolumn{1}{|c|}{}                                & TVM                                                    & 69.17                                                       & \textcolor{red}{69.18}                                                 & 12.15                                                          & 7.56                                                           & 20.33                                                       & \textcolor{blue}{59.84}                                                  \\ \cline{2-8} 
\multicolumn{1}{|c|}{}                                & SR                                                     & \textcolor{blue}{77.85}                                                       & 40.24                                                 & 6.94                                                           & 0.10                                                           & 6.93                                                        & 33.14                                                  \\ \cline{2-8} 
\multicolumn{1}{|c|}{}                                & NRP (resG)                                              & 72.45                                                       & 58.96                                                 & 19.66                                                          & 4.34                                                           & 12.18                                                       & 58.17                                                  \\ \cline{2-8} 
\multicolumn{1}{|c|}{}                                & NRP                                                    & 74.58                                                       & 58.60                                                 & 14.97                                                          & 5.53                                                           & 13.22                                                       & 57.69                                                  \\ \cline{2-8} 
\multicolumn{1}{|c|}{}                                & \textbf{GS ($i=7$)}                                            & \textbf{60.33}                                              & \textbf{57.88}                                        & \textbf{\textcolor{blue}{28.15}}                                                 & \textbf{\textcolor{blue}{25.30}}                                                 & \textbf{\textcolor{blue}{45.04}}                                              & \textbf{57.96}                                         \\ \cline{2-8} 
\multicolumn{1}{|c|}{}                                & \textbf{NCIS ($i=7$)}                                          & \textbf{68.54}                                              & \textbf{\textcolor{blue}{63.96}}                                        & \textbf{\textcolor{red}{34.18}}                                                 & \textbf{\textcolor{red}{33.98}}                                                 & \textbf{\textcolor{red}{49.26}}                                              & \textbf{\textcolor{red}{64.03}}                                         \\ \hline \hline
\multicolumn{1}{|c|}{\multirow{9}{*}{\rotatebox[origin=c]{90}{ResNeXT-101}}}    & \begin{tabular}[c]{@{}c@{}}W/o defense\end{tabular} & 79.21                                                       & 9.43                                                  & 8.59                                                           & 0.61                                                           & 7.81                                                        & 13.84                                                  \\ \cline{2-8} 
\multicolumn{1}{|c|}{}                                & JPEG                                                   & \textcolor{red}{78.28}                                                       & 43.23                                                 & 8.46                                                           & 0.65                                                           & 7.96                                                        & 36.77                                                  \\ \cline{2-8} 
\multicolumn{1}{|c|}{}                                & FS                                                     & 77.54                                                       & 47.88                                                 & 8.61                                                           & 0.62                                                           & 7.94                                                        & 46.14                                                  \\ \cline{2-8} 
\multicolumn{1}{|c|}{}                                & TVM                                                    & 70.66                                                       & 59.47                                                 & 12.52                                                          & 6.68                                                           & 20.67                                                       & 61.05                                                  \\ \cline{2-8} 
\multicolumn{1}{|c|}{}                                & SR                                                     & \textcolor{blue}{78.08}                                                       & 40.45                                                 & 8.59                                                           & 0.65                                                           & 8.0                                                         & 34.31                                                  \\ \cline{2-8} 
\multicolumn{1}{|c|}{}                                & NRP (resG)                                              & 73.65                                                       & 59.04                                                 & 20.57                                                          & 4.14                                                           & 13.00                                                       & 59.53                                                  \\ \cline{2-8} 
\multicolumn{1}{|c|}{}                                & NRP                                                    & 75.28                                                       & 58.30                                                 & 15.88                                                          & 5.23                                                           & 13.58                                                       & 58.15                                                  \\ \cline{2-8} 
\multicolumn{1}{|c|}{}                                & \textbf{GS ($i=7$)}                                            & \textbf{63.78}                                              & \textbf{\textcolor{blue}{60.77}}                                        & \textbf{\textcolor{blue}{28.01}}                                                 & \textbf{\textcolor{blue}{23.50}}                                                 & \textbf{\textcolor{blue}{46.43}}                                              & \textbf{\textcolor{blue}{61.18}}                                         \\ \cline{2-8} 
\multicolumn{1}{|c|}{}                                & \textbf{NCIS ($i=7$)}                                          & \textbf{70.08}                                              & \textbf{\textcolor{red}{65.12}}                                        & \textbf{\textcolor{red}{36.47}}                                                 & \textbf{\textcolor{red}{35.53}}                                                 & \textbf{\textcolor{red}{51.46}}                                              & \textbf{\textcolor{red}{65.57}}                                         \\ \hline \hline
\multicolumn{1}{|c|}{\multirow{9}{*}{\rotatebox[origin=c]{90}{RegNet-32G}}}     & \begin{tabular}[c]{@{}c@{}}W/o defense\end{tabular} & 80.43                                                       & 9.19                                                  & 7.77                                                           & 0.39                                                           & 7.46                                                        & 11.88                                                  \\ \cline{2-8} 
\multicolumn{1}{|c|}{}                                & JPEG                                                   & \textcolor{red}{79.04}                                                       & 52.56                                                 & 7.82                                                           & 0.38                                                           & 7.65                                                        & 47.83                                                  \\ \cline{2-8} 
\multicolumn{1}{|c|}{}                                & FS                                                     & 77.86                                                       & 51.62                                                 & 7.77                                                           & 0.40                                                           & 7.49                                                        & 47.85                                                  \\ \cline{2-8} 
\multicolumn{1}{|c|}{}                                & TVM                                                    & 67.06                                                       & 59.37                                                 & 16.72                                                          & 14.02                                                          & 28.66                                                       & 60.21                                                  \\ \cline{2-8} 
\multicolumn{1}{|c|}{}                                & SR                                                     & \textcolor{blue}{78.38}                                                       & 48.16                                                 & 7.77                                                           & 0.33                                                           & 7.54                                                        & 40.43                                                  \\ \cline{2-8} 
\multicolumn{1}{|c|}{}                                & NRP (resG)                                              & 74.22                                                       & 58.48                                                 & 20.55                                                          & 4.35                                                           & 14.20                                                       & 58.18                                                  \\ \cline{2-8} 
\multicolumn{1}{|c|}{}                                & NRP                                                    & 76.03                                                       & 57.00                                                 & 16.05                                                          & 4.72                                                           & 14.06                                                       & 56.77                                                  \\ \cline{2-8} 
\multicolumn{1}{|c|}{}                                & \textbf{GS ($i=5$)}                                            & \textbf{65.22}                                              & \textbf{\textcolor{blue}{62.71}}                                        & \textbf{\textcolor{blue}{30.83}}                                                 & \textbf{\textcolor{blue}{28.04}}                                                 & \textbf{\textcolor{blue}{51.24}}                                              & \textbf{\textcolor{blue}{63.01}}                                         \\ \cline{2-8} 
\multicolumn{1}{|c|}{}                                & \textbf{NCIS ($i=5$)}                                          & \textbf{71.50}                                              & \textbf{\textcolor{red}{67.37}}                                        & \textbf{\textcolor{red}{41.84}}                                                 & \textbf{\textcolor{red}{31.13}}                                                 & \textbf{\textcolor{red}{53.97}}                                              & \textbf{\textcolor{red}{67.28}}                                         \\ \hline
\end{tabular}
}\vspace{-.2in}
\label{table:whitebox1}
\end{table}

\begin{figure*}[t]
\centering 
\subfigure[Untargeted PGD with various $\epsilon$]
{\includegraphics[width=0.24\linewidth]{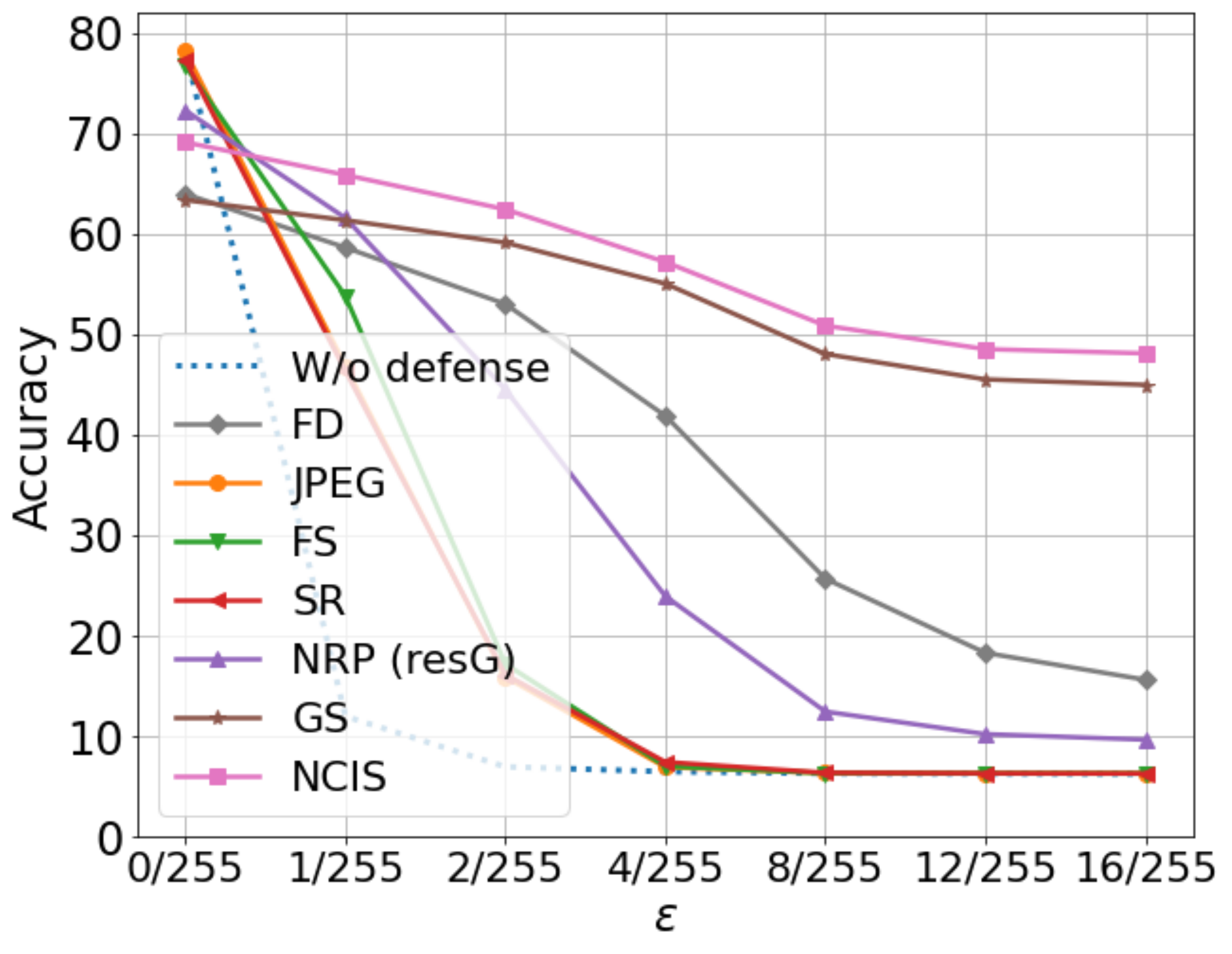}
\label{figure:whitebox_untargeted_eps}}
\subfigure[Untargeted PGD with various iters]
{\includegraphics[width=0.24\linewidth]{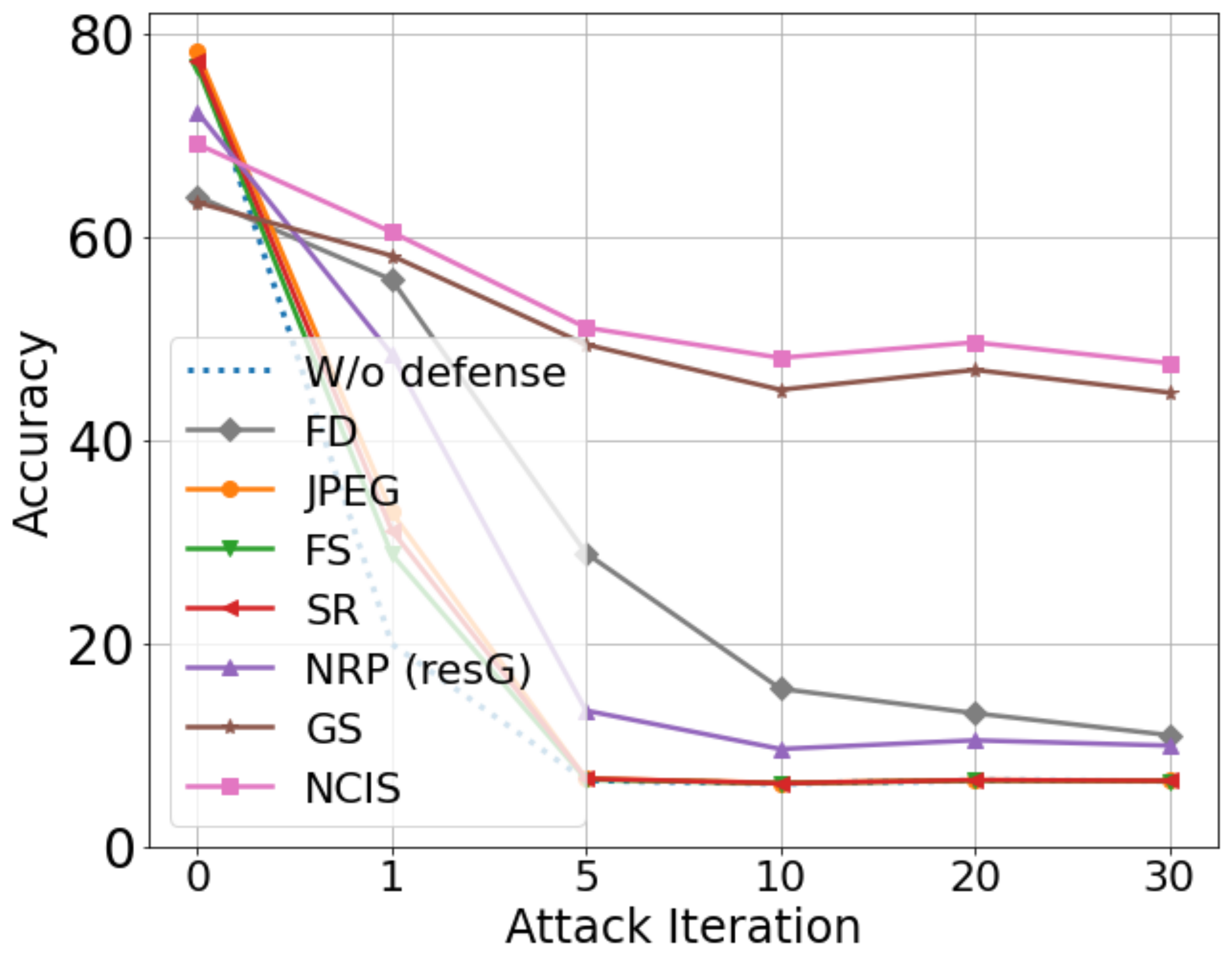}
\label{figure:whitebox_untargeted_iter}}
\subfigure[Targeted PGD with various $\epsilon$]
{\includegraphics[width=0.24\linewidth]{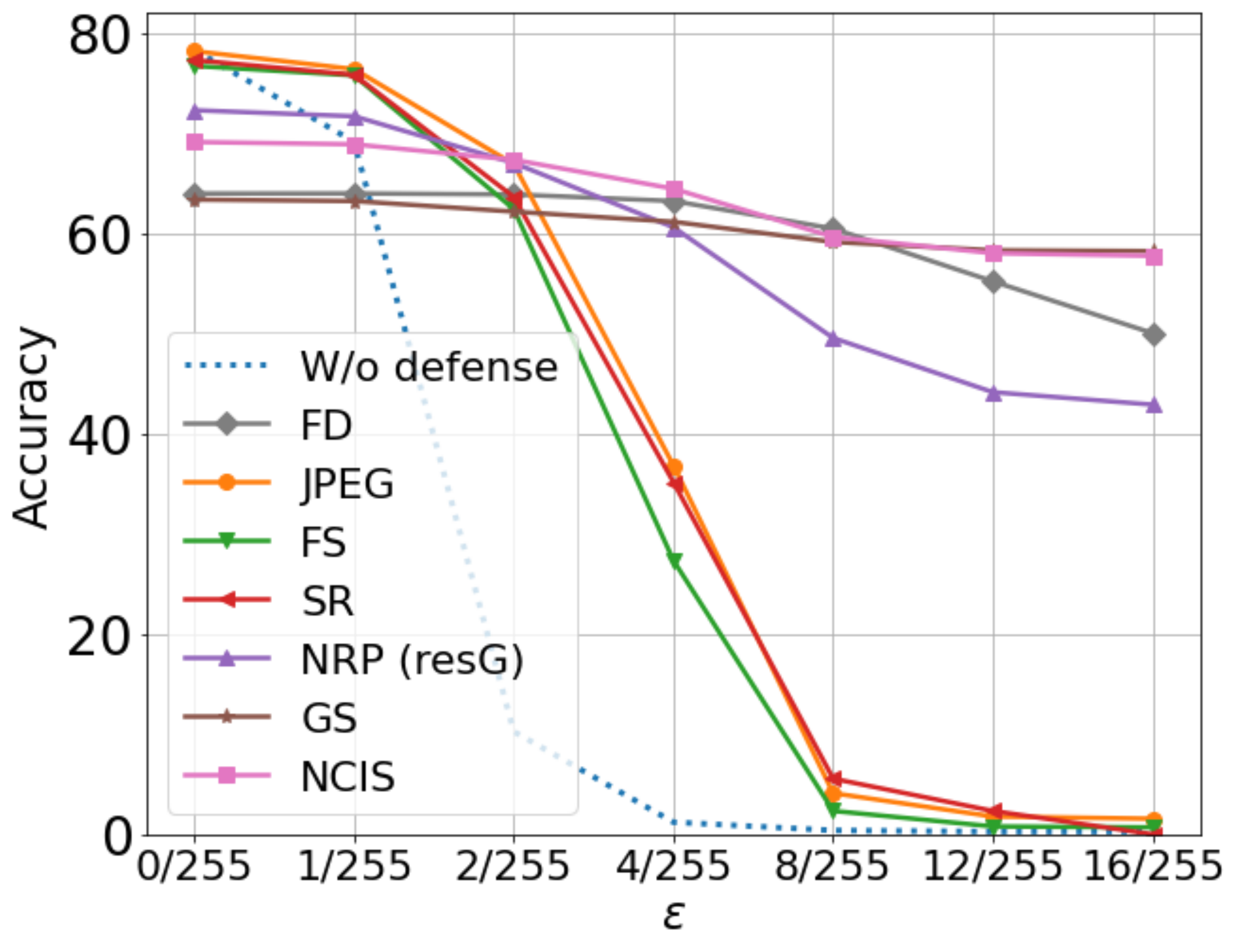}
\label{figure:whitebox_targeted_eps}}
\subfigure[Targeted PGD with various iters]
{\includegraphics[width=0.24\linewidth]{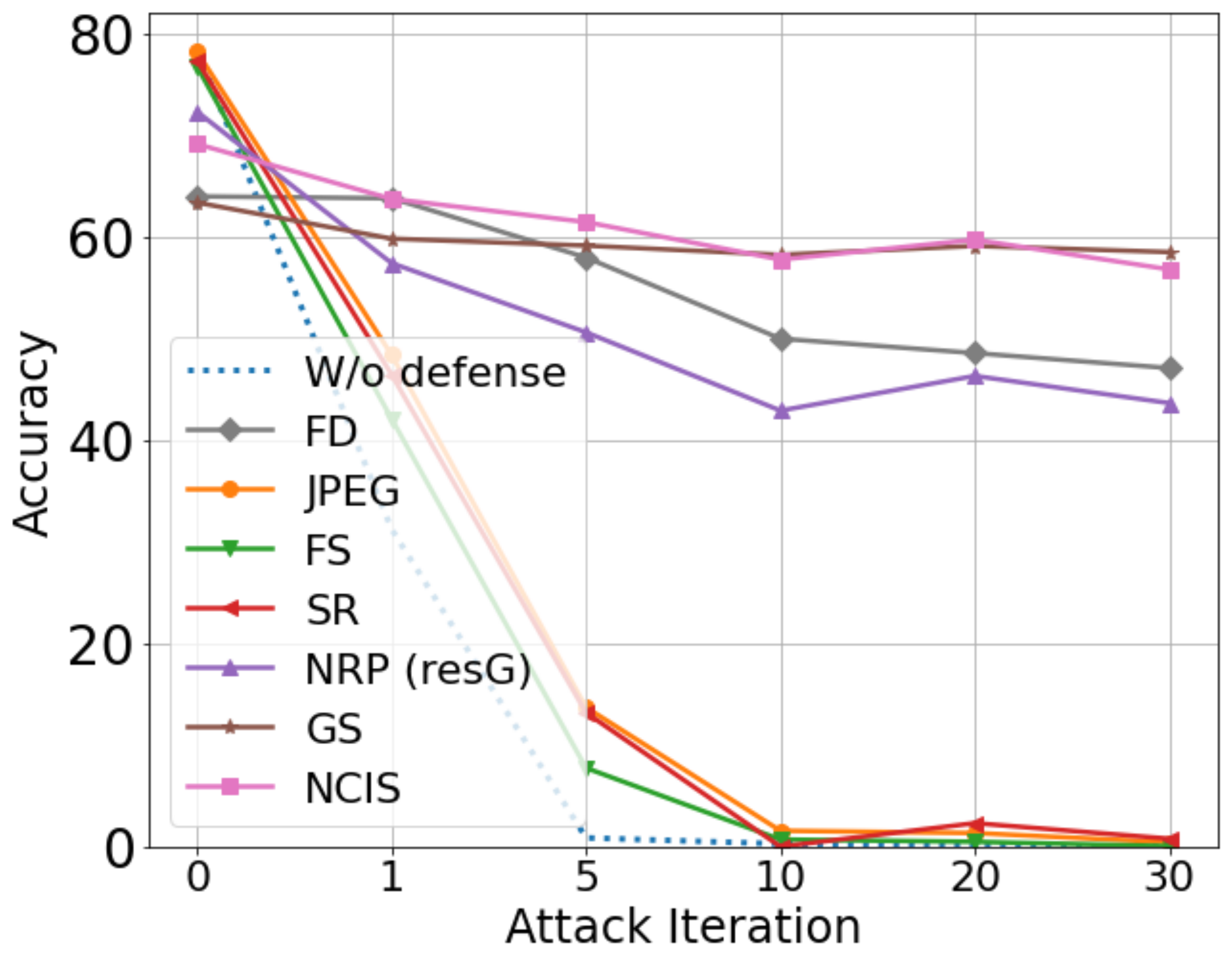}
\label{figure:whitebox_targeted_iter}}\vspace{-.15in}
\caption{Experimental results against $L_\infty$ white-box PGD attacks with ResNet-152. For the experiments on various $\epsilon$, we set $\alpha = 1.6/255$, where $\alpha$ is a step size, with 10 attack iterations.
In the case of experiments for attack iterations, we equally set $\epsilon = 16/255$, and used $\alpha = 1.6/255$ if the number of attack iteration is lower than 10, and used $\alpha = 1/255$ otherwise.}
\vspace{-.2in}
\label{figure:whitebox_pgd}
 \end{figure*}
 
\subsection{Experimental results for white-box attacks}

\begin{table*}[t]
\caption{Experimental results for vision APIs. \textbf{Boldface}, \textcolor{red}{red} and \textcolor{blue}{blue} each denotes the proposed,  highest and second highest result.}\label{table:api_table}
\centering
\vspace{-.1in}
\smallskip\noindent
\resizebox{.9\linewidth}{!}{
\begin{tabular}{|c||cccc||cccc||cccc|}
\hline
\multirow{2}{*}{\begin{tabular}[c]{@{}c@{}}Standard \\ / Robust Acc.\end{tabular}} & \multicolumn{4}{c||}{Prediction Accuracy}                                                                                                                 & \multicolumn{4}{c||}{Top-1 Accuracy}                                                                                                                       & \multicolumn{4}{c|}{Top-5 Accuracy}                                                                                                                       \\ \cline{2-13} 
                                                                                 & \multicolumn{1}{c|}{AWS}                  & \multicolumn{1}{c|}{Azure}                & \multicolumn{1}{c|}{Clarifai}             & Google               & \multicolumn{1}{c|}{AWS}                  & \multicolumn{1}{c|}{Azure}                & \multicolumn{1}{c|}{Clarifai}             & Google               & \multicolumn{1}{c|}{AWS}                  & \multicolumn{1}{c|}{Azure}                & \multicolumn{1}{c|}{Clarifai}             & Google               \\ \hline \hline
W/o defense                                                                      & \multicolumn{1}{c|}{1.00 / 0.00}          & \multicolumn{1}{c|}{1.00 / 0.00}          & \multicolumn{1}{c|}{1.00 / 0.00}          & 1.00 / 0.00          & \multicolumn{1}{c|}{1.00 / 0.00}          & \multicolumn{1}{c|}{1.00 / 0.00}          & \multicolumn{1}{c|}{1.00 / 0.00}          & 1.00 / 0.00          & \multicolumn{1}{c|}{1.00 / 0.00}          & \multicolumn{1}{c|}{1.00 / 0.00}          & \multicolumn{1}{c|}{1.00 / 0.00}          & 1.00 / 0.00          \\ \hline
JPEG                                                                             & \multicolumn{1}{c|}{\textcolor{red}{0.78} / 0.12}          & \multicolumn{1}{c|}{\textcolor{red}{0.82} / 0.08}          & \multicolumn{1}{c|}{0.78 / 0.58}          & \textcolor{red}{0.75} / 0.10          & \multicolumn{1}{c|}{\textcolor{blue}{0.75} / 0.04}          & \multicolumn{1}{c|}{0.72 / 0.02}          & \multicolumn{1}{c|}{\textcolor{blue}{0.79} / 0.45}          & \textcolor{red}{0.60} / 0.04          & \multicolumn{1}{c|}{\textcolor{red}{0.91} / 0.14}          & \multicolumn{1}{c|}{0.94 / 0.12}          & \multicolumn{1}{c|}{0.95 / \textcolor{blue}{0.80}}          & \textcolor{red}{0.84} / 0.13          \\ \hline
FS                                                                               & \multicolumn{1}{c|}{0.60 / 0.21}          & \multicolumn{1}{c|}{0.60 / 0.05}          & \multicolumn{1}{c|}{0.76 / 0.56}          & 0.55 / 0.11          & \multicolumn{1}{c|}{0.44 / 0.08}          & \multicolumn{1}{c|}{0.42 / 0.06}          & \multicolumn{1}{c|}{0.69 / 0.45}          & 0.43 / 0.07          & \multicolumn{1}{c|}{0.72 / 0.23}          & \multicolumn{1}{c|}{0.67 / 0.10}          & \multicolumn{1}{c|}{0.94 / 0.77}          & 0.67 / 0.14          \\ \hline
SR                                                                               & \multicolumn{1}{c|}{\textcolor{red}{0.78} / 0.19}          & \multicolumn{1}{c|}{\textcolor{blue}{0.81} / 0.12}          & \multicolumn{1}{c|}{\textcolor{blue}{0.85} / 0.62}          & 0.69 / 0.13          & \multicolumn{1}{c|}{\textcolor{blue}{0.72} / 0.06}          & \multicolumn{1}{c|}{\textcolor{blue}{0.75} / 0.10}          & \multicolumn{1}{c|}{\textcolor{red}{0.87} / \textcolor{blue}{0.52}}          & 0.46 / 0.07          & \multicolumn{1}{c|}{\textcolor{blue}{0.89} / \textcolor{blue}{0.25}}          & \multicolumn{1}{c|}{\textcolor{red}{0.95} / 0.17}          & \multicolumn{1}{c|}{\textcolor{red}{0.99} / \textcolor{blue}{0.80}}          & \textcolor{black}{0.77} / 0.18          \\ \hline
NRP (resG)                                                                       & \multicolumn{1}{c|}{0.45 / 0.19}          & \multicolumn{1}{c|}{0.47 / 0.16}          & \multicolumn{1}{c|}{0.45/ 0.19}           & 0.39 / 0.14          & \multicolumn{1}{c|}{0.30 / 0.09}          & \multicolumn{1}{c|}{0.34 / 0.12}          & \multicolumn{1}{c|}{0.30 / 0.09}          & 0.26 / 0.09          & \multicolumn{1}{c|}{0.57 / 0.18}          & \multicolumn{1}{c|}{0.54 / 0.18}          & \multicolumn{1}{c|}{0.57 / 0.18}          & 0.47 / 0.16          \\ \hline
\textbf{GS (K=5)}                                                                & \multicolumn{1}{c|}{\textbf{\textcolor{blue}{0.68} / \textcolor{blue}{0.26}}} & \multicolumn{1}{c|}{\textbf{0.63 / \textcolor{blue}{0.23}}} & \multicolumn{1}{c|}{\textbf{0.79 / \textcolor{red}{0.65}}} & \textbf{0.55 / \textcolor{blue}{0.21}} & \multicolumn{1}{c|}{\textbf{0.45 / \textcolor{blue}{0.11}}} & \multicolumn{1}{c|}{\textbf{0.53 / \textcolor{red}{0.20}}} & \multicolumn{1}{c|}{\textbf{0.72 / \textcolor{red}{0.55}}} & \textbf{0.42 / \textcolor{blue}{0.11}} & \multicolumn{1}{c|}{\textbf{0.80 / \textcolor{blue}{0.25}}} & \multicolumn{1}{c|}{\textbf{\textcolor{blue}{0.78} / \textcolor{red}{0.30}}} & \multicolumn{1}{c|}{\textbf{\textcolor{red}{0.99} / \textcolor{red}{0.86}}} & \textbf{0.69 / \textcolor{blue}{0.23}} \\ \hline
\textbf{NCIS}                                                                    & \multicolumn{1}{c|}{\textbf{0.74 / \textcolor{red}{0.28}}} & \multicolumn{1}{c|}{\textbf{0.80 / 0.21}} & \multicolumn{1}{c|}{\textbf{\textcolor{red}{0.86} / \textcolor{blue}{0.62}}} & \textbf{\textcolor{blue}{0.72} / \textcolor{red}{0.27}} & \multicolumn{1}{c|}{\textbf{0.60 / \textcolor{red}{0.13}}} & \multicolumn{1}{c|}{\textbf{\textcolor{red}{0.78} / \textcolor{blue}{0.13}}} & \multicolumn{1}{c|}{\textbf{0.76 / \textcolor{red}{0.55}}} & \textbf{\textcolor{blue}{0.54} / \textcolor{red}{0.20}} & \multicolumn{1}{c|}{\textbf{0.88 / \textcolor{red}{0.31}}} & \multicolumn{1}{c|}{\textbf{\textcolor{red}{0.95} / \textcolor{blue}{0.29}}} & \multicolumn{1}{c|}{\textbf{\textcolor{blue}{0.98} / \textcolor{blue}{0.80}}} & \textbf{\textcolor{blue}{0.83} / \textcolor{red}{0.37}} \\ \hline
\end{tabular}
}\vspace{-.2in}
\label{table:api}
\end{table*}

\paragraph{Experiments with various white-box attacks}
Table \ref{table:whitebox1} shows experimental results of each defense method against five attacks for four different ImageNet pretrained classification models.
First, as for standard accuracy, JPEG achieved the best results. However, as already presented in \cite{(benchmarking)dong2020benchmarking, guo2017countering}, traditional input transformation-based methods are easily broken by white-box attacks. 
Second, GS achieves superior robust accuracy in most cases and even outperforms NRP and NRP (resG). \yj{One possible reason} NRP and NRP (resG) show deteriorated performance, compared to its original paper, is that the proposed loss function for training NRP and NRP (resG) is not well generalized to purify adversarial examples generated from various types of attack.
On the other hand, because our GS is based on the findings of the characteristic of adversarial noise, it can be applied to purify varied types of adversarial examples.
Furthermore, NCIS consistently surpasses robust accuracy of GS and other baselines.

Additionally, we compared inference time, GPU memory requirement, and the number of parameters of each method in S.M. NRP has $\times 41.5$ model parameters and requires $\times 14.7$ GPU memory with $14\%$ slow inference time than NCIS.
Since we already observe that there is no big difference in performance between NRP and NRP (resG) and the computational cost of NRP is too huge, \yj{we only use NRP (resG) for remaining experiments.}
 \vspace{-.2in}
\paragraph{Experiments with PGD attack with various settings}
To evaluate the proposed method more rigorously, we experimented with both  \textit{targeted} and \textit{untargeted} $L_\infty$ PGD attack with various $\epsilon$ and the number of attack iterations.
Additionally, we report the result of FD \cite{(featuredenoising)xie2019feature} as the baseline.

Figure \ref{figure:whitebox_untargeted_eps} and \ref{figure:whitebox_untargeted_iter} show the experimental results for \textit{untargeted} $L_\infty$ PGD attack in ResNet-152~\cite{(resnet)he2016deep}.
As $\epsilon$ and the number of attack iterations increase, the robust accuracy of baseline defense methods goes significantly down. Besides, FD has lower standard accuracy (63.96) compared to NCIS. Moreover, input transformation-based methods and NRP are crumbling like a wreck when the attack setting is strong. 
However, NCIS and GS achieve the highest robust accuracy, showing up to four times robust accuracy of FD. Besides, both standard and robust accuracy of NCIS are higher than GS in all settings. 

On the other hand, Figure \ref{figure:whitebox_targeted_eps} and \ref{figure:whitebox_targeted_iter} show experiments on \textit{targeted} $L_\infty$ PGD attack. We observe that the tendency of experimental results has changed from the \textit{untargeted} results.
First, FD and NRP (resG) show better performance than the \textit{untargeted} attack case. We presume that both FD and NRP (resG) have a generalization problem and fail to stably defense all types of attacks.
Second, the robust accuracy of \naeun{FD and NRP} degrade as the number of iterations and $\epsilon$ increase, resulting in lower robust accuracy than our methods against strong attacks.
On top of this, NCIS and GS achieve superior robust accuracy  against strong attacks.
See S.M for the results on other classifiers, showing similar tendency.
\subsection{Experimental results for black-box attacks}
\begin{figure}[h]
\centering 
 \vspace{-.15in}
\subfigure[Experiments for various $\epsilon$]
{\includegraphics[width=0.48\linewidth]{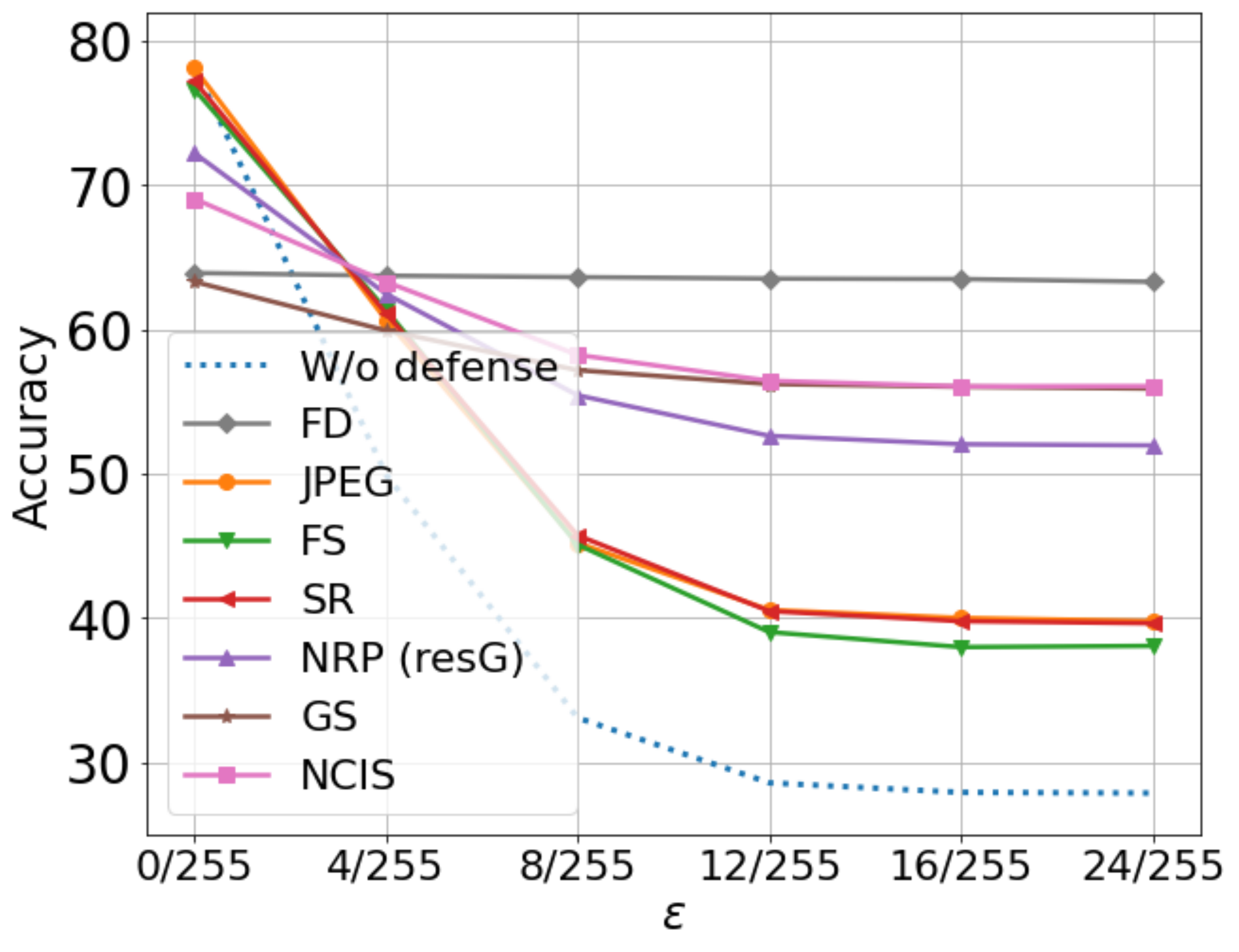}
\label{figure:blackbox_eps_resnet}}
\subfigure[Experiments for various iters]
{\includegraphics[width=0.48\linewidth]{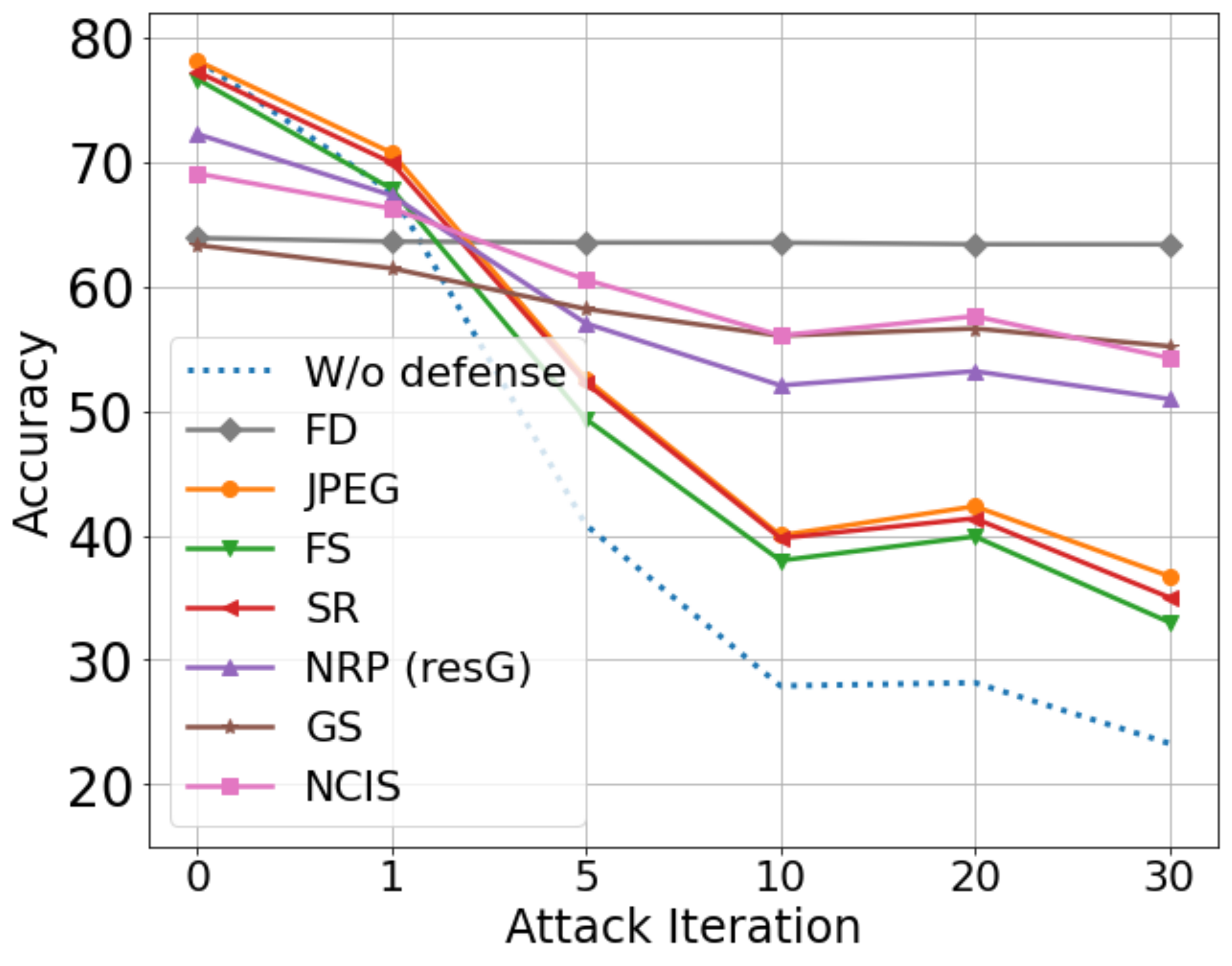}
\label{figure:blackbox_eps_wideresnet}}\vspace{-.15in}
\caption{Experimental results of transfer-based black-box attack with $L_\infty$ PGD attack for ResNet-152.}
\vspace{-.15in}
\label{figure:blackbox_pgd}
 \end{figure}

\paragraph{Transfer-based black-box attack}
Figure \ref{figure:blackbox_pgd} shows ResNet-152 results attacked by adversarial examples generated by attacking WideResNet-101~\cite{(WResNet)zagoruyko2016wide}.
The notable discovery is that FD shows the almost constant but most robust result for all cases. 
Also, different from white-box attack cases, NRP (resG) achieves the competitive performance compared to other input transformation-based methods, supporting the result of their paper.
However, note that our GS and NCIS not only surpass the performance of NRP (resG) but also NCIS achieves the competitive performance compared to FD when considering both standard and robust accuracy. Additionally, we conducted experiments with the state-of-the-art score-based black-box attack (Square \cite{(square)andriushchenko2020square}), and the experimental results are proposed in S.M.

\vspace{-.2in}
\paragraph{Experiment with APIs}
\cite{(ensemble_attack)liu2016delving} showed existing APIs for multi-label classification can be fooled by ensemble transfer-based black-box attack.
Based on this finding, we propose a new experiment for evaluating purification methods using APIs of Azure, AWS, Clarifai and Google.
From this experiment, we evaluate how well each purifier can restore the top five labels predicted from a clean image when given strong adversarial examples. 
To evaluate each purifier, we used three evaluation metrics: \textit{Prediction Accuracy}, \textit{Top-1 Accuracy} and \textit{Top-5 Accuracy}. 
The additional details of metrics and used hyperparameters are noted in S.M. 

Table \ref{table:api_table} shows experimental results on the generated test dataset.
First, we observe that traditional input transformation-based methods show generally high standard accuracy with competitive robust accuracy compared to NRP (resG).
We believe that this result shows another example of insufficient generalization of NRP (resG) for purifying various types of adversarial example.
Second, NCIS and GS show the most uniformly competitive performance for various APIs.
Among them, NCIS achieves better purification performance than GS, considering the average of standard and robust accuracy.

Note that this setting is challenging to both attacker and purifier since there is no model nor detailed information about API. 
However, we believe that this setting is more common in real world than the settings of white-box or strong adaptive attack.
Hence, this benchmark is practical and useful, especially for evaluating the efficiency of purifiers properly.



\subsection{Ablation study}

To verify each proposed module, we conducted ablation study and the results are shown in Table \ref{table:ablation_study}.
For experiments, we used the ImageNet validation dataset and adversarial examples of it generated by $L_\infty$ PGD  ($\epsilon=16/255, \alpha=1.6/255$) attack with 10 attack iterations.
The first row shows the result of NCIS with complete components.
The second row is excluding GS ($K=11$) from NCIS, which is named as FBI-E ($m=2$). It shows that both standard and robust accuracy slightly drop because it becomes difficult to reconstruct given images. 
The third and fourth row show the result of FBI + GS ($K=11$) and FBI respectively.
We clearly observe that not only inference time and GPU memory requirement significantly increase but also both standard and robust accuracy decrease after removing the extension operations.
The fifth and sixth row is the result of both GS cases and, as already checked in previous experiments, GS ($K=5$) achieves remarkable performance but GS ($K=11$) alone doesn't.

\begin{table}[h]
\vspace{-.1in}
\caption{Experimental results for ablation studies.}
\vspace{-.1in}
\centering
\smallskip\noindent
\resizebox{.98\linewidth}{!}{
\begin{tabular}{|c|c|c|c||c|c|c|c|}
\hline
\begin{tabular}[c]{@{}c@{}}GS\\ ($K=5$)\end{tabular} & \begin{tabular}[c]{@{}c@{}}GS\\ ($K=11$)\end{tabular} & FBI        & \begin{tabular}[c]{@{}c@{}}Extension\\ ($m=2$)\end{tabular}  & \begin{tabular}[c]{@{}c@{}}Standard\\ Accuracy\end{tabular} & \begin{tabular}[c]{@{}c@{}}Robust\\ Accuracy\end{tabular} & \begin{tabular}[c]{@{}c@{}}Inference\\ Time\end{tabular} & \begin{tabular}[c]{@{}c@{}}GPU\\ Memory\end{tabular} \\ \hline \hline
\textbf{\xmark}                                             & \textbf{\cmark}                                              & \textbf{\cmark} & \textbf{\cmark} & \textbf{69.07}                                              & \textbf{48.06}                                            & \textbf{0.0779}                                          & \textbf{0.60G}                                       \\ \hline 
\xmark                                                      & \xmark                                                       & \cmark          & \cmark          & 65.08                                                       & 46.07                                                     & 0.0669                                                   & 0.60G                                                \\ \hline
\xmark                                                      & \cmark                                                       & \cmark          & \xmark          & 66.51                                                       & 21.37                                                     & 0.1743                                                   & 2.13G                                                \\ \hline
\xmark                                                      & \xmark                                                       & \cmark          & \xmark          & 67.62                                                       & 39.93                                                     & 0.1636                                                   & 2.13G                                                \\ \hline
\cmark                                                      & \xmark                                                       & \xmark          & \xmark          & 63.32                                                       & 44.92                                                     & $4\times 10^{-5}$                                                  & 0.002G                                               \\ \hline
X                                                      & \cmark                                                       & \xmark          & \xmark          & 21.35                                                       & 19.98                                                     & $4\times 10^{-5}$                                                  & 0.002G                                               \\ \hline
\end{tabular}}
\label{table:ablation_study}
\vspace{-.2in}
\end{table}

\section{Concluding Remarks}
We proposed Neural Contextual Iterative Smoothing (NCIS) for adversarial purification against white- and black-box attack to the classifier.
Starting from the novel observation that adversarial noise has almost zero mean and a symmetric distribution, we showed that the simple iterative Gaussian smoothing can purify adversarial perturbations.
To further improve it, we proposed the learnable neural network-based smoothing function, named as NCIS, based on the novel usage of FBI-Net combining newly proposed two operations for increasing computational efficiency.
From the extensive experiments, we observed NCIS robustly purifies adversarial examples generated from various types of attack (\textit{e.g.}, $L_2$/$L_\infty$ white-box attacks, targeted/untargeted $L_\infty$ PGD white-box attack, and transfer-based black-box attack) without requiring any adversarial training or re-training of the classification model.
We discussed the limitations of our work in S.M.



{\small
\bibliographystyle{ieee_fullname}
\bibliography{egbib}

\begin{thebibliography}{10}\itemsep=-1pt

\bibitem{(square)andriushchenko2020square}
Maksym Andriushchenko, Francesco Croce, Nicolas Flammarion, and Matthias Hein.
\newblock Square attack: a query-efficient black-box adversarial attack via
  random search.
\newblock In {\em European Conference on Computer Vision}, pages 484--501.
  Springer, 2020.

\bibitem{athalye2018obfuscated}
Anish Athalye, Nicholas Carlini, and David Wagner.
\newblock Obfuscated gradients give a false sense of security: Circumventing
  defenses to adversarial examples.
\newblock In {\em International Conference on Machine Learning}, pages
  274--283. PMLR, 2018.

\bibitem{(obfuscated)athalye2018obfuscated}
Anish Athalye, Nicholas Carlini, and David Wagner.
\newblock Obfuscated gradients give a false sense of security: Circumventing
  defenses to adversarial examples.
\newblock In {\em International conference on machine learning}, pages
  274--283. PMLR, 2018.

\bibitem{(advtr_survey)bai2021recent}
Tao Bai, Jinqi Luo, Jun Zhao, Bihan Wen, and Qian Wang.
\newblock Recent advances in adversarial training for adversarial robustness.
\newblock {\em arXiv preprint arXiv:2102.01356}, 2021.

\bibitem{(fbi)byun2021fbi}
Jaeseok Byun, Sungmin Cha, and Taesup Moon.
\newblock Fbi-denoiser: Fast blind image denoiser for poisson-gaussian noise.
\newblock {\em arXiv preprint arXiv:2105.10967}, 2021.

\bibitem{(evaluating)carlini2019evaluating}
Nicholas Carlini, Anish Athalye, Nicolas Papernot, Wieland Brendel, Jonas
  Rauber, Dimitris Tsipras, Ian Goodfellow, Aleksander Madry, and Alexey
  Kurakin.
\newblock On evaluating adversarial robustness.
\newblock {\em arXiv preprint arXiv:1902.06705}, 2019.

\bibitem{(C&W)carlini2017towards}
Nicholas Carlini and David Wagner.
\newblock Towards evaluating the robustness of neural networks.
\newblock In {\em 2017 ieee symposium on security and privacy (sp)}, pages
  39--57. IEEE, 2017.

\bibitem{(NAIDE)cha2018neural}
Sungmin Cha and Taesup Moon.
\newblock Neural adaptive image denoiser.
\newblock In {\em IEEE International Conference on Acoustics, Speech and Signal
  Processing (ICASSP)}, pages 2981--2985, 2018.

\bibitem{(FCAIDE)cha2019fully}
Sungmin Cha and Taesup Moon.
\newblock Fully convolutional pixel adaptive image denoiser.
\newblock In {\em IEEE International Conference on Computer Vision (ICCV)},
  pages 4160--4169, 2019.

\bibitem{(zoo)chen2017zoo}
Pin-Yu Chen, Huan Zhang, Yash Sharma, Jinfeng Yi, and Cho-Jui Hsieh.
\newblock Zoo: Zeroth order optimization based black-box attacks to deep neural
  networks without training substitute models.
\newblock In {\em Proceedings of the 10th ACM workshop on artificial
  intelligence and security}, pages 15--26, 2017.

\bibitem{(certified)cohen2019certified}
Jeremy Cohen, Elan Rosenfeld, and Zico Kolter.
\newblock Certified adversarial robustness via randomized smoothing.
\newblock In {\em International Conference on Machine Learning}, pages
  1310--1320. PMLR, 2019.

\bibitem{(imagenet)deng2009imagenet}
Jia Deng, Wei Dong, Richard Socher, Li-Jia Li, Kai Li, and Li Fei-Fei.
\newblock Imagenet: A large-scale hierarchical image database.
\newblock In {\em 2009 IEEE conference on computer vision and pattern
  recognition}, pages 248--255. Ieee, 2009.

\bibitem{(advertorch)ding2019advertorch}
Gavin~Weiguang Ding, Luyu Wang, and Xiaomeng Jin.
\newblock {AdverTorch} v0.1: An adversarial robustness toolbox based on
  pytorch.
\newblock {\em arXiv preprint arXiv:1902.07623}, 2019.

\bibitem{(benchmarking)dong2020benchmarking}
Yinpeng Dong, Qi-An Fu, Xiao Yang, Tianyu Pang, Hang Su, Zihao Xiao, and Jun
  Zhu.
\newblock Benchmarking adversarial robustness on image classification.
\newblock In {\em Proceedings of the IEEE/CVF Conference on Computer Vision and
  Pattern Recognition}, pages 321--331, 2020.

\bibitem{(MIFGSM)dong2018boosting}
Yinpeng Dong, Fangzhou Liao, Tianyu Pang, Hang Su, Jun Zhu, Xiaolin Hu, and
  Jianguo Li.
\newblock Boosting adversarial attacks with momentum.
\newblock In {\em Proceedings of the IEEE conference on computer vision and
  pattern recognition}, pages 9185--9193, 2018.

\bibitem{(JPEG)dziugaite2016study}
Gintare~Karolina Dziugaite, Zoubin Ghahramani, and Daniel~M Roy.
\newblock A study of the effect of jpg compression on adversarial images.
\newblock {\em arXiv preprint arXiv:1608.00853}, 2016.

\bibitem{(FGSM)goodfellow2014explaining}
Ian~J Goodfellow, Jonathon Shlens, and Christian Szegedy.
\newblock Explaining and harnessing adversarial examples.
\newblock {\em arXiv preprint arXiv:1412.6572}, 2014.

\bibitem{(skewness)groeneveld1984measuring}
Richard~A Groeneveld and Glen Meeden.
\newblock Measuring skewness and kurtosis.
\newblock {\em Journal of the Royal Statistical Society: Series D (The
  Statistician)}, 33(4):391--399, 1984.

\bibitem{guo2017countering}
Chuan Guo, Mayank Rana, Moustapha Cisse, and Laurens Van Der~Maaten.
\newblock Countering adversarial images using input transformations.
\newblock {\em arXiv preprint arXiv:1711.00117}, 2017.

\bibitem{(resnet)he2016deep}
Kaiming He, Xiangyu Zhang, Shaoqing Ren, and Jian Sun.
\newblock Deep residual learning for image recognition.
\newblock In {\em Proceedings of the IEEE conference on computer vision and
  pattern recognition}, pages 770--778, 2016.

\bibitem{(NES)ilyas2018black}
Andrew Ilyas, Logan Engstrom, Anish Athalye, and Jessy Lin.
\newblock Black-box adversarial attacks with limited queries and information.
\newblock In {\em International Conference on Machine Learning}, pages
  2137--2146. PMLR, 2018.

\bibitem{(advertorch)kim2020torchattacks}
Hoki Kim.
\newblock Torchattacks: A pytorch repository for adversarial attacks.
\newblock {\em arXiv preprint arXiv:2010.01950}, 2020.

\bibitem{(N2V)krull2018noise2void}
Alexander Krull, Tim-Oliver Buchholz, and Florian Jug.
\newblock Noise2void-learning denoising from single noisy images.
\newblock In {\em IEEE Conference on Computer Vision and Pattern Recognition
  (CVPR)}, pages 2129--2137, 2019.

\bibitem{(BIM)kurakin2016adversarial}
Alexey Kurakin, Ian Goodfellow, Samy Bengio, et~al.
\newblock Adversarial examples in the physical world, 2016.

\bibitem{(HQDenoising)laine2019high}
Samuli Laine, Tero Karras, Jaakko Lehtinen, and Timo Aila.
\newblock High-quality self-supervised deep image denoising.
\newblock In {\em Advances in Neural Information Processing Systems (NIPS)},
  pages 6968--6978, 2019.

\bibitem{(denoiser)liao2018defense}
Fangzhou Liao, Ming Liang, Yinpeng Dong, Tianyu Pang, Xiaolin Hu, and Jun Zhu.
\newblock Defense against adversarial attacks using high-level representation
  guided denoiser.
\newblock In {\em Proceedings of the IEEE Conference on Computer Vision and
  Pattern Recognition}, pages 1778--1787, 2018.

\bibitem{(ensemble_attack)liu2016delving}
Yanpei Liu, Xinyun Chen, Chang Liu, and Dawn Song.
\newblock Delving into transferable adversarial examples and black-box attacks.
\newblock {\em arXiv preprint arXiv:1611.02770}, 2016.

\bibitem{(PGD)madry2017towards}
Aleksander Madry, Aleksandar Makelov, Ludwig Schmidt, Dimitris Tsipras, and
  Adrian Vladu.
\newblock Towards deep learning models resistant to adversarial attacks.
\newblock {\em arXiv preprint arXiv:1706.06083}, 2017.

\bibitem{(DeepFool)moosavi2016deepfool}
Seyed-Mohsen Moosavi-Dezfooli, Alhussein Fawzi, and Pascal Frossard.
\newblock Deepfool: a simple and accurate method to fool deep neural networks.
\newblock In {\em Proceedings of the IEEE conference on computer vision and
  pattern recognition}, pages 2574--2582, 2016.

\bibitem{(SR)mustafa2019image}
Aamir Mustafa, Salman~H Khan, Munawar Hayat, Jianbing Shen, and Ling Shao.
\newblock Image super-resolution as a defense against adversarial attacks.
\newblock {\em IEEE Transactions on Image Processing}, 29:1711--1724, 2019.

\bibitem{(NRP)naseer2020self}
Muzammal Naseer, Salman Khan, Munawar Hayat, Fahad~Shahbaz Khan, and Fatih
  Porikli.
\newblock A self-supervised approach for adversarial robustness.
\newblock In {\em Proceedings of the IEEE/CVF Conference on Computer Vision and
  Pattern Recognition}, pages 262--271, 2020.

\bibitem{(ART)art2018}
Maria-Irina Nicolae, Mathieu Sinn, Minh~Ngoc Tran, Beat Buesser, Ambrish Rawat,
  Martin Wistuba, Valentina Zantedeschi, Nathalie Baracaldo, Bryant Chen, Heiko
  Ludwig, Ian Molloy, and Ben Edwards.
\newblock Adversarial robustness toolbox v1.2.0.
\newblock {\em CoRR}, 1807.01069, 2018.

\bibitem{(regnet)radosavovic2020designing}
Ilija Radosavovic, Raj~Prateek Kosaraju, Ross Girshick, Kaiming He, and Piotr
  Doll{\'a}r.
\newblock Designing network design spaces.
\newblock In {\em Proceedings of the IEEE/CVF Conference on Computer Vision and
  Pattern Recognition}, pages 10428--10436, 2020.

\bibitem{(tvm)rudin1992nonlinear}
Leonid~I Rudin, Stanley Osher, and Emad Fatemi.
\newblock Nonlinear total variation based noise removal algorithms.
\newblock {\em Physica D: nonlinear phenomena}, 60(1-4):259--268, 1992.

\bibitem{(denoisedsmoothing)salman2020denoised}
Hadi Salman, Mingjie Sun, Greg Yang, Ashish Kapoor, and J~Zico Kolter.
\newblock Denoised smoothing: A provable defense for pretrained classifiers.
\newblock {\em Advances in Neural Information Processing Systems}, 33, 2020.

\bibitem{(defense-gan)samangouei2018defense}
Pouya Samangouei, Maya Kabkab, and Rama Chellappa.
\newblock Defense-gan: Protecting classifiers against adversarial attacks using
  generative models.
\newblock {\em arXiv preprint arXiv:1805.06605}, 2018.

\bibitem{(advtrfree)shafahi2019adversarial}
Ali Shafahi, Mahyar Najibi, Amin Ghiasi, Zheng Xu, John Dickerson, Christoph
  Studer, Larry~S Davis, Gavin Taylor, and Tom Goldstein.
\newblock Adversarial training for free!
\newblock {\em arXiv preprint arXiv:1904.12843}, 2019.

\bibitem{(ape-gan)shen2017ape}
Shiwei Shen, Guoqing Jin, Ke Gao, and Yongdong Zhang.
\newblock Ape-gan: Adversarial perturbation elimination with gan.
\newblock {\em arXiv preprint arXiv:1707.05474}, 2017.

\bibitem{(SOAP)shi2020online}
Changhao Shi, Chester Holtz, and Gal Mishne.
\newblock Online adversarial purification based on self-supervised learning.
\newblock In {\em International Conference on Learning Representations}, 2020.

\bibitem{shi2021online}
Changhao Shi, Chester Holtz, and Gal Mishne.
\newblock Online adversarial purification based on self-supervision.
\newblock {\em arXiv preprint arXiv:2101.09387}, 2021.

\bibitem{(adaptive_attack)tramer2020adaptive}
Florian Tramer, Nicholas Carlini, Wieland Brendel, and Aleksander Madry.
\newblock On adaptive attacks to adversarial example defenses.
\newblock {\em arXiv preprint arXiv:2002.08347}, 2020.

\bibitem{(SPSA)uesato2018adversarial}
Jonathan Uesato, Brendan O’donoghue, Pushmeet Kohli, and Aaron Oord.
\newblock Adversarial risk and the dangers of evaluating against weak attacks.
\newblock In {\em International Conference on Machine Learning}, pages
  5025--5034. PMLR, 2018.

\bibitem{(DBSN)wu2020unpaired}
Xiaohe Wu, Ming Liu, Yue Cao, Dongwei Ren, and Wangmeng Zuo.
\newblock Unpaired learning of deep image denoising.
\newblock In {\em European Conference on Computer Vision (ECCV)}, pages
  352--368, 2020.

\bibitem{(featuredenoising)xie2019feature}
Cihang Xie, Yuxin Wu, Laurens van~der Maaten, Alan~L Yuille, and Kaiming He.
\newblock Feature denoising for improving adversarial robustness.
\newblock In {\em Proceedings of the IEEE/CVF Conference on Computer Vision and
  Pattern Recognition}, pages 501--509, 2019.

\bibitem{(DIFGSM)xie2019improving}
Cihang Xie, Zhishuai Zhang, Yuyin Zhou, Song Bai, Jianyu Wang, Zhou Ren, and
  Alan~L Yuille.
\newblock Improving transferability of adversarial examples with input
  diversity.
\newblock In {\em Proceedings of the IEEE/CVF Conference on Computer Vision and
  Pattern Recognition}, pages 2730--2739, 2019.

\bibitem{(resnext)xie2017aggregated}
Saining Xie, Ross Girshick, Piotr Doll{\'a}r, Zhuowen Tu, and Kaiming He.
\newblock Aggregated residual transformations for deep neural networks.
\newblock In {\em Proceedings of the IEEE conference on computer vision and
  pattern recognition}, pages 1492--1500, 2017.

\bibitem{(FS)xu2017feature}
Weilin Xu, David Evans, and Yanjun Qi.
\newblock Feature squeezing: Detecting adversarial examples in deep neural
  networks.
\newblock {\em arXiv preprint arXiv:1704.01155}, 2017.

\bibitem{(input_transform_bitred)xu2017feature}
Weilin Xu, David Evans, and Yanjun Qi.
\newblock Feature squeezing: Detecting adversarial examples in deep neural
  networks.
\newblock {\em arXiv preprint arXiv:1704.01155}, 2017.

\bibitem{(adp)yoon2021adversarial}
Jongmin Yoon, Sung~Ju Hwang, and Juho Lee.
\newblock Adversarial purification with score-based generative models.
\newblock {\em arXiv preprint arXiv:2106.06041}, 2021.

\bibitem{(WResNet)zagoruyko2016wide}
Sergey Zagoruyko and Nikos Komodakis.
\newblock Wide residual networks.
\newblock {\em arXiv preprint arXiv:1605.07146}, 2016.

\bibitem{(dncnn)zhang2017beyond}
Kai Zhang, Wangmeng Zuo, Yunjin Chen, Deyu Meng, and Lei Zhang.
\newblock Beyond a gaussian denoiser: Residual learning of deep cnn for image
  denoising.
\newblock {\em IEEE transactions on image processing}, 26(7):3142--3155, 2017.

\end{thebibliography}


\begin{thebibliography}{10}\itemsep=-1pt

\bibitem{(square)andriushchenko2020square}
Maksym Andriushchenko, Francesco Croce, Nicolas Flammarion, and Matthias Hein.
\newblock Square attack: a query-efficient black-box adversarial attack via
  random search.
\newblock In {\em European Conference on Computer Vision}, pages 484--501.
  Springer, 2020.

\bibitem{(obfuscated)athalye2018obfuscated}
Anish Athalye, Nicholas Carlini, and David Wagner.
\newblock Obfuscated gradients give a false sense of security: Circumventing
  defenses to adversarial examples.
\newblock In {\em International conference on machine learning}, pages
  274--283. PMLR, 2018.

\bibitem{(fbi)byun2021fbi}
Jaeseok Byun, Sungmin Cha, and Taesup Moon.
\newblock Fbi-denoiser: Fast blind image denoiser for poisson-gaussian noise.
\newblock {\em arXiv preprint arXiv:2105.10967}, 2021.

\bibitem{(evaluating)carlini2019evaluating}
Nicholas Carlini, Anish Athalye, Nicolas Papernot, Wieland Brendel, Jonas
  Rauber, Dimitris Tsipras, Ian Goodfellow, Aleksander Madry, and Alexey
  Kurakin.
\newblock On evaluating adversarial robustness.
\newblock {\em arXiv preprint arXiv:1902.06705}, 2019.

\bibitem{(C&W)carlini2017towards}
Nicholas Carlini and David Wagner.
\newblock Towards evaluating the robustness of neural networks.
\newblock In {\em 2017 ieee symposium on security and privacy (sp)}, pages
  39--57. IEEE, 2017.

\bibitem{(advertorch)ding2019advertorch}
Gavin~Weiguang Ding, Luyu Wang, and Xiaomeng Jin.
\newblock {AdverTorch} v0.1: An adversarial robustness toolbox based on
  pytorch.
\newblock {\em arXiv preprint arXiv:1902.07623}, 2019.

\bibitem{(benchmarking)dong2020benchmarking}
Yinpeng Dong, Qi-An Fu, Xiao Yang, Tianyu Pang, Hang Su, Zihao Xiao, and Jun
  Zhu.
\newblock Benchmarking adversarial robustness on image classification.
\newblock In {\em Proceedings of the IEEE/CVF Conference on Computer Vision and
  Pattern Recognition}, pages 321--331, 2020.

\bibitem{guo2017countering}
Chuan Guo, Mayank Rana, Moustapha Cisse, and Laurens Van Der~Maaten.
\newblock Countering adversarial images using input transformations.
\newblock {\em arXiv preprint arXiv:1711.00117}, 2017.

\bibitem{(resnet)he2016deep}
Kaiming He, Xiangyu Zhang, Shaoqing Ren, and Jian Sun.
\newblock Deep residual learning for image recognition.
\newblock In {\em Proceedings of the IEEE conference on computer vision and
  pattern recognition}, pages 770--778, 2016.

\bibitem{(advertorch)kim2020torchattacks}
Hoki Kim.
\newblock Torchattacks: A pytorch repository for adversarial attacks.
\newblock {\em arXiv preprint arXiv:2010.01950}, 2020.

\bibitem{(ensemble_attack)liu2016delving}
Yanpei Liu, Xinyun Chen, Chang Liu, and Dawn Song.
\newblock Delving into transferable adversarial examples and black-box attacks.
\newblock {\em arXiv preprint arXiv:1611.02770}, 2016.

\bibitem{(PGD)madry2017towards}
Aleksander Madry, Aleksandar Makelov, Ludwig Schmidt, Dimitris Tsipras, and
  Adrian Vladu.
\newblock Towards deep learning models resistant to adversarial attacks.
\newblock {\em arXiv preprint arXiv:1706.06083}, 2017.

\bibitem{(NRP)naseer2020self}
Muzammal Naseer, Salman Khan, Munawar Hayat, Fahad~Shahbaz Khan, and Fatih
  Porikli.
\newblock A self-supervised approach for adversarial robustness.
\newblock In {\em Proceedings of the IEEE/CVF Conference on Computer Vision and
  Pattern Recognition}, pages 262--271, 2020.

\bibitem{(regnet)radosavovic2020designing}
Ilija Radosavovic, Raj~Prateek Kosaraju, Ross Girshick, Kaiming He, and Piotr
  Doll{\'a}r.
\newblock Designing network design spaces.
\newblock In {\em Proceedings of the IEEE/CVF Conference on Computer Vision and
  Pattern Recognition}, pages 10428--10436, 2020.

\bibitem{(tvm)rudin1992nonlinear}
Leonid~I Rudin, Stanley Osher, and Emad Fatemi.
\newblock Nonlinear total variation based noise removal algorithms.
\newblock {\em Physica D: nonlinear phenomena}, 60(1-4):259--268, 1992.

\bibitem{(adaptive_attack)tramer2020adaptive}
Florian Tramer, Nicholas Carlini, Wieland Brendel, and Aleksander Madry.
\newblock On adaptive attacks to adversarial example defenses.
\newblock {\em arXiv preprint arXiv:2002.08347}, 2020.

\bibitem{(resnext)xie2017aggregated}
Saining Xie, Ross Girshick, Piotr Doll{\'a}r, Zhuowen Tu, and Kaiming He.
\newblock Aggregated residual transformations for deep neural networks.
\newblock In {\em Proceedings of the IEEE conference on computer vision and
  pattern recognition}, pages 1492--1500, 2017.

\bibitem{(WResNet)zagoruyko2016wide}
Sergey Zagoruyko and Nikos Komodakis.
\newblock Wide residual networks.
\newblock {\em arXiv preprint arXiv:1605.07146}, 2016.

\end{thebibliography}
}

\end{document}


\title{Supplementary Materials for \\ NCIS: Neural Contextual Iterative Smoothing for \\ Purifying Adversarial Perturbations}

\author{Sungmin Cha\textsuperscript{\rm 1}, Naeun Ko\textsuperscript{\rm 2}, Youngjoon Yoo\textsuperscript{\rm 2, \rm 3}, and Taesup Moon\textsuperscript{\rm 1}\thanks{Corresponding author (E-mail: \texttt{tsmoon@snu.ac.kr})} \\\

\textsuperscript{\rm 1}Department of Electrical and Computer Engineering, Seoul National University\\
\textsuperscript{\rm 2}Face, NAVER Clova 
\textsuperscript{\rm 3}NAVER AI Lab\\
{\tt\small sungmin.cha@snu.ac.kr, \tt\small naeun.ko@navercorp.com, \tt\small youngjoon.yoo@navercorp.com, \tt\small tsmoon@snu.ac.kr}
}
\maketitle

\section{Additional Analysis for Adversarial Noise}

\begin{figure*}[h]
\centering 
\subfigure[Distribution of mean]
{\includegraphics[width=0.31\linewidth]{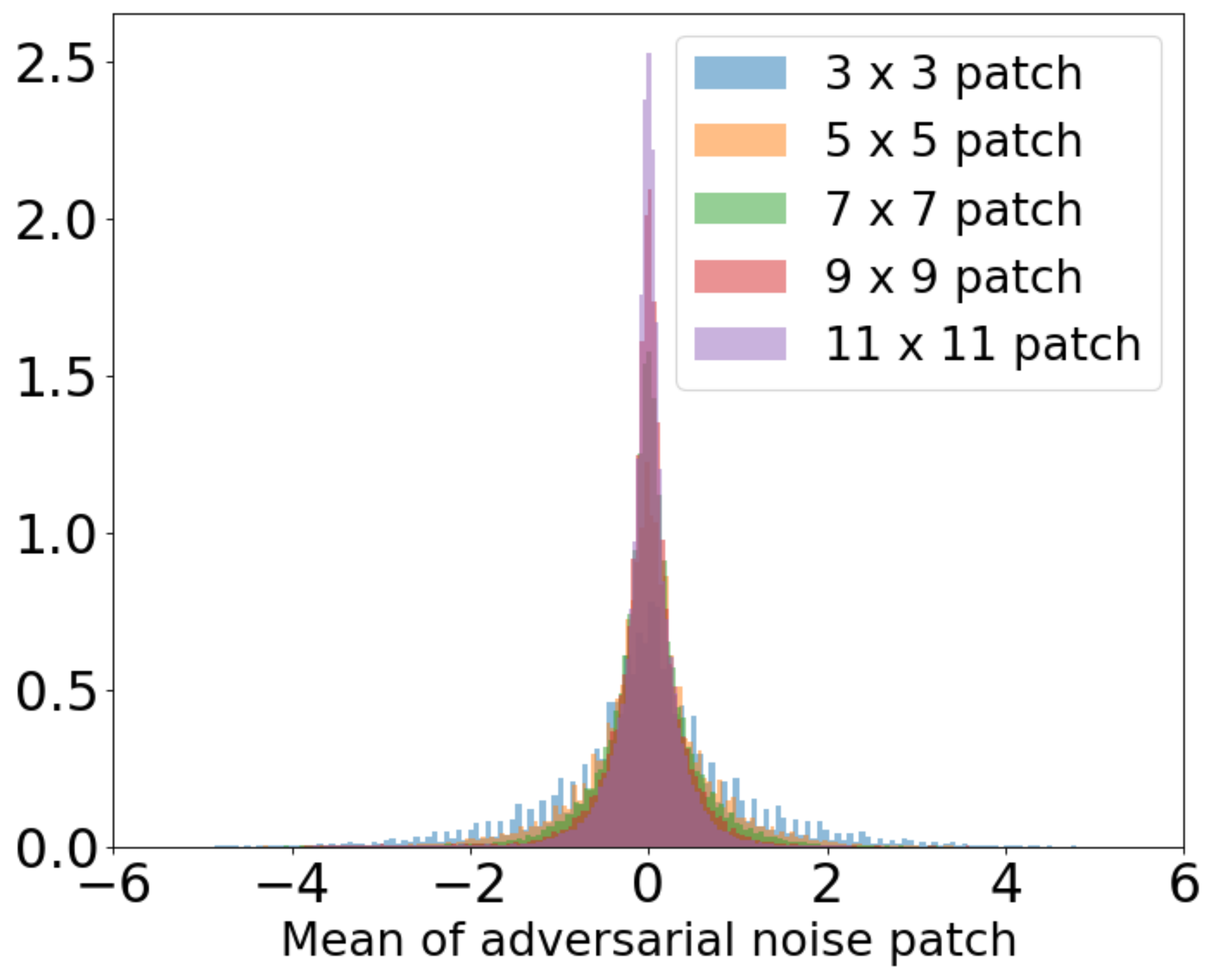}
\label{figure:motivation_distribution_pgd_targeted}}
\subfigure[Skewness]
{\includegraphics[width=0.305\linewidth]{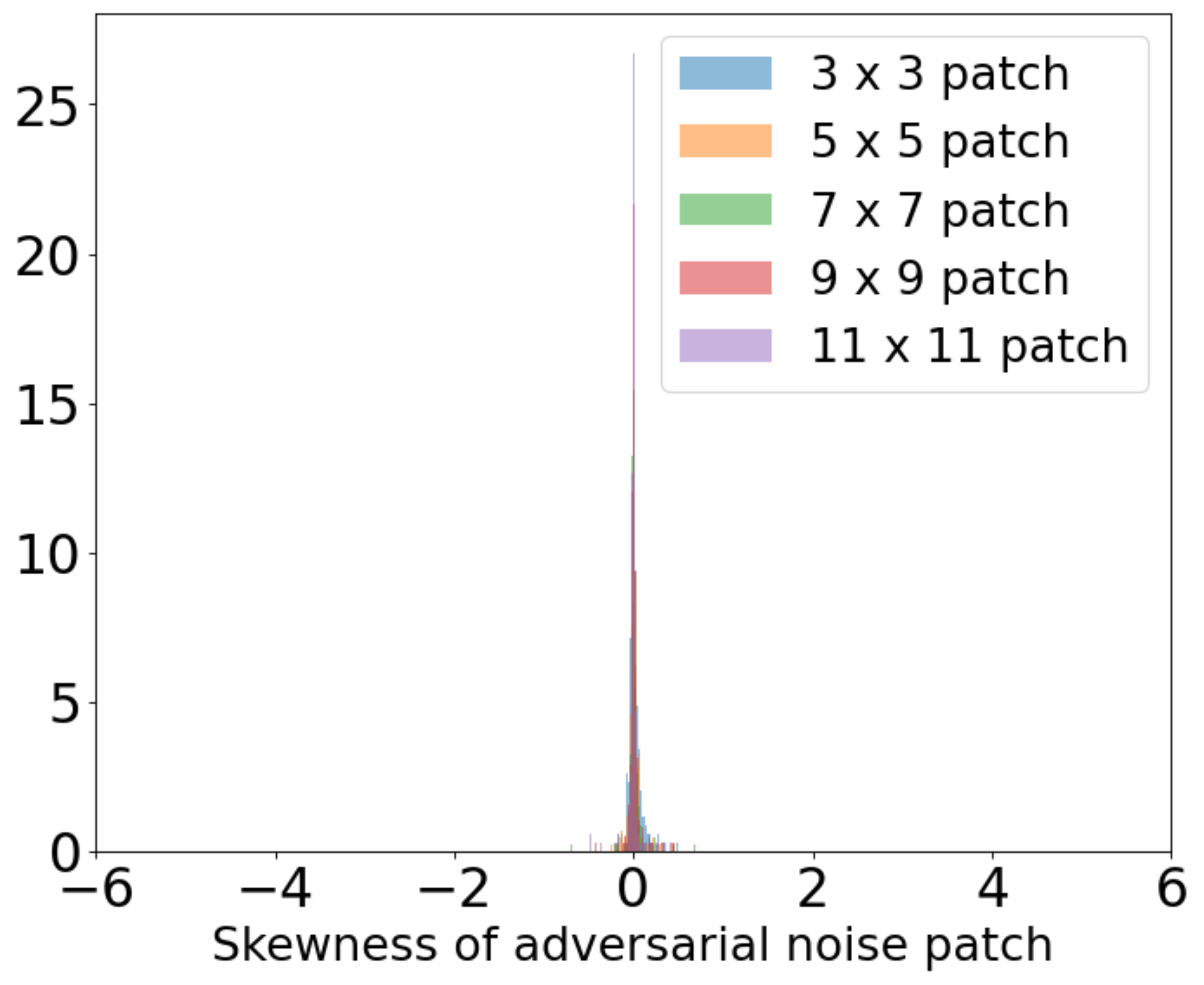}
\label{figure:motivation_skewnessn_pgd_targeted}}
\vspace{-.15in}
\caption{Experimental analysis for adversarial examples generated from targeted $L_{\infty}$ PGD \cite{(PGD)madry2017towards} ($\alpha$ = 1.6 / 255, where $\alpha$ is a step size) attack with 10 attack iterations.}\label{figure:motivation_pgd_targeted}
 \end{figure*}
 
 \begin{figure*}[h]
\centering 
\subfigure[Distribution of mean]
{\includegraphics[width=0.315\linewidth]{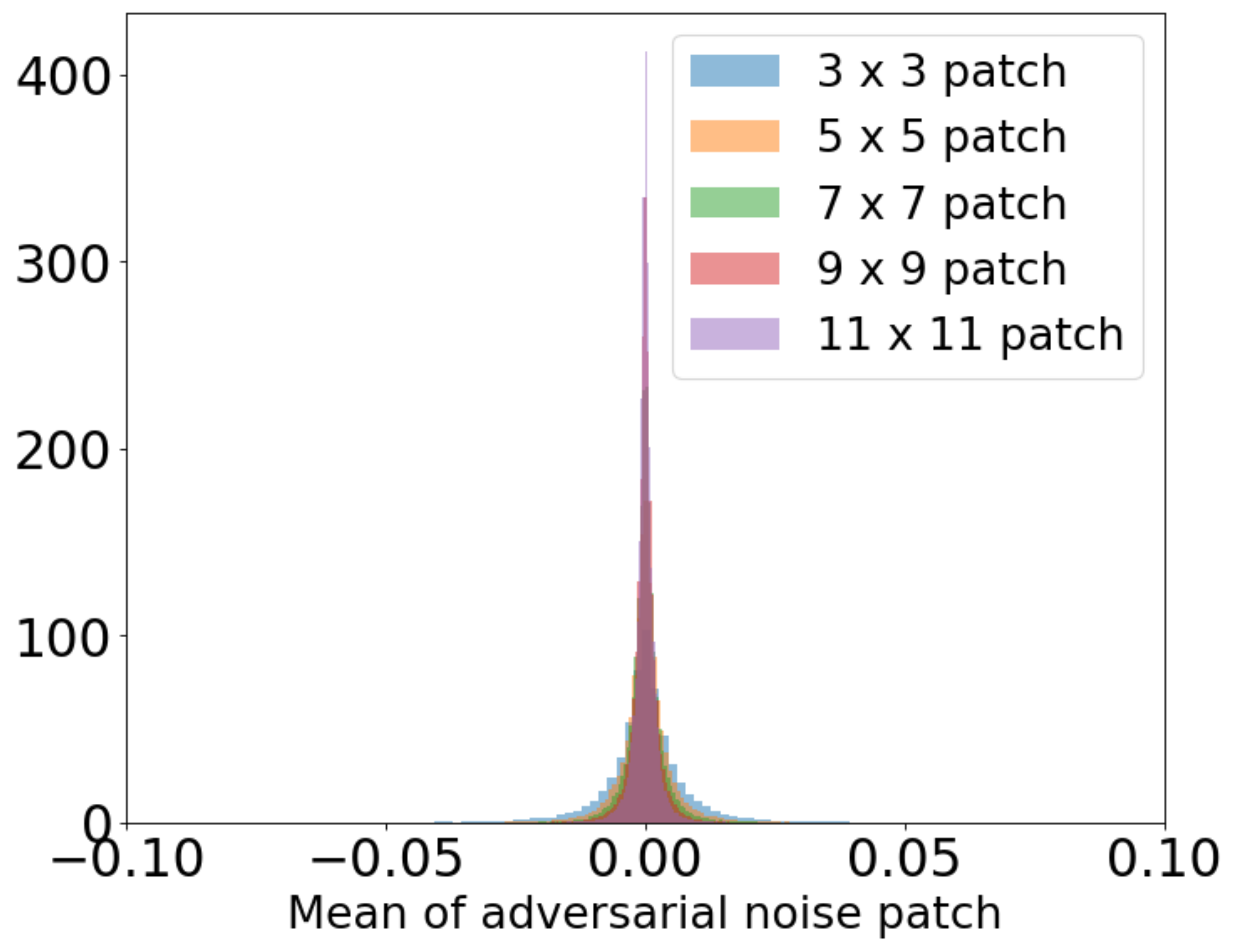}
\label{figure:motivation_distribution_pgd_l2}}
\subfigure[Skewness]
{\includegraphics[width=0.30\linewidth]{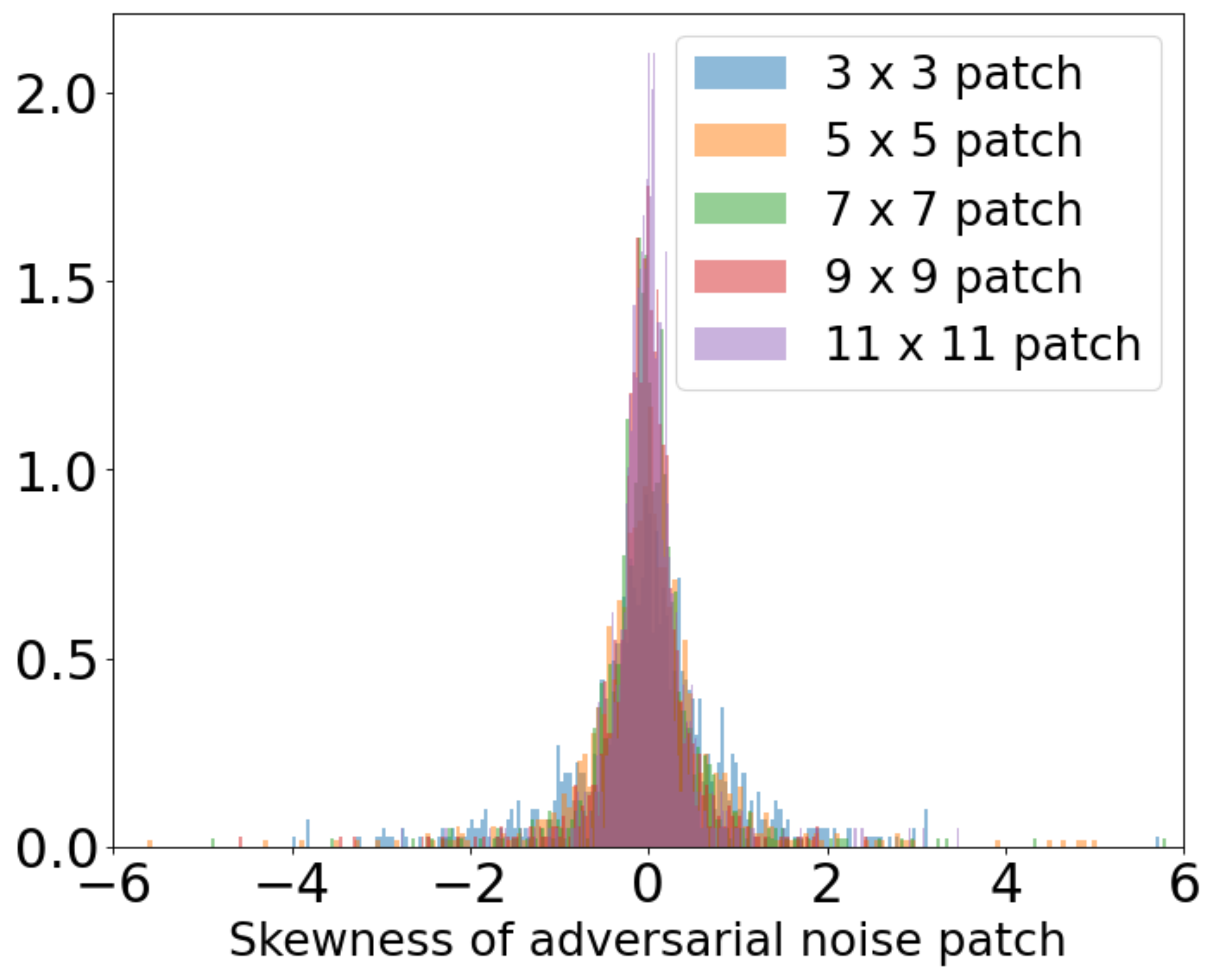}
\label{figure:motivation_skewnessn_pgd_l2}}
\vspace{-.15in}
\caption{Experimental analysis for adversarial examples generated from targeted $L_2$ PGD \cite{(PGD)madry2017towards} ($\alpha$ = 1 / 255, where $\alpha$ is a step size) attack with 10 attack iterations.}\label{figure:motivation_pgd_l2}
 \end{figure*}
 
  \begin{figure*}[h]
\centering 
\subfigure[Distribution of mean]
{\includegraphics[width=0.30\linewidth]{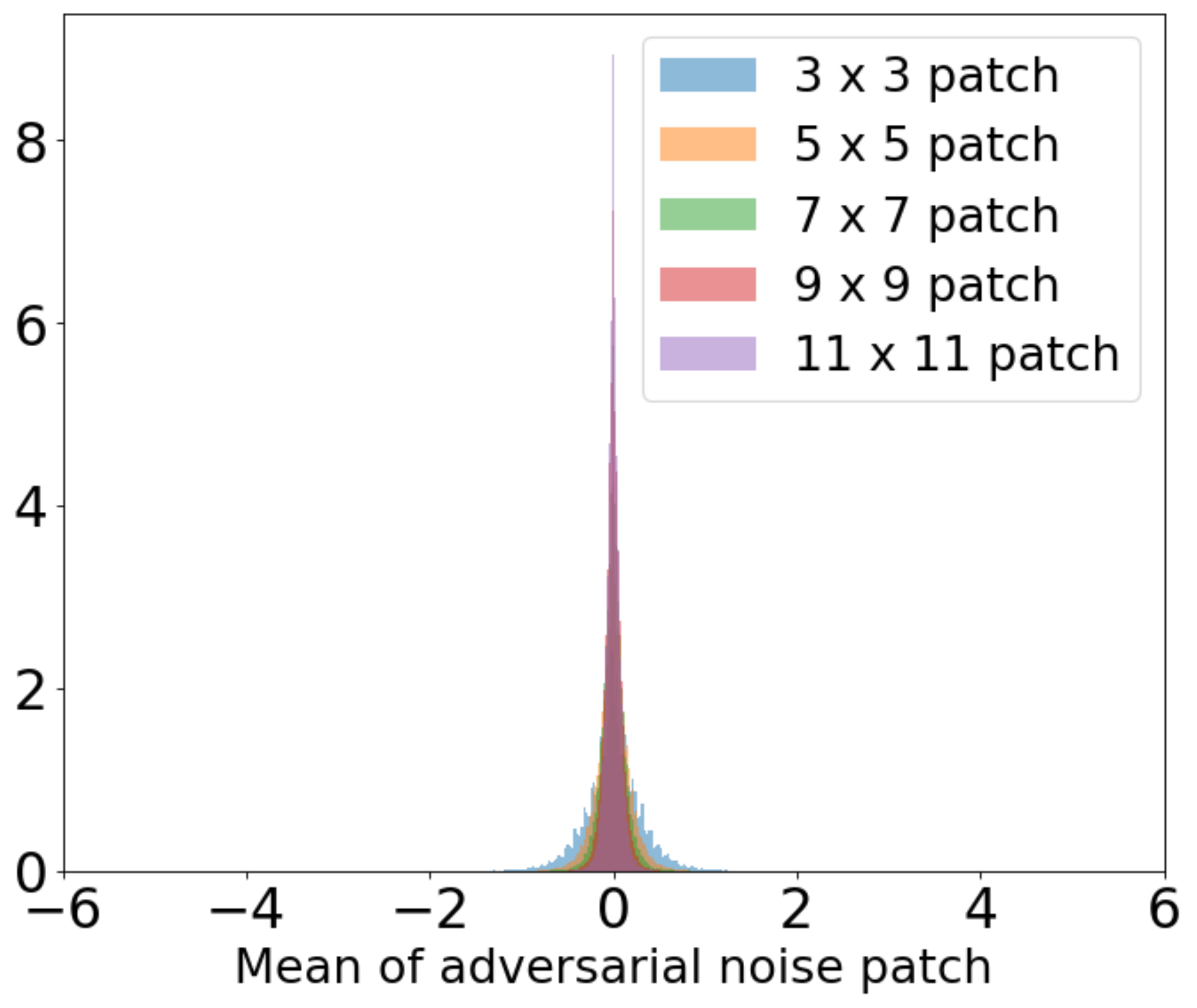}
\label{figure:motivation_distribution_cw_l2}}
\subfigure[Skewness]
{\includegraphics[width=0.31\linewidth]{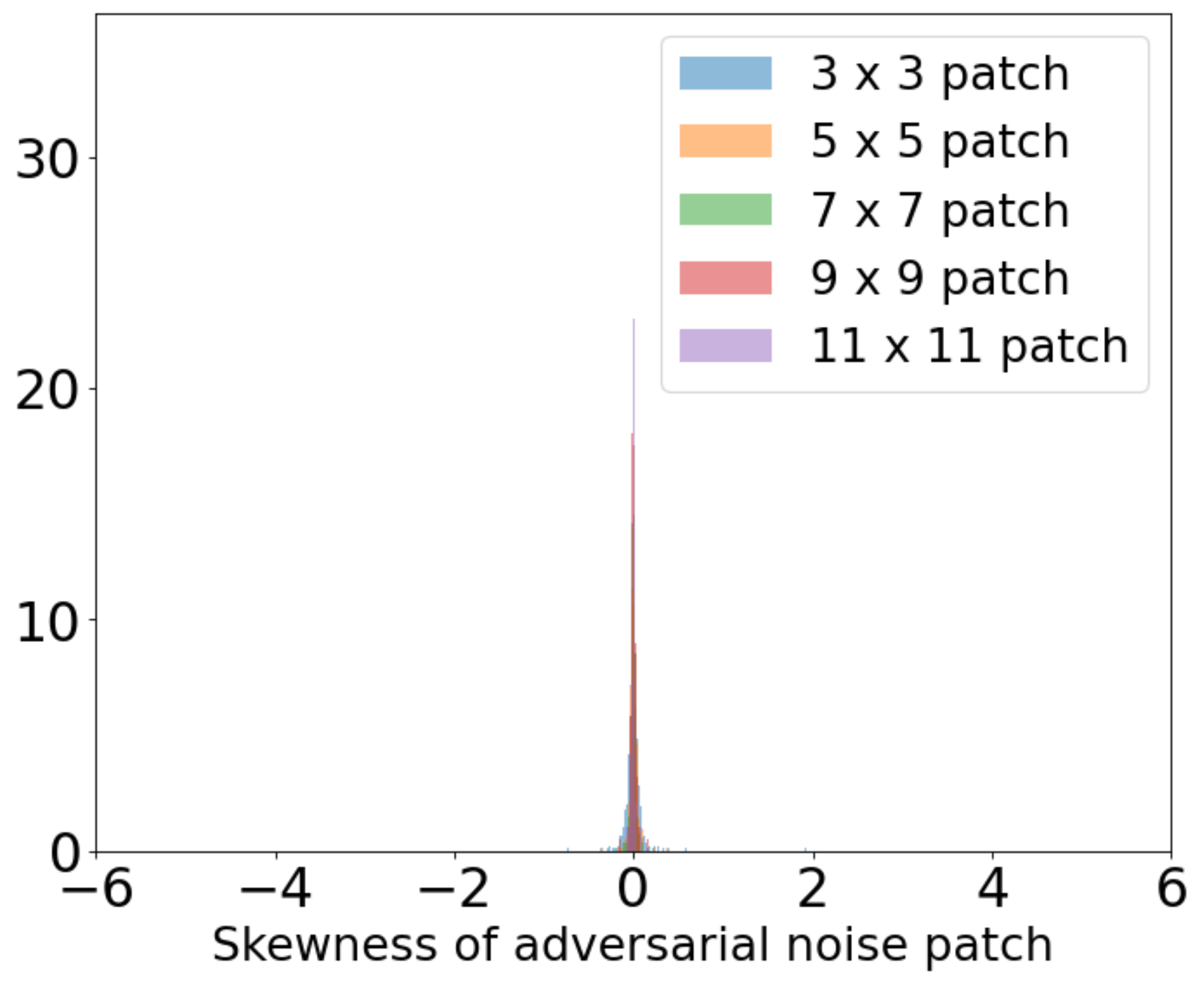}
\label{figure:motivation_skewnessn_cw_l2}}
\vspace{-.15in}
\caption{Experimental analysis for adversarial examples generated from targeted $L_2$ CW \cite{(C&W)carlini2017towards} attack with 10 attack iterations. All other hyperparameters set to default hyperparameters of \cite{(advertorch)kim2020torchattacks}}\label{figure:motivation_cw}
 \end{figure*}

Figure \ref{figure:motivation_pgd_targeted}, \ref{figure:motivation_pgd_l2}, and \ref{figure:motivation_cw} show additional experimental results of adversarial noise.
We conducted the experiments on the same dataset used in Section 3.2 of the manuscript but with different attack methods. 
Figure \ref{figure:motivation_pgd_targeted} is the result of targeted $L_\infty$ PGD \cite{(PGD)madry2017towards} attack with the same $\epsilon$, $\alpha$ and attack iterations proposed in Section 3.2 of the manuscript. 
Figure \ref{figure:motivation_pgd_l2} shows analysis of untargeted $L_2$ PGD attack with $\epsilon = \{1,2,3,4,5\}$ and Figure \ref{figure:motivation_cw} is about $L_2$ CW~\cite{(C&W)carlini2017towards} attack with 10 attack iterations.
From all the experimental results, we observe that the patches of adversarial noises generated from untargeted / targeted and $L_2$/$L_\infty$ optimization-based adversarial attacks consistently show more or less zero mean and have symmetric distribution.
We believe that these results support how our proposed methods could achieve such strong purification results against various types of attacks, as shown in Table 1 of the manuscript.


\section{Architectural Details and Experiments for NCIS}
\subsection{Architectural Details on FBI-Net}

We implemented FBI-Net by slightly modifying FBI-Denoiser \cite{(fbi)byun2021fbi}'s official code. In the original paper, they composed FBI-Net with 17 layers, 64 convolutional filters for each layer. For our method, we changed the number of layers to 8 for all experiments including FBI-Net, FBI-E and NCIS.

\subsection{The number of training data for training NCIS}

For self-supervised training of NCIS, we randomly selected and used only 5\% images of the ImageNet training dataset since there was no significant difference even when more images were used, as shown in Table \ref{table:fbi_dataset_ratio}.

\begin{table}[h]
\caption{Experimental results of NCIS ($i = 7$, $K = 11$, $m = 2$) trained by different number of images. For all experiments, we used ResNet-152 as the classification model and evaluate each case with the ImageNet validation dataset. For generating adversarial examples, we attacked each image using untargeted $L_\infty$ PGD ($\epsilon = 16/255, \alpha = 1.6/255$) attack with 10 attack iterations. We only experimented with a single seed.}
\centering
\smallskip\noindent
\resizebox{.5\linewidth}{!}{
\begin{tabular}{|c||c|c|}
\hline
 ResNet-152           & Standard Accuracy & Robust Accuracy \\ \hline \hline
\textbf{NCIS (5\%)}  & \textbf{69.07}             & \textbf{46.49}           \\ \hline
NCIS (10\%) & 68.92             & 49.14           \\ \hline
NCIS (15\%) & 68.84             & 46.93           \\ \hline
NCIS (20\%) & 68.93             & 47.84           \\ \hline
NCIS (30\%) & 68.99             & 46.84           \\ \hline
\end{tabular}
}
\label{table:fbi_dataset_ratio}
\end{table}
\newpage

\subsection{Selecting the number of iterations $i$ for NCIS ($K = 11, m = 2$)}

\begin{figure*}[h]
\centering 
\subfigure[Standard accuracy]
{\includegraphics[width=0.32\linewidth]{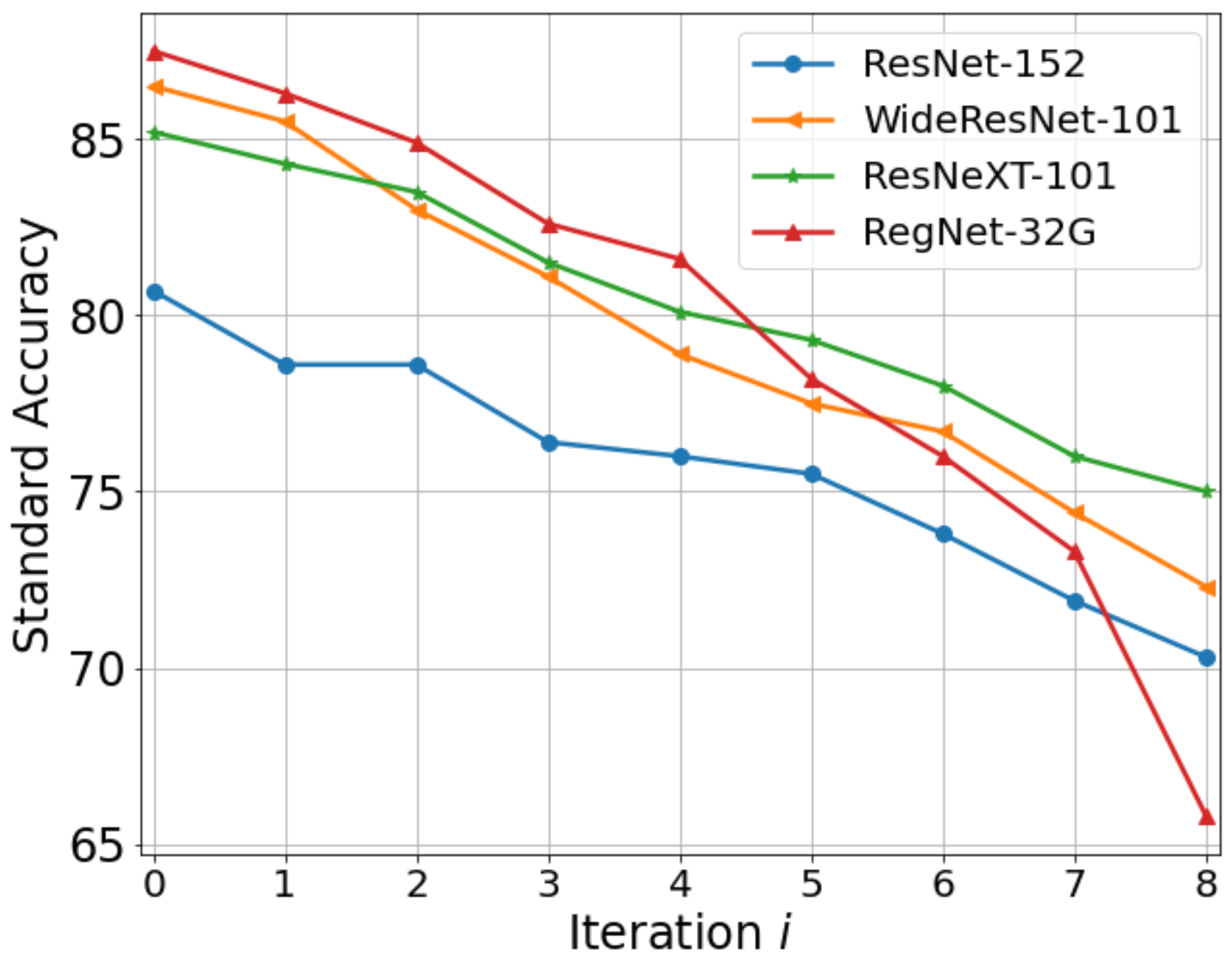}}
\subfigure[Robust accuracy]
{\includegraphics[width=0.32\linewidth]{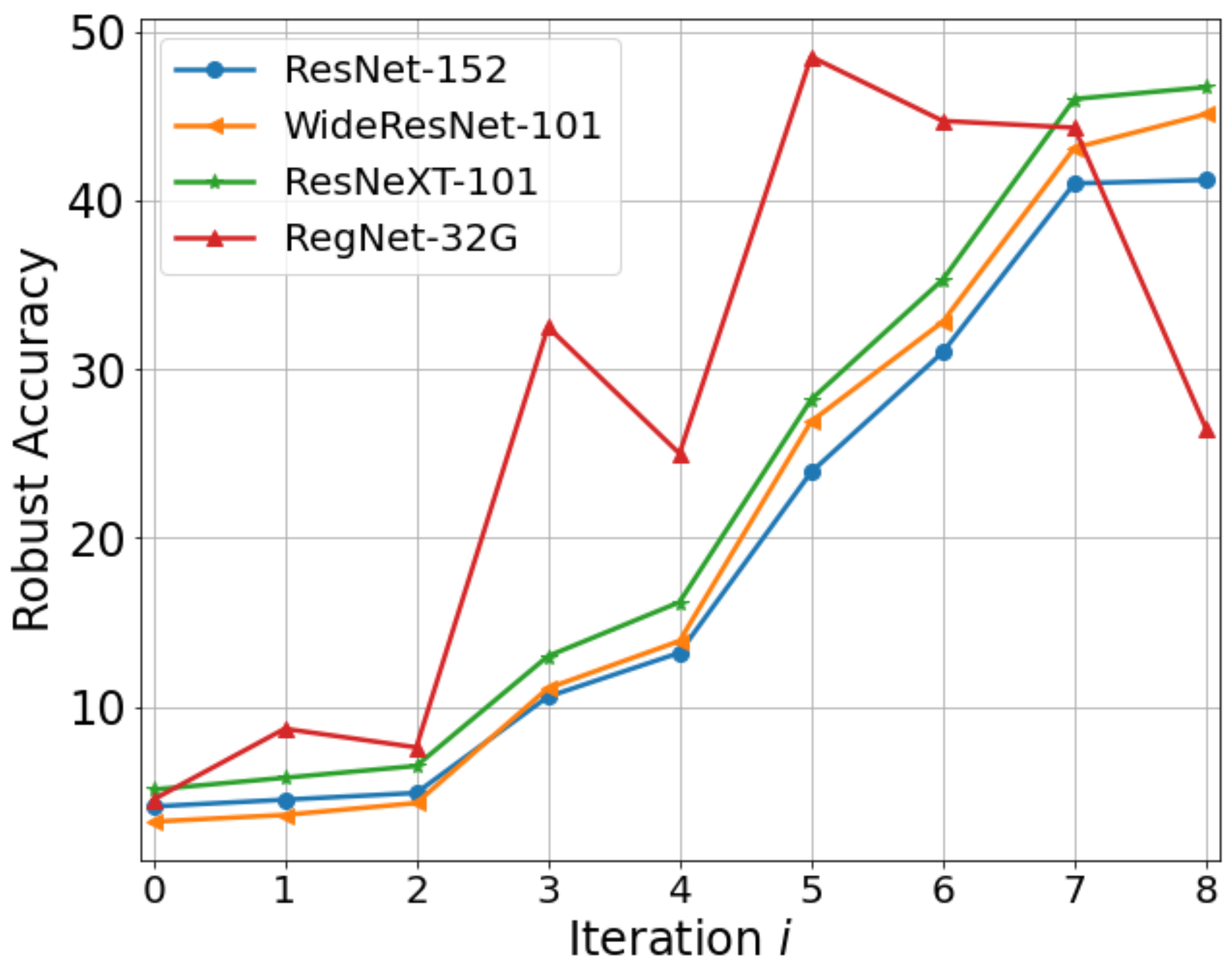}}
\subfigure[Average of standard and robust accuracy]
{\includegraphics[width=0.32\linewidth]{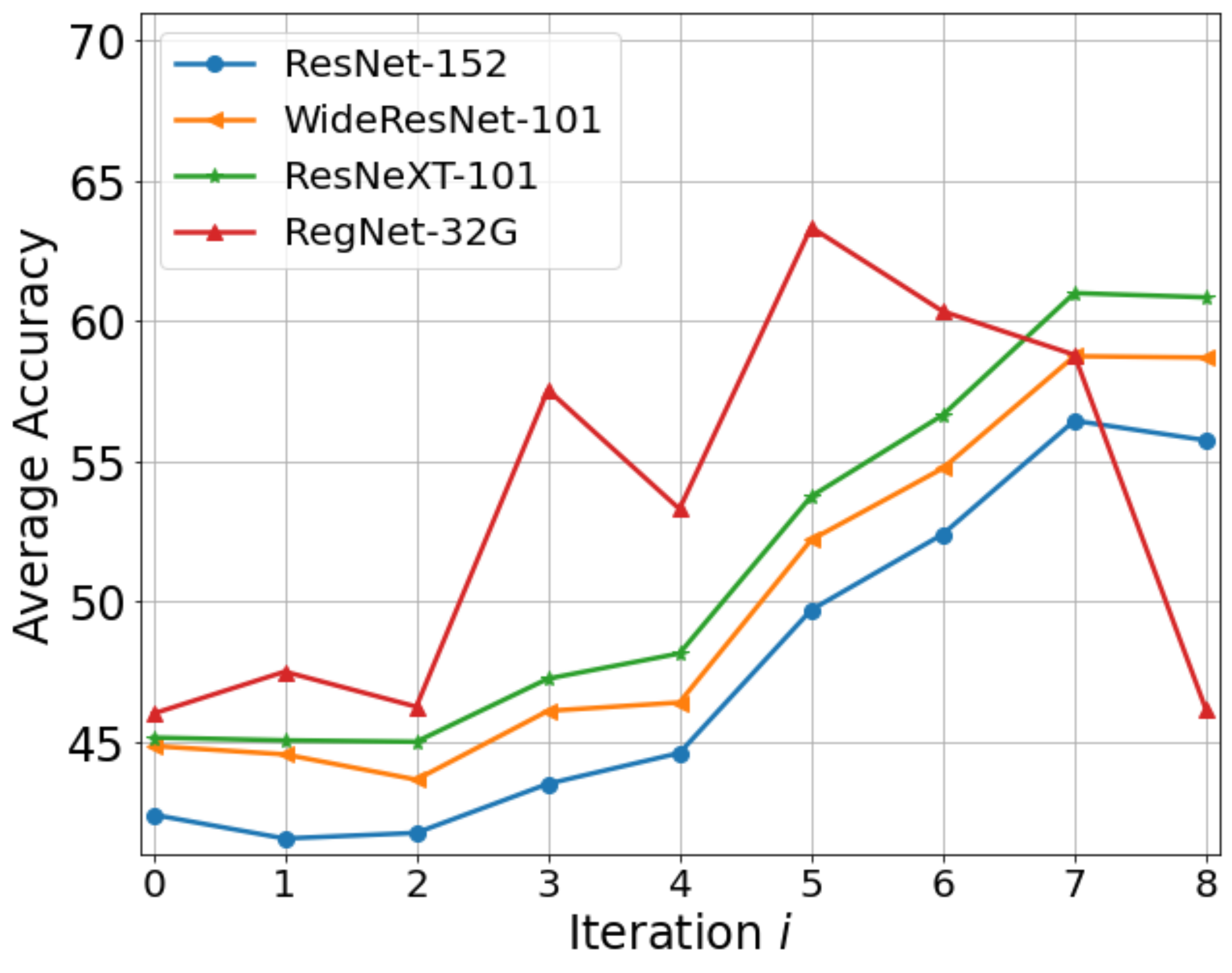}}
\caption{Experimental results of selecting the number of iterations $i$ of NCIS for each classification model. We used randomly sampled 1,000 images from ImageNet training dataset and adversarial examples generated by $L_\infty$ PGD ($\epsilon = 16 / 255$, $\alpha = 1.6/255$) attack with 10 attack iterations.}
\label{figure:selecting_iteration}
 \end{figure*}

Also, we conducted experiments, as in Figure \ref{figure:selecting_iteration}, to select $i$ (number of iterations for iterative smoothing) of NCIS for each classification model.
Considering average of standard and robust accuracy, \textbf{$i = 7$} is the best iteration number for ResNet-152~\cite{(resnet)he2016deep}, WideResNet-101~\cite{(WResNet)zagoruyko2016wide} and ResNeXT-101~\cite{(resnext)xie2017aggregated} and \textbf{$i = 5$} is best for RegNet-32G~\cite{(regnet)radosavovic2020designing}.
Note that, for all experiments, the selected $i$ for each classification model was used fixedly.




\subsection{Finding the best configuration of NCIS}

\begin{figure*}[h]
\centering 
\subfigure[Standard accuracy]
{\includegraphics[width=0.32\linewidth]{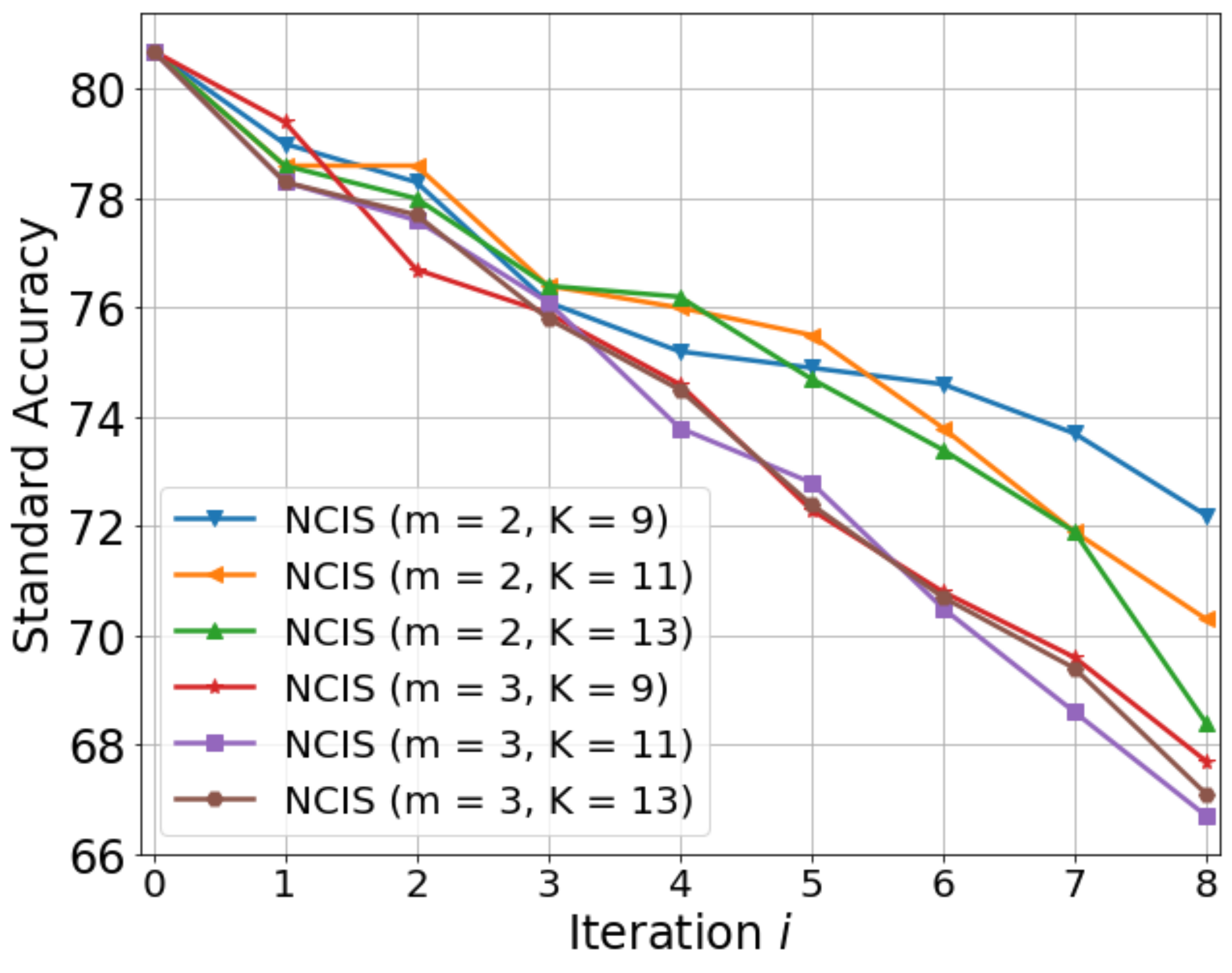}}
\subfigure[Robust accuracy]
{\includegraphics[width=0.32\linewidth]{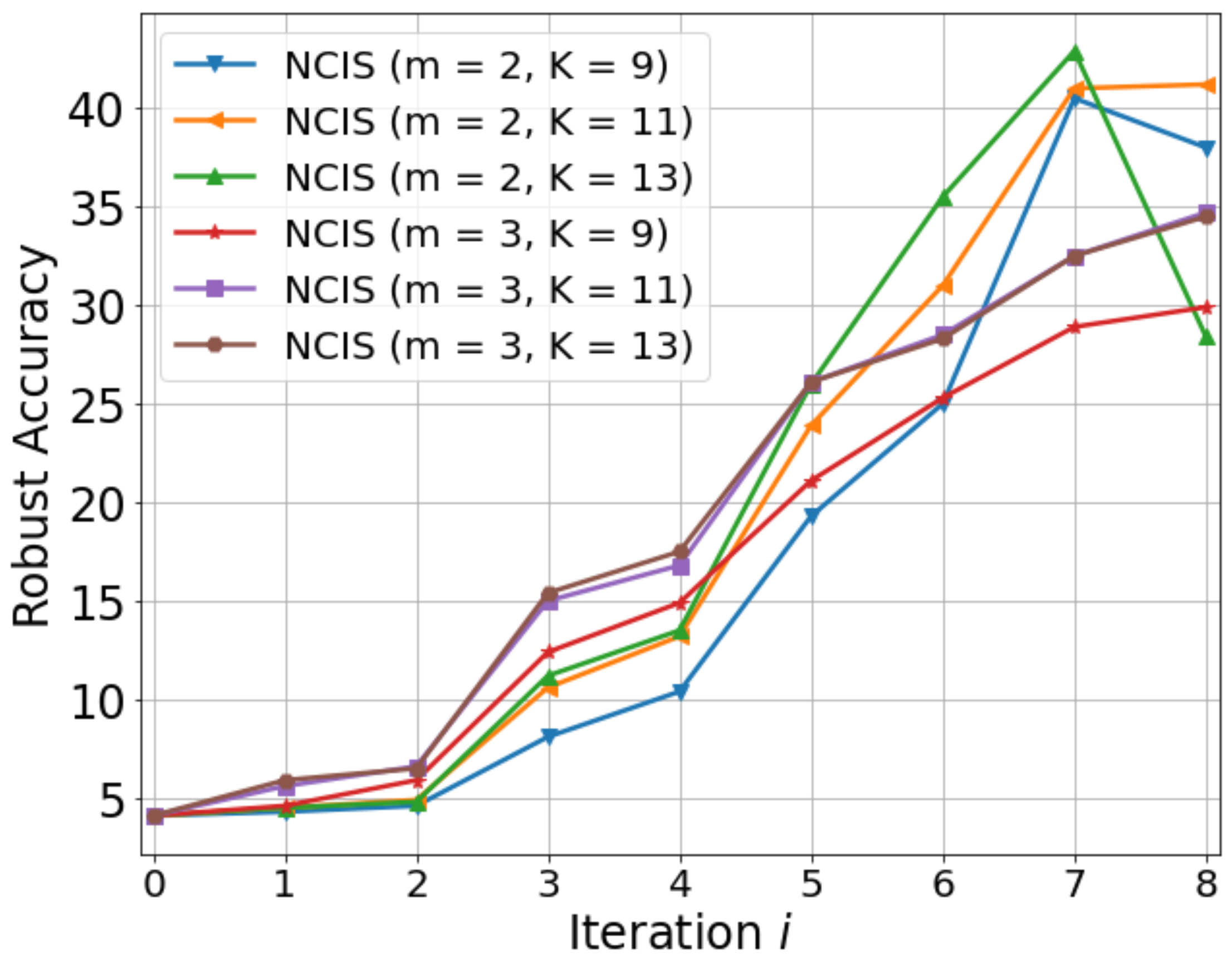}}
\subfigure[Average of standard and robust accuracy]
{\includegraphics[width=0.32\linewidth]{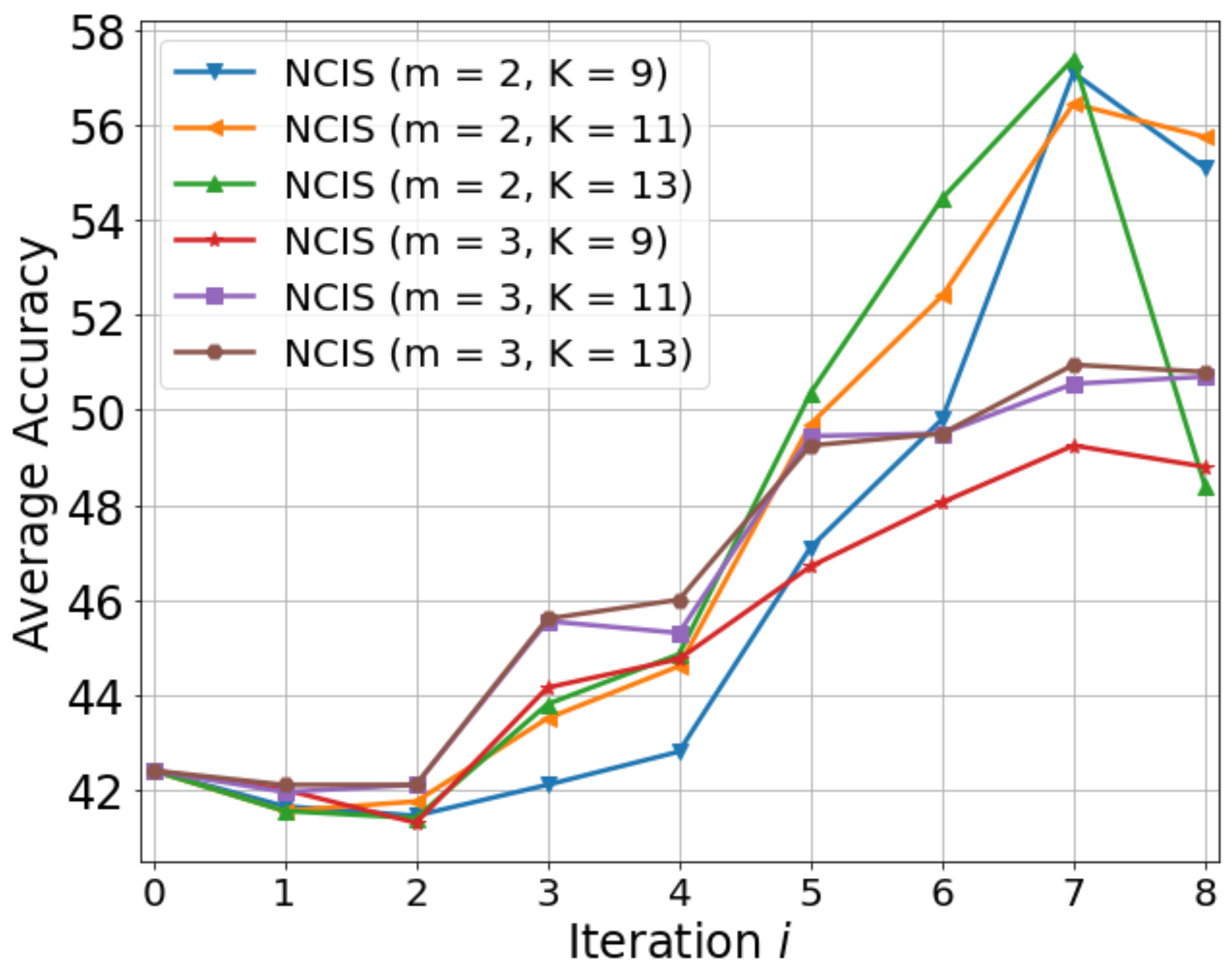}}
\caption{Experimental results of variants of NCIS. Experiments are conducted with ImageNet pretrained ResNet-152. We randomly sampled 1,000 images from ImageNet training dataset and generated adversarial examples using $L_\infty$ PGD ($\epsilon = 16 / 255$, $\alpha = 1.6/255$) attack with 10 attack iterations.}
\label{figure:selecting_ncis_model}
 \end{figure*}

Figure \ref{figure:selecting_ncis_model} shows the experimental results of various types of NCIS.
Note that all NCIS are trained with 5\% images of ImageNet training dataset as already proposed in the previous section.
First, both robust and standard accuracy of NCIS with $m = 3$ is clearly lower than NCIS with $m = 2$ because reconstruction difficulty increases as the reshape size of FBI-E becomes bigger.
Second, among the results of NCIS with $m = 2$ at $i = 7$, NCIS ($m = 2$, $K = 13$) achieves slightly better performance compared to NCIS ($m = 2$, $K = 11$).
However, robust accuracy of both NCIS ($m = 2$, $K = 13$) and NCIS ($m = 2$, $K = 9$) significantly decrease at $i = 8$ where NCIS ($m = 2$, $K = 11$) does not.
In this regard, we are concerned that NCIS ($m = 2$, $K = 13$) and NCIS ($m = 2$, $K = 9$) might be sensitive to the number of iterations even though they are slightly ahead in performance.
Therefore, we selected NCIS ($m = 2$, $K = 11$) as the representative model of NCIS and conducted all experiments with it.

 \newpage

\section{Additional Experimental Results}
\subsection{Comparison of inference time, GPU memory, and the number of parameters}

Table \ref{table:time_memory_parameters} shows the comparison of inference time, GPU memory requirement, and the number of parameters for purifying a single image. GPU memory was measured on an image of size 224x224.
First, we can check that traditional input transformation-based methods consistently show fast inference time, except for TVM~\cite{(tvm)rudin1992nonlinear}, with no GPU memory requirement. However, as already proposed in the manuscript and \cite{guo2017countering}, these methods are easily broken by strong white-box attacks.
Second, the original NRP has a large number of parameters and requires a huge GPU memory for purifying a single image. We think this is a fatal weakness from a practical point of view. To overcome this limitation, the author of NRP newly proposed a lightweight version of NRP, denoted as NRP (resG), at their official code. NRP (resG) significantly reduces inference time, GPU memory requirement, and the number of parameters. However, both NRP and NRP (resG) have the generalization issue and cannot purify several types of adversarial examples well, as proposed in the manuscript. In addition, NCIS is slower than NRP (resG), but shows faster inference time and memory requirements than NRP.
Note that the number of parameters of NCIS is significantly lower than NRP variants.
GS is the most computationally efficient compared to other baselines.  
Even though our NCIS is slow and has high computational cost than GS and NRP (resG), we would like to emphasize that NCIS generally achieved superior results against various attacks than GS, and also much better results than both NRP and NRP (resG) when considering both standard and robust accuracy, as already shown in the manuscript.

\begin{table}[h]
\caption{Comparison of computational efficiency.}
\centering
\smallskip\noindent
\resizebox{.8\linewidth}{!}{
\begin{tabular}{|c||c|c|c|c||c|c||c|c|}
\hline
                                                         & JPEG   & FS     & TVM    & SR     & \begin{tabular}[c]{@{}c@{}}NRP\\ (resG)\end{tabular} & NRP    & \begin{tabular}[c]{@{}c@{}}\textbf{GS} \\($i = 7$)\end{tabular}     & \begin{tabular}[c]{@{}c@{}}\textbf{NCIS}\\ ($i = 7$)\end{tabular}   \\ \hline \hline
\begin{tabular}[c]{@{}c@{}}Inference\\ Time\end{tabular} & 0.0070 & 0.0007 & 1.1259 & 0.0084 & 0.0007                                               & 0.0892 & \textbf{0.0004} & \textbf{0.0779} \\ \hline
\begin{tabular}[c]{@{}c@{}}GPU\\ Memory\end{tabular}     & -      & -      & -      & -      & 0.43G                                               & 9.86G  & \textbf{0.002G} & \textbf{0.60G} \\ \hline
\# of Parameters                                         & -      & -      & -      & -      & 1.70M                                                & 16.6M  & \textbf{-}      & \textbf{0.40M}  \\ \hline
\end{tabular}
}
\label{table:time_memory_parameters}
\end{table}

\newpage

\subsection{White-box PGD attack on other classification models}

\begin{figure*}[h]
\centering 
\subfigure[Various $\epsilon$ (WideResNet-101~\cite{(WResNet)zagoruyko2016wide})]
{\includegraphics[width=0.32\linewidth]{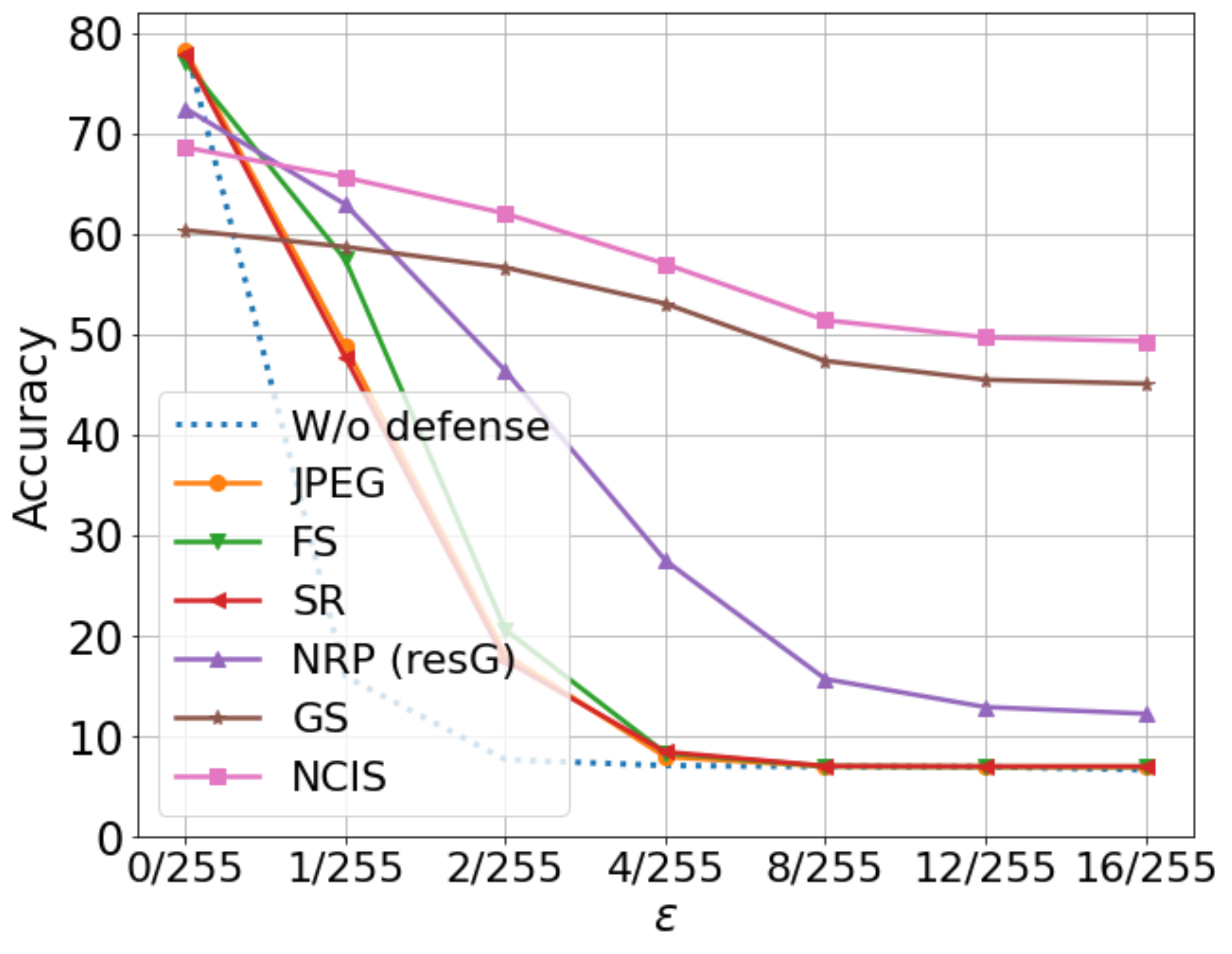}}
\subfigure[Various $\epsilon$ (ResNeXT-101~\cite{(resnext)xie2017aggregated})]
{\includegraphics[width=0.32\linewidth]{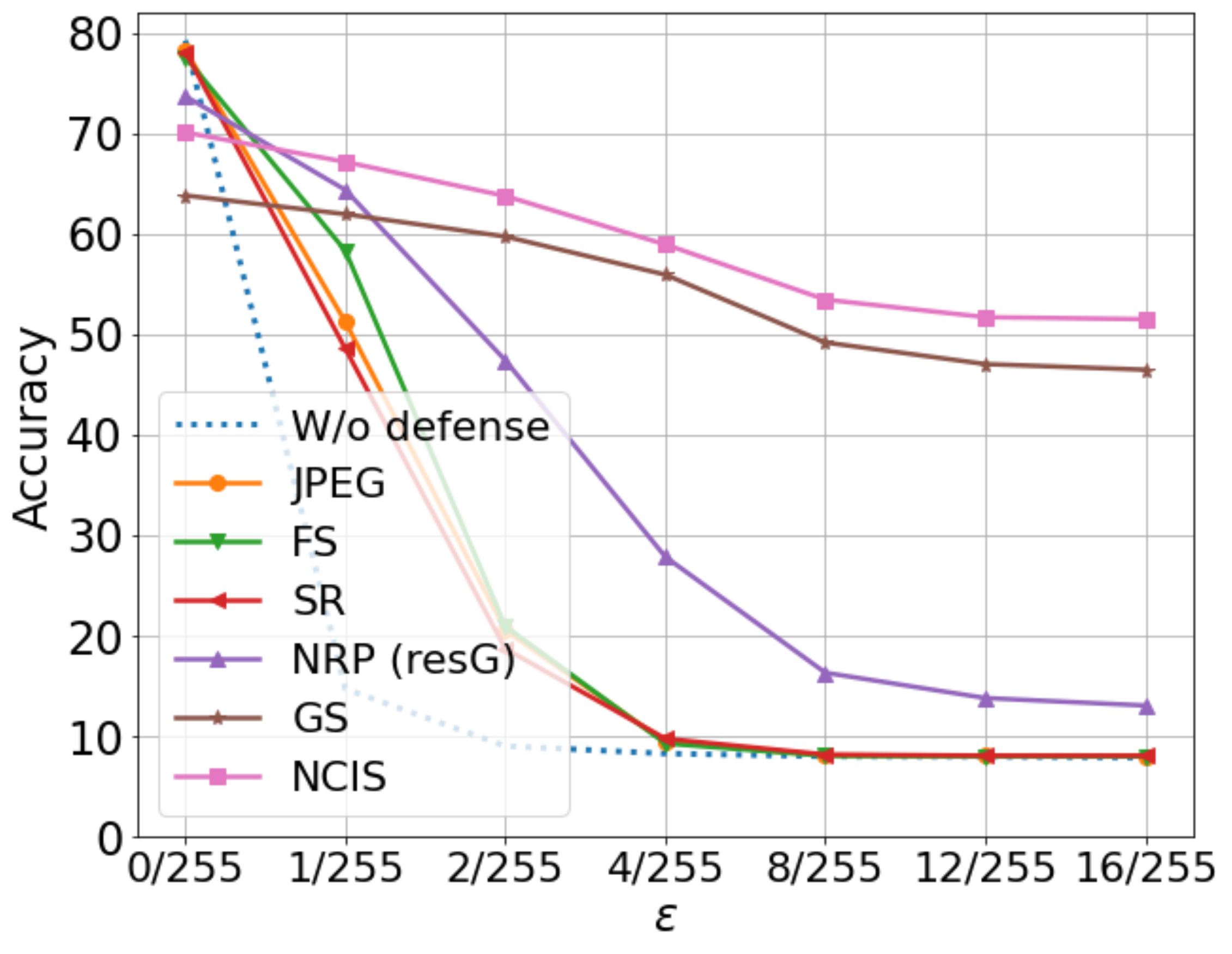}}
\subfigure[Various $\epsilon$ (RegNet-32G~\cite{(regnet)radosavovic2020designing})]
{\includegraphics[width=0.32\linewidth]{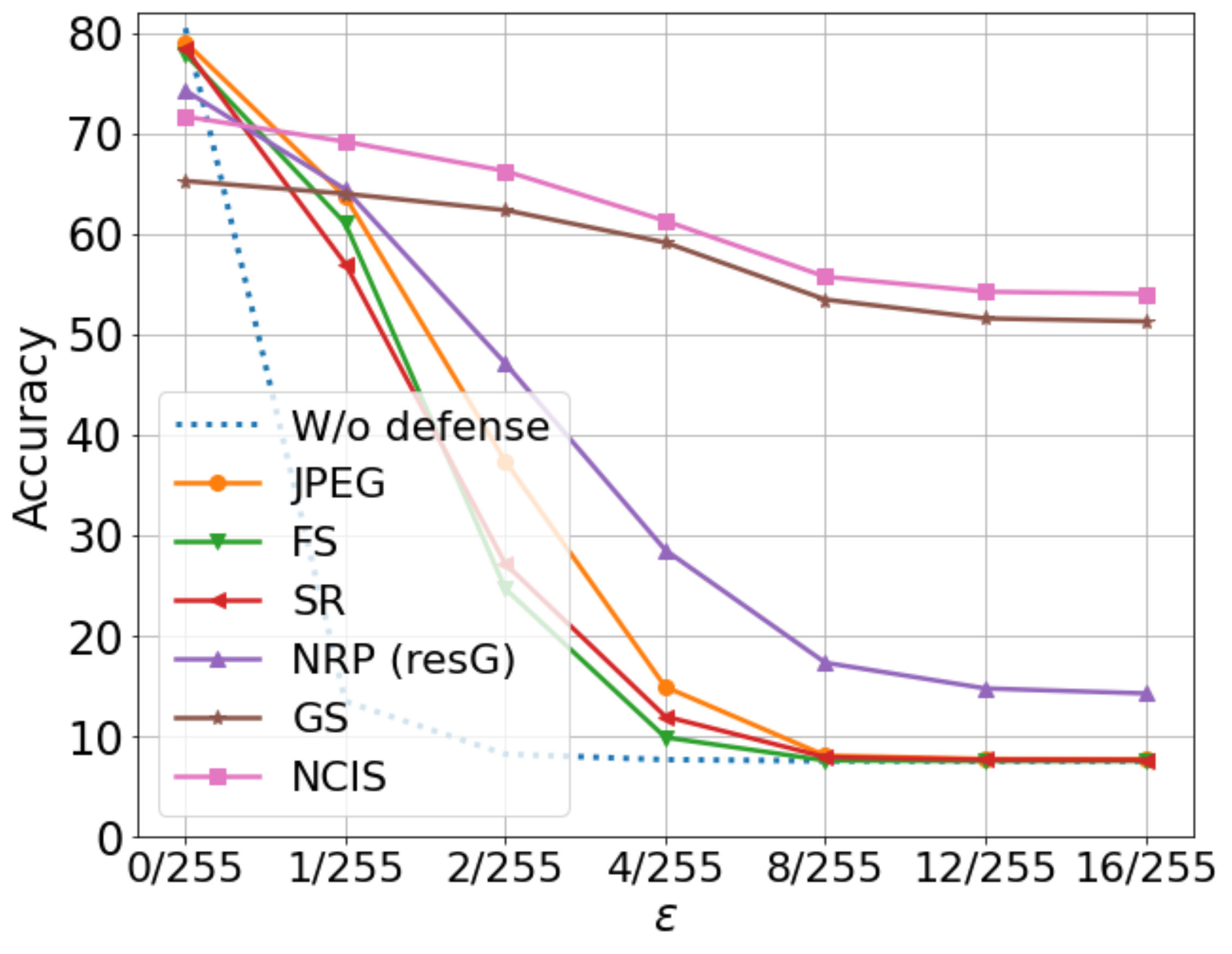}}
\subfigure[Various attack iters (WideResNet-101~\cite{(WResNet)zagoruyko2016wide})]
{\includegraphics[width=0.32\linewidth]{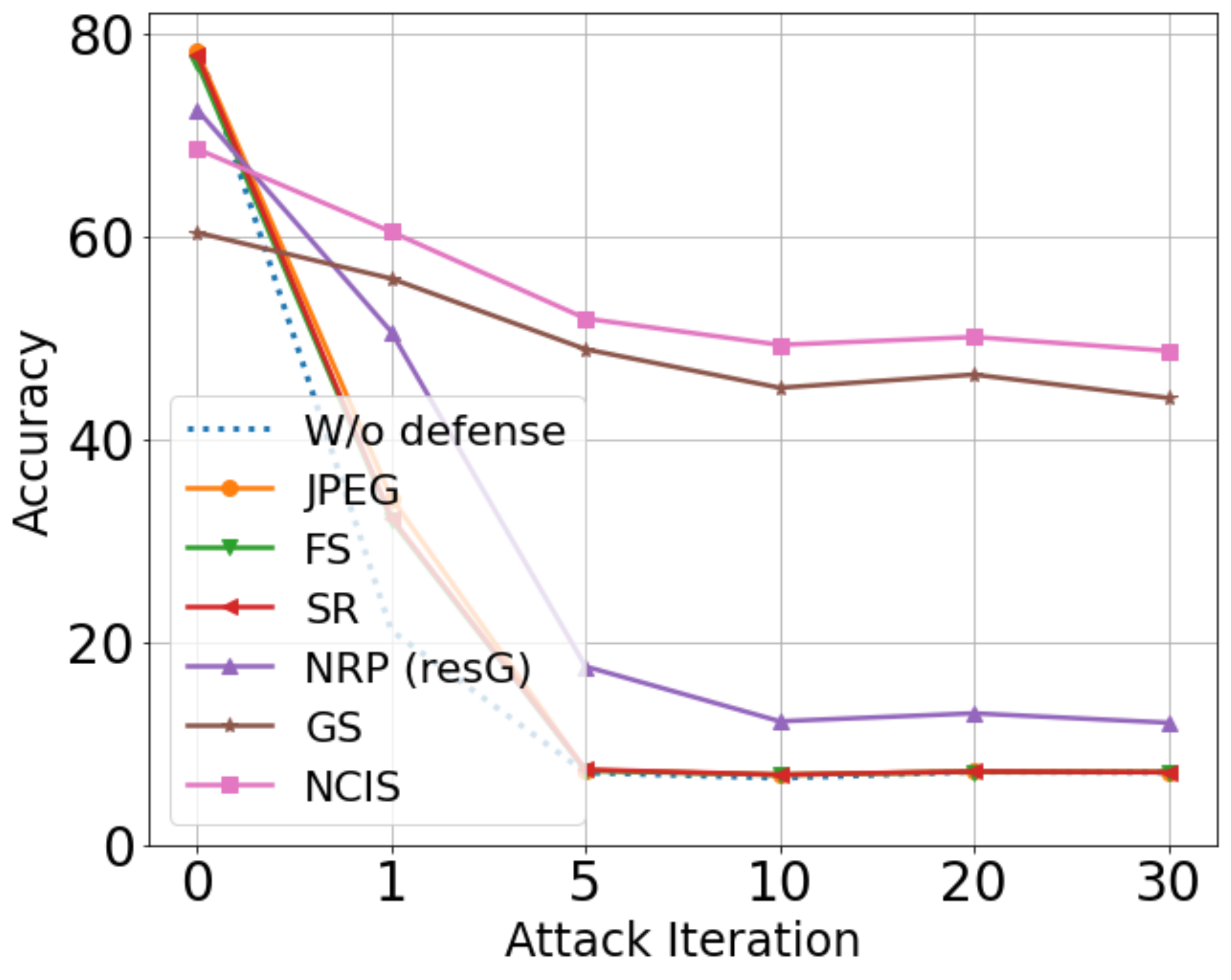}}
\subfigure[Various attack iters (ResNeXT-101~\cite{(resnext)xie2017aggregated})]
{\includegraphics[width=0.32\linewidth]{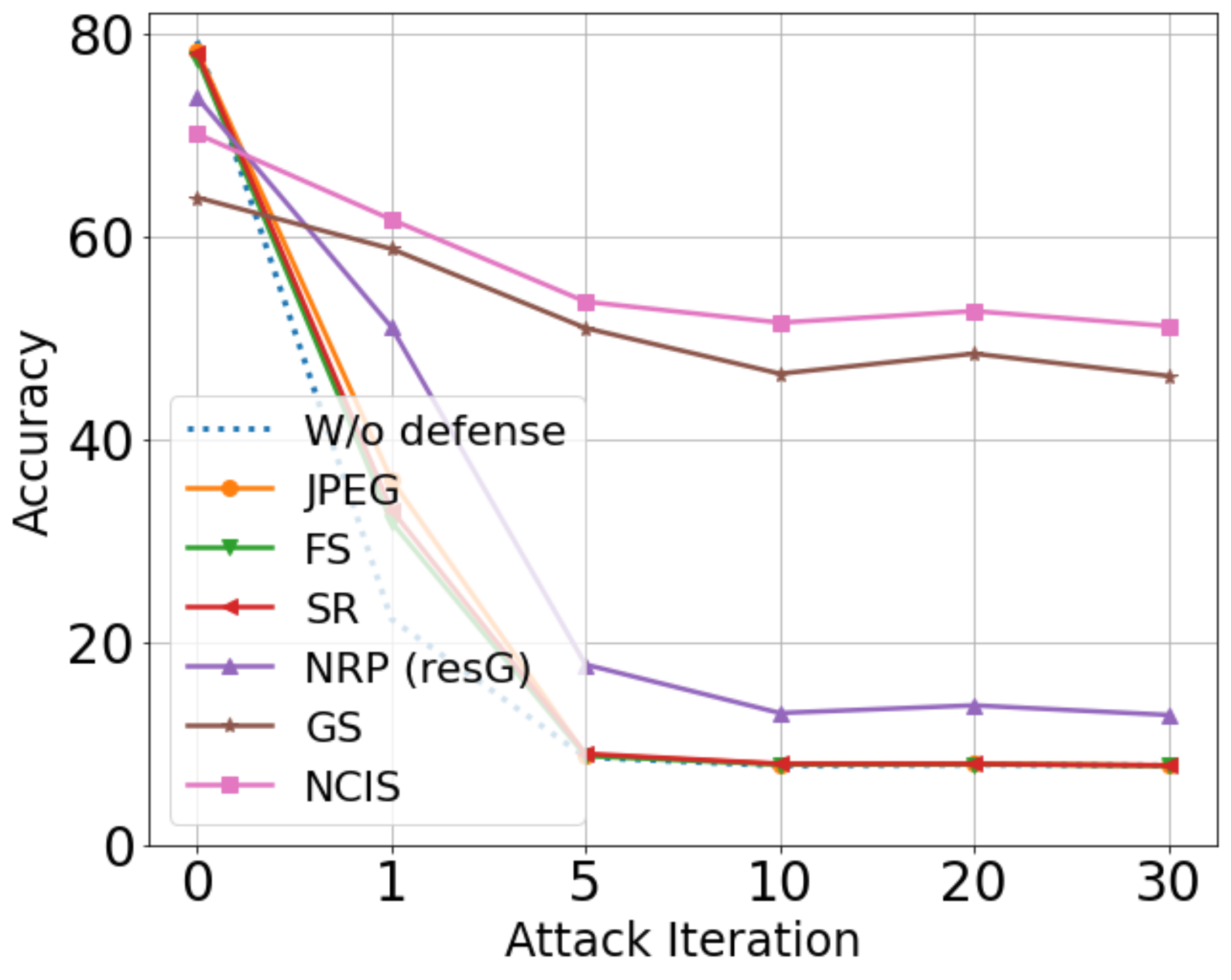}}
\subfigure[Various attack iters (RegNet-32G~\cite{(regnet)radosavovic2020designing})]
{\includegraphics[width=0.32\linewidth]{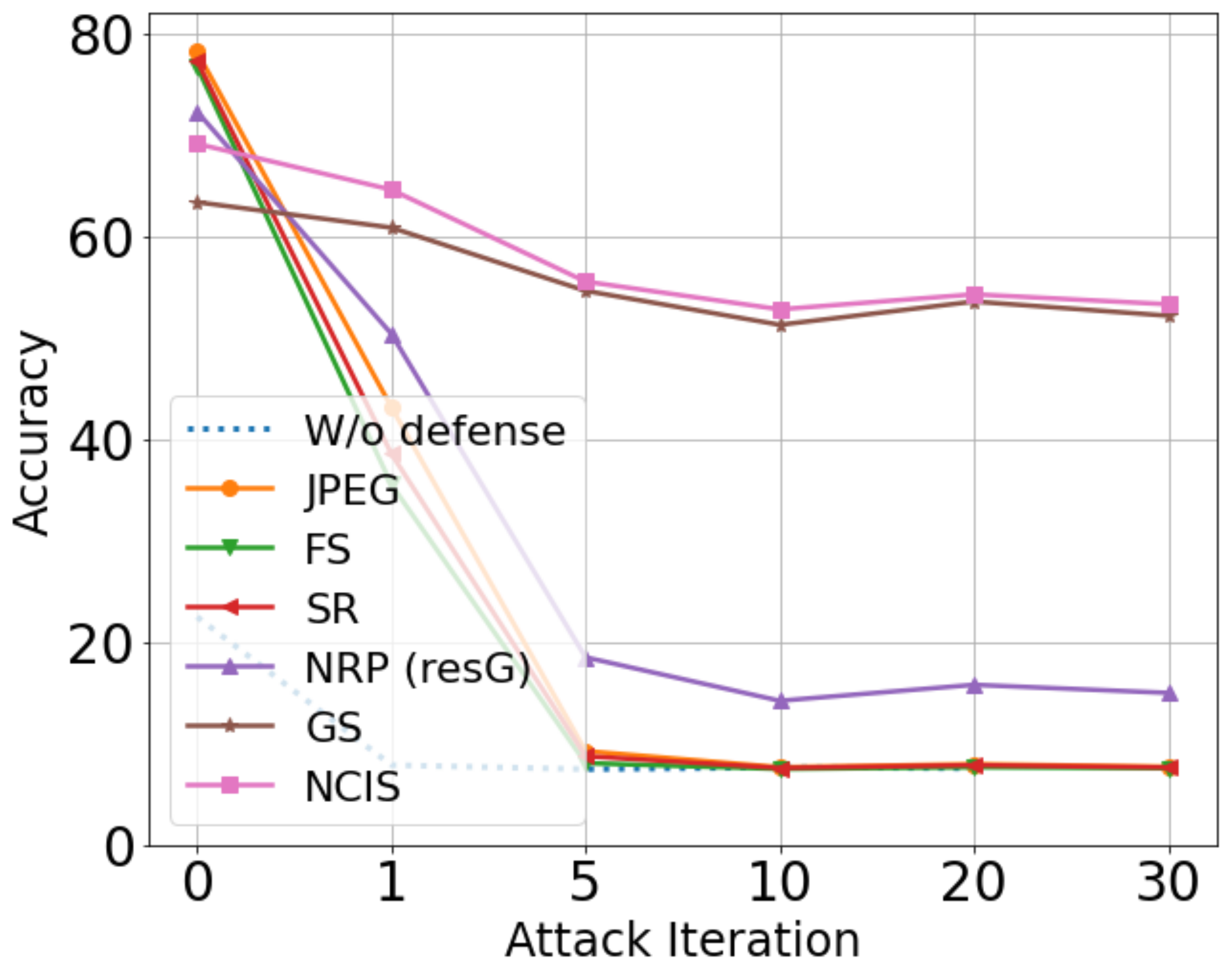}}
\caption{Experimental results against $L_\infty$ white-box PGD attacks on WideResNet-101/ResNeXT-101/RegNet-32G. For the experiments on various $\epsilon$, we set step size $\alpha = 1.6/255$ and attack iterations = 10.
For various attack iterations experiments, we equally set $\epsilon = 16/255$, and $\alpha = 1.6/255$ if the number of attack iteration is lower than 10, and set $\alpha = 1/255$ otherwise.}
\label{figure:whitebox_pgd_others}
 \end{figure*}

Figure \ref{figure:whitebox_pgd_others} shows additional experimental results against white-box attacks on other classification models. We clearly observe that our proposed methods surpass other baselines on all classification models.
Notably, we see that the performance gap between NCIS and GS is slightly wider than the gap on ResNet-152 which was shown in the manuscript.
We believe these results show our methods generally purify overall adversarial examples generated from various types of the classification model.

\newpage
\subsection{Transfer-based black-box attack on other classification models}

\begin{figure*}[h]
\centering 
\subfigure[Various $\epsilon$ (WideResNet-101~\cite{(WResNet)zagoruyko2016wide})]
{\includegraphics[width=0.32\linewidth]{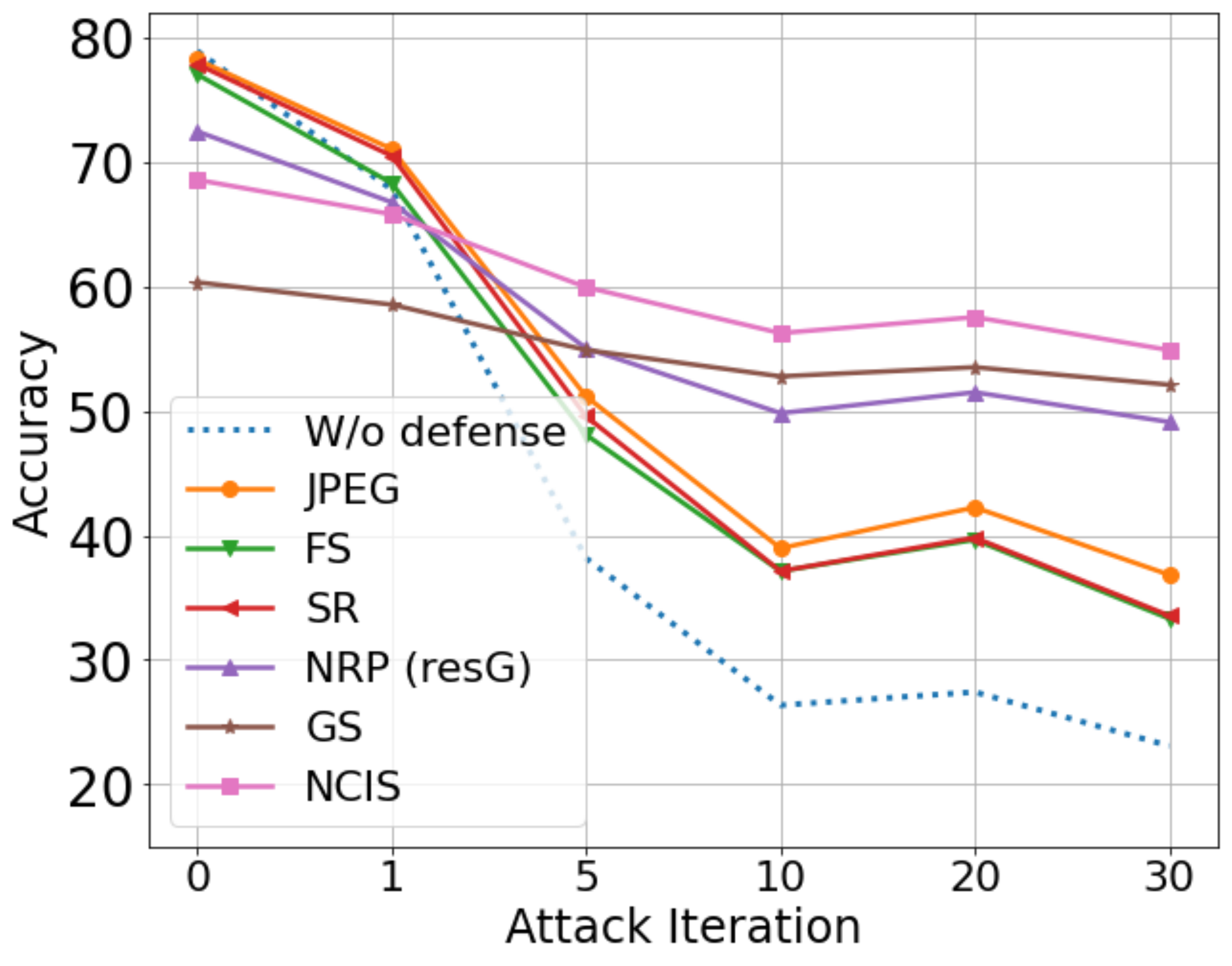}}
\subfigure[Various $\epsilon$ (ResNeXT-101~\cite{(resnext)xie2017aggregated})]
{\includegraphics[width=0.32\linewidth]{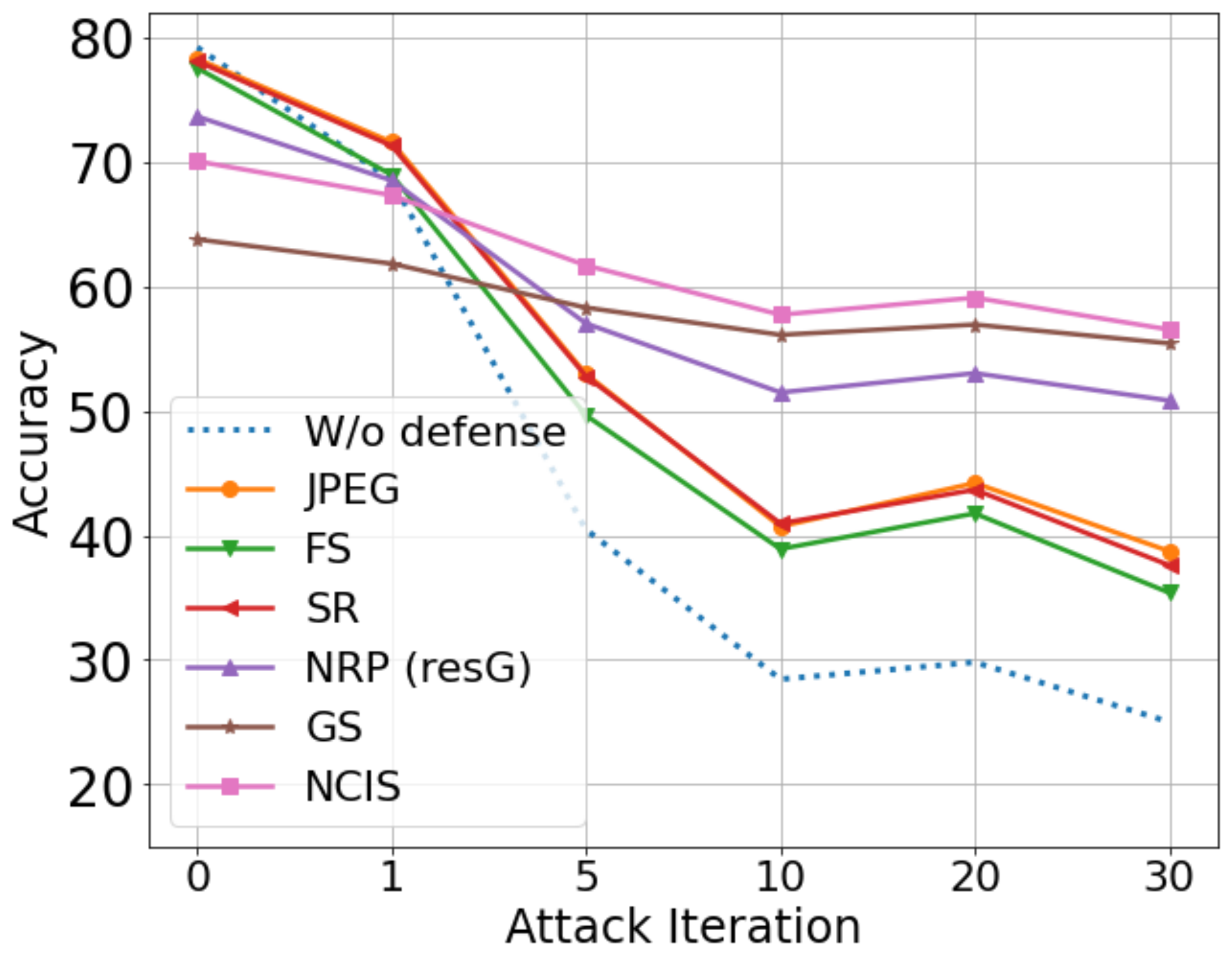}}
\subfigure[Various $\epsilon$ (RegNet-32G~\cite{(regnet)radosavovic2020designing})]
{\includegraphics[width=0.32\linewidth]{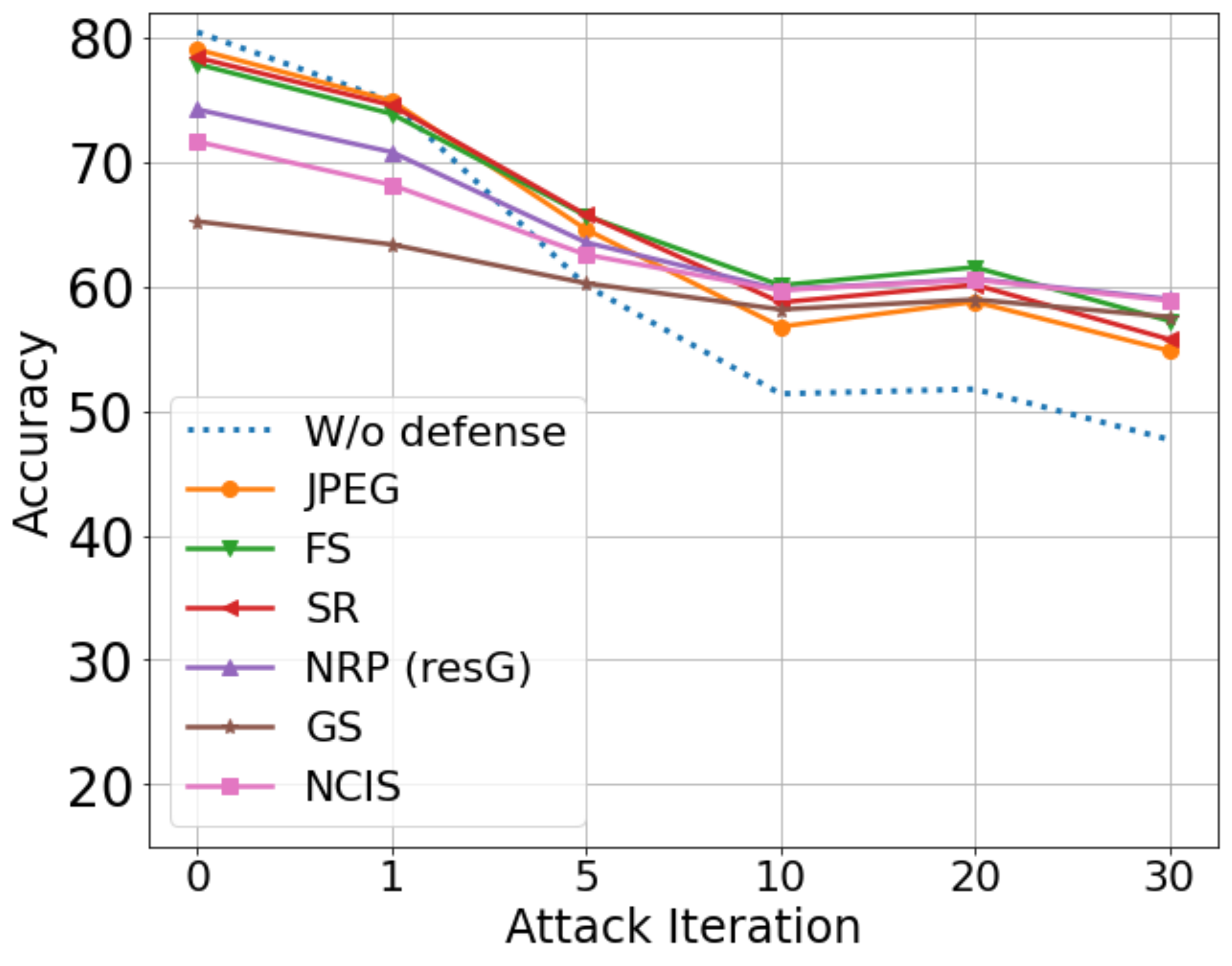}}
\subfigure[Various attack iters (WideResNet-101~\cite{(WResNet)zagoruyko2016wide})]
{\includegraphics[width=0.32\linewidth]{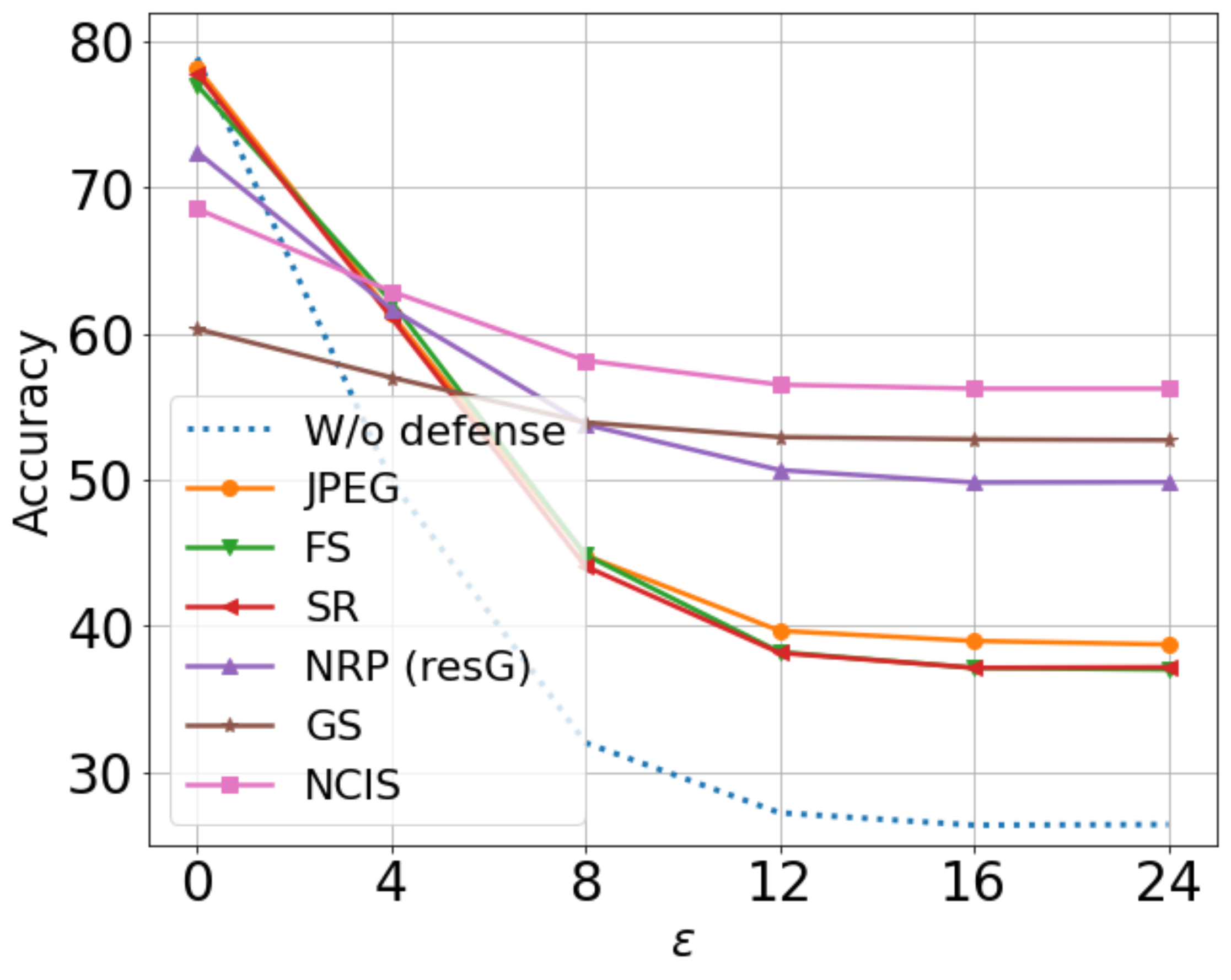}}
\subfigure[Various attack iters (ResNeXT-101~\cite{(resnext)xie2017aggregated})]
{\includegraphics[width=0.32\linewidth]{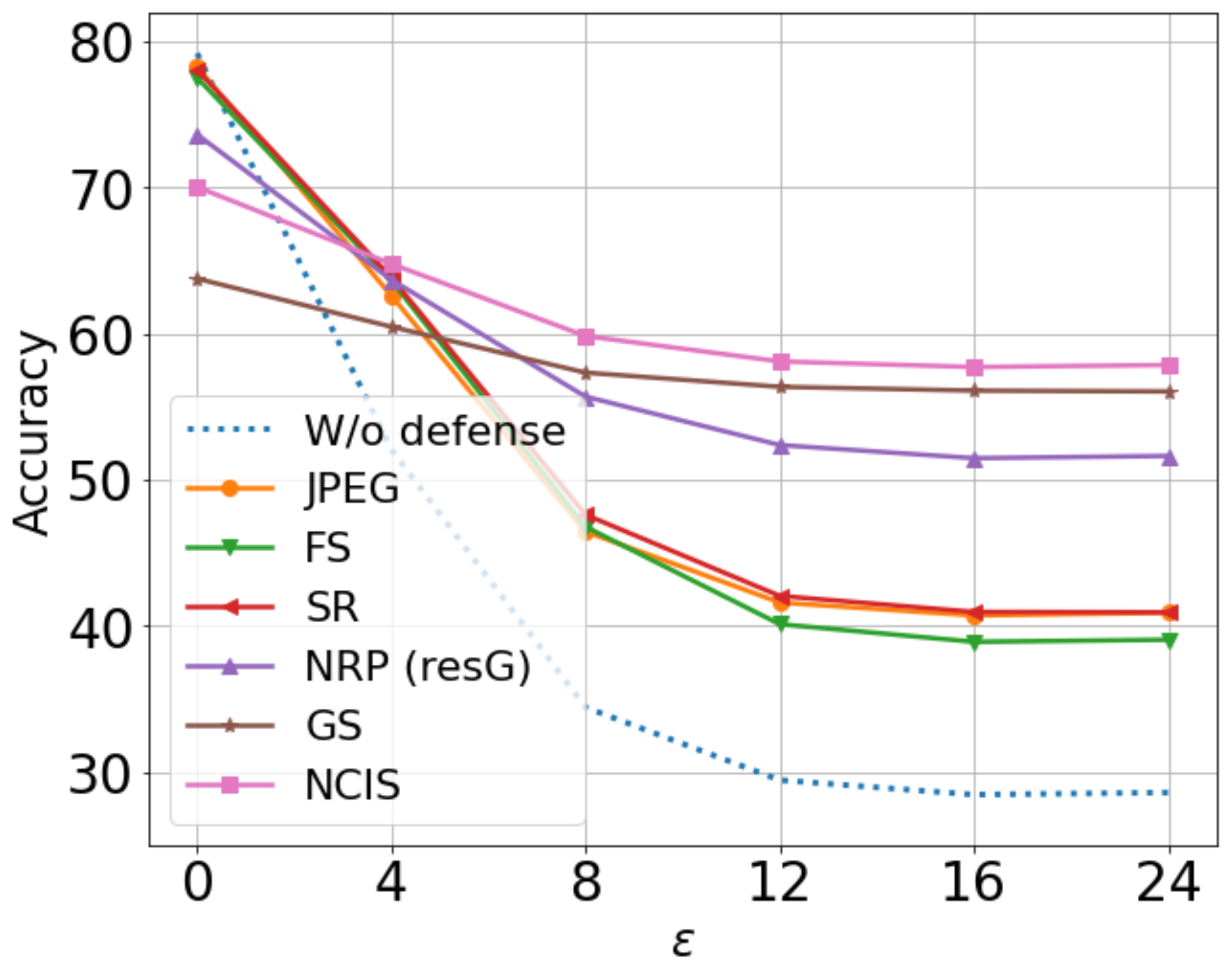}}
\subfigure[Various attack iters (RegNet-32G~\cite{(regnet)radosavovic2020designing})]
{\includegraphics[width=0.32\linewidth]{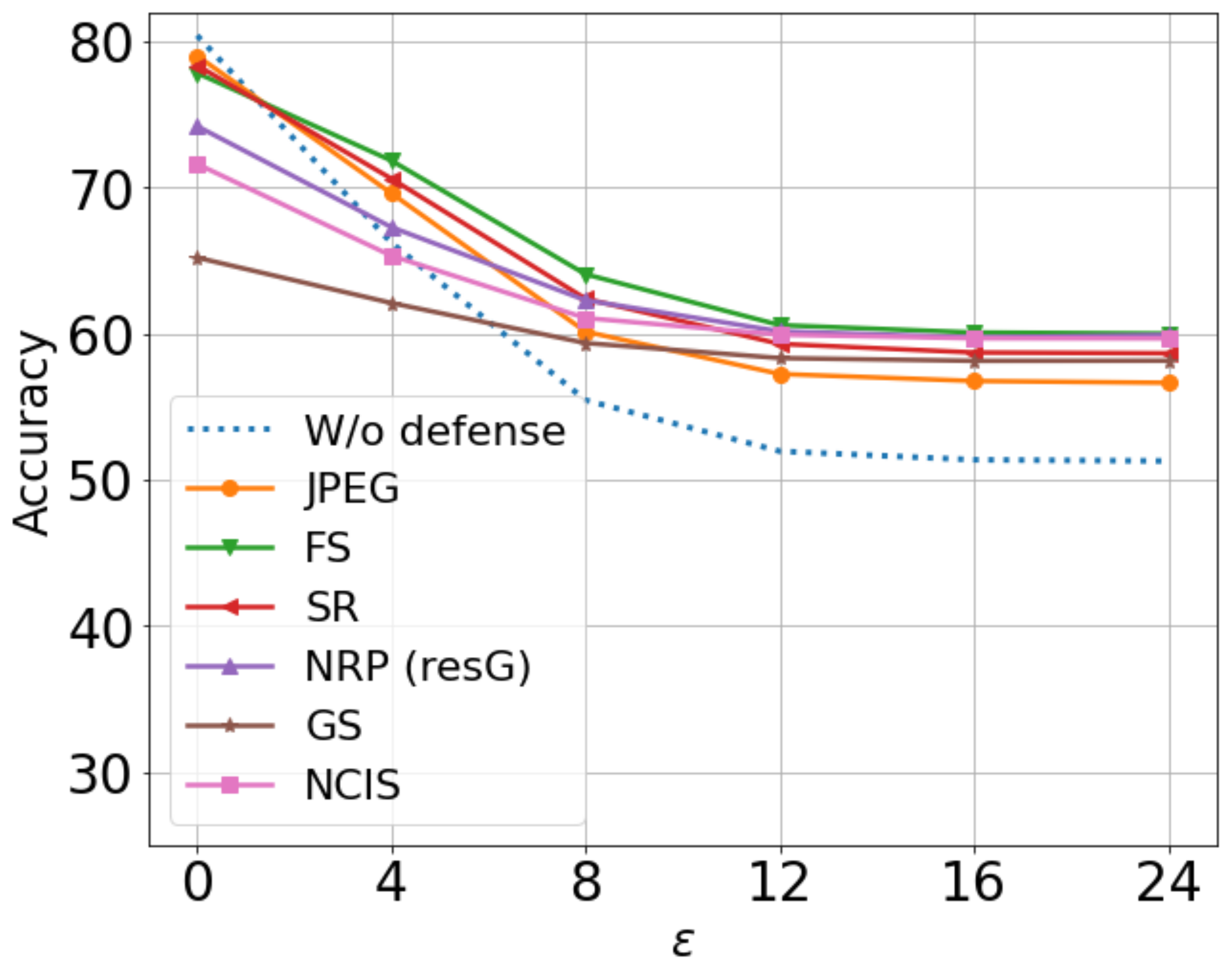}}
\caption{Experimental results of transfer-based black-box attack with $L_\infty$ PGD attack on WideResNet-101 (substitute model: ResNet-152)/ResNeXT-101 (substitute model: ResNet-152)/RegNet-32G (substitute model: WideResNet-101).
For the experiments on various $\epsilon$, we set step size $\alpha = 1.6/255$ and attack iterations = 10.
In the case of experiments on attack iterations, we equally set $\epsilon = 16/255$, and $\alpha = 1.6/255$ if the number of attack iteration is lower than 10, and set $\alpha = 1/255$ otherwise.}
\label{figure:transfer_blackbox_pgd_others}
 \end{figure*}

Following the manuscript, Figure \ref{figure:transfer_blackbox_pgd_others} shows the experimental results against transfer-based black-box attacks on other classification models. We set a substitute model for each classification model and generated adversarial examples by attacking it. The experimental results demonstrate that GS and NCIS achieve superior results than NRP (resG) in most cases. Also, the same result is shown once again that the performance gap between NCIS and GS is slightly wider than the gap on WideResNet-101.
Note that transfer-based black-box attacks on RegNet-32G were not as successful as attacks on other classification models so that there is not much difference in robust accuracy between defense methods.

\newpage
 
\subsection{Score-based black-box attack}

\begin{table}[h]
\caption{Experimental results for Square attack.}
\vspace{-.1in}
\centering
\smallskip\noindent
\resizebox{.4\linewidth}{!}{
\begin{tabular}{|cc||c|c|}
\hline
\multicolumn{2}{|c|}{Model / Defense}                                                                      & Square ($L_\infty$) & Square ($L_2$) \\ \hline \hline
\multicolumn{1}{|c|}{\multirow{8}{*}{{\rotatebox[origin=c]{90}{ResNet-152}}}} & \begin{tabular}[c]{@{}c@{}}W/o \\ defense\end{tabular} & 38.00               & 37.68          \\ \cline{2-4} 
\multicolumn{1}{|c|}{}                            & JPEG                                                   & 58.81               & 63.20          \\ \cline{2-4} 
\multicolumn{1}{|c|}{}                            & FS                                                     & 62.57               & 63.68          \\ \cline{2-4} 
\multicolumn{1}{|c|}{}                            & TVM                                                    & 51.23               & 58.05          \\ \cline{2-4} 
\multicolumn{1}{|c|}{}                            & SR                                                     & 58.99               & 60.04          \\ \cline{2-4} 
\multicolumn{1}{|c|}{}                            & NRP (resG)                                             & 53.10               & 59.25          \\ \cline{2-4} 
\multicolumn{1}{|c|}{}                            & \textbf{GS}                                            & \textbf{55.31}      & \textbf{51.57} \\ \cline{2-4} 
\multicolumn{1}{|c|}{}                            & \textbf{NCIS}                                          & \textbf{58.52}      & \textbf{55.95} \\ \hline
\end{tabular}
}
\label{table:square_blackbox}
\end{table}

Following the suggestion of \cite{(evaluating)carlini2019evaluating}, we evaluate both baselines and our proposed method against score-based black-box attack.
We select Square~\cite{(square)andriushchenko2020square}, which is the state-of-the-art and query efficient black-box attack method, for the experiment.
For $L_2$ Square attack, we set $\epsilon = 5, p = 0.1$ with 300 queries and, for $L_\infty$ Square attack, we set $\epsilon = 8/255, p = 0.1$ with 300 queries. Other hyperparameters are set to default values as \cite{(advertorch)kim2020torchattacks}.

Figure \ref{table:square_blackbox} shows the experimental results against $L_\infty$ and $L_2$ Square attacks.
First, we observe that attack success rate of Square is still low compared to transfer-based black-box attacks and it requires more than 300 queries to successfully fool the classification model.
Second, as already shown in \cite{(benchmarking)dong2020benchmarking}, the traditional input transformation-based methods (\textit{e.g.} JPEG, FS, TVM and SR) show robust performance compared to others, such as white-box and transfer-based black-box attacks.
Third, our proposed methods (NCIS and GS) and NRP (resG) accomplish competitive performance, and NCIS consistently surpasses GS.
When comparing NCIS with NRP (resG), we observe that NCIS obtain higher robust accuracy than NRP (resG) in Square ($L_\infty$) but lower robust accuracy in Square ($L_2$).

\begin{figure*}[h]
\centering 
\includegraphics[width=0.7\linewidth]{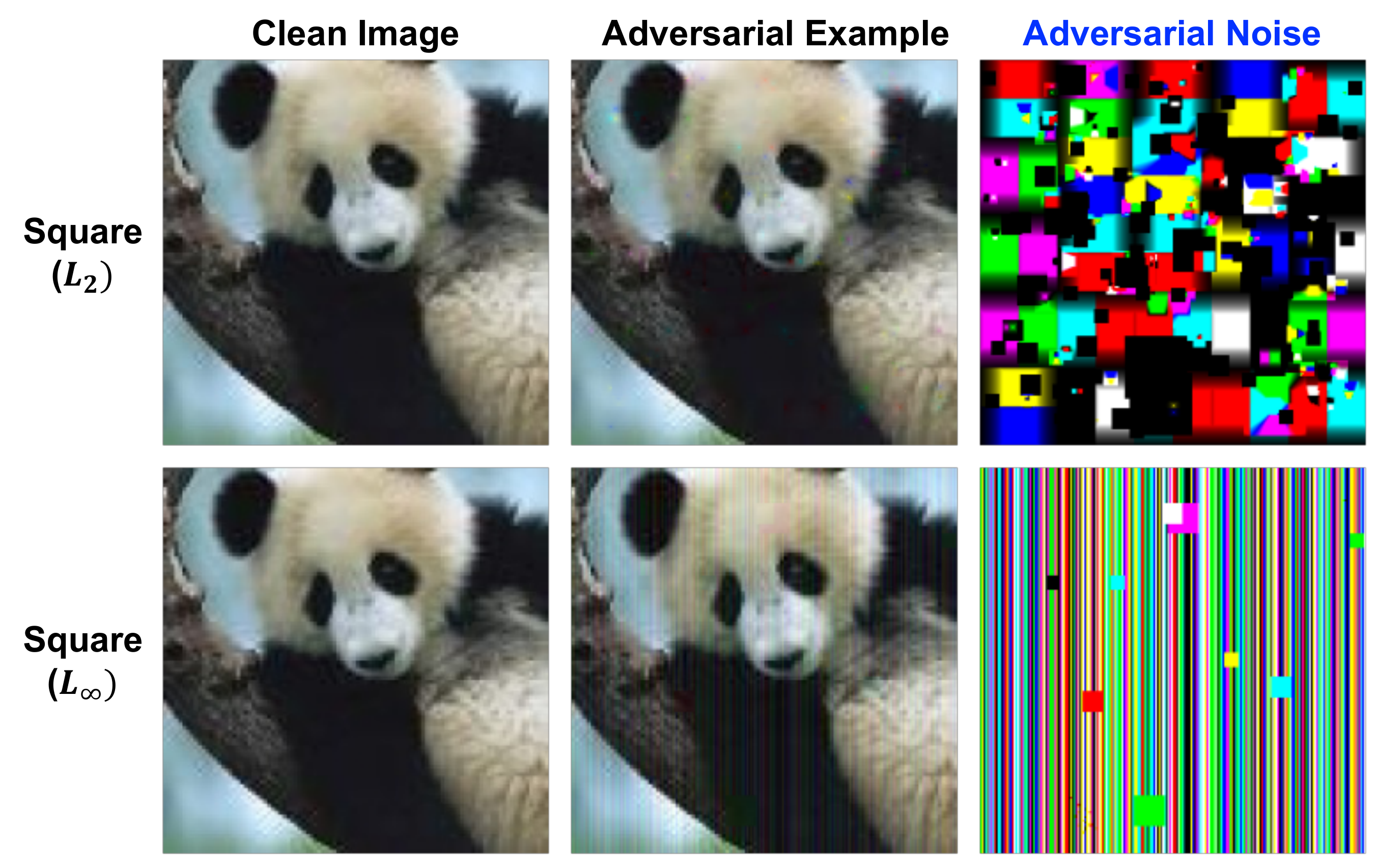}
\caption{Visualizations of adversarial example and adversarial noise generated by Square \cite{(square)andriushchenko2020square} attack.}
\label{figure:analysis_square}
 \end{figure*}

As an analysis, we visualized adversarial examples and noises generated by Square attack in Figure \ref{figure:analysis_square}. 
We numerically checked that the mean of the adversarial noise of an entire image is almost zero.
However, the figure clearly shows that adversarial noise generated by Square has a structured noise different from adversarial noise generated by the optimization-based white-box attack (\textit{e.g} PGD).
We believe that this is the reason why both our proposed method and NRP (resG) show lower performance compared to input transformation-based methods, this type of adversarial noise is not a form considered in both methods.

\newpage

\subsection{Experimental results with dynamic inference against the purifier-aware white-box attack}

\begin{table}[h]
\caption{Experimental results with dynamic inference.}
\vspace{-.1in}
\centering
\smallskip\noindent
\resizebox{.7\linewidth}{!}{
\begin{tabular}{|c||c|c|c|}
\hline
ResNet-152             & Standard Accuracy & White-box PGD & BPDA \cite{(obfuscated)athalye2018obfuscated} White-box PGD \\ \hline \hline
NRP (resG)   & 66.56 & 37.30         & 33.52              \\ \hline
GS           & 63.06 & 44.46          & 7.90               \\ \hline
NCIS ($i = 4$) & 66.58 & 45.64         & 24.56              \\ \hline
NCIS ($i = 5$) & 61.20 & 48.42         & 33.20              \\ \hline
\end{tabular}}
\label{table:bpda_white_box}
\end{table}

Even though our paper focuses on the purifier-blind white-box attack, we conducted experiments on the purifier-aware white-box attack (in other words, strong adaptive attack~\cite{(obfuscated)athalye2018obfuscated, (adaptive_attack)tramer2020adaptive}) by following the suggestion of \cite{(adaptive_attack)tramer2020adaptive}.
We used BPDA~\cite{(obfuscated)athalye2018obfuscated} implemented in \cite{(advertorch)ding2019advertorch} for the purifier-aware white-box attack and applied it to $L_\infty$ PGD attack.
We used randomly sampled 10\% data of ImageNet validation dataset and used untargeted $L_\infty$ PGD attack ($\epsilon = 16, \alpha = 1.6/255$) with 10 attack iterations for all experiments.
We applied dynamic inference against BPDA white-box attack as proposed in the Supplementary Material of NRP~\cite{(NRP)naseer2020self} to our methods and implemented it by referring their code.
Same as default hyperparameters, we set $\epsilon$ for dynamic inference to $16/255$ and add Gaussian noise with zero mean and $\sigma = 0.05$ to a given input image at each inference.

Table \ref{table:bpda_white_box} shows the experimental results.
We experimentally found that when dynamic inference is applied, the optimal number of iterations ($i = 4$ or $i = 5$) for NCIS changes from the original number ($i=7$ for ResNet-152) because of the added noises to the given input image.
From the table, we observe that NRP (resG) and NCIS ($i = 4$) achieves superior standard accuracy.
Besides, we clearly observe once again that for the purifier-blind white-box PGD attack (the results of White-box PGD in the table), NCIS and GS outperform NRP (resG). Moreover, in BPDA white-box PGD attack, GS shows a significant drop in performance but NCIS ($i = 5$) shows a competitive robust accuracy compared to NRP (resG).

In conclusion, though we mainly consider the purifier-blind white-box attack, our NCIS can be robust to not only the purifier-blind attack but also the purifier-aware attack using BPDA, by easily applying dynamic inference proposed in the NRP paper.


\newpage
\section{Additional Details on API Experiments}
\subsection{Dataset generation}

For generating benchmark datasets, we sampled images from ImageNet training dataset and generated adversarial examples of it using  transfer-based black-box attack using ensemble of five classification models, ResNet-152, VGG-19, GoogleNet, ResNeXT-101, WideResNet-101, based on \cite{(ensemble_attack)liu2016delving}.
We attacked each image with targeted $L_{\infty}$ PGD $(\epsilon=16/255, \alpha=1.6/255)$ attack with 10 attack iterations, and then made a pair of a clean and an adversarial example.
Then, we queried each pair and stored them only when all the top five predicted labels of the clean and adversarial example were totally different.
The above process was performed on all the four APIs, and we respectively sampled 100 pairs for test dataset and 20 pairs for validation dataset.
In other words, we sampled 100 pairs of test dataset and 20 pairs of validation dataset for each API respectively.
We will open all generated datasets publicly.

\subsection{Evaluation metrics}

First, \textbf{Prediction Accuracy} is the measure for the same number of labels among top five labels between the predicted label of the purified image and the predicted label of the clean image.
Second, \textbf{Top-1 Accuracy} is the measure for whether the Top-1 label of the purified image is same as the Top-1 label of the clean image.
Finally, \textbf{Top-5 Accuracy} is the measure for whether the Top-1 label of the clean image exists within the Top-5 predicted labels of the purified image.

\subsection{Hyperparameter selection}

\begin{figure*}[h]
\centering 
\subfigure[Experimental results of GS]
{\includegraphics[width=0.32\linewidth]{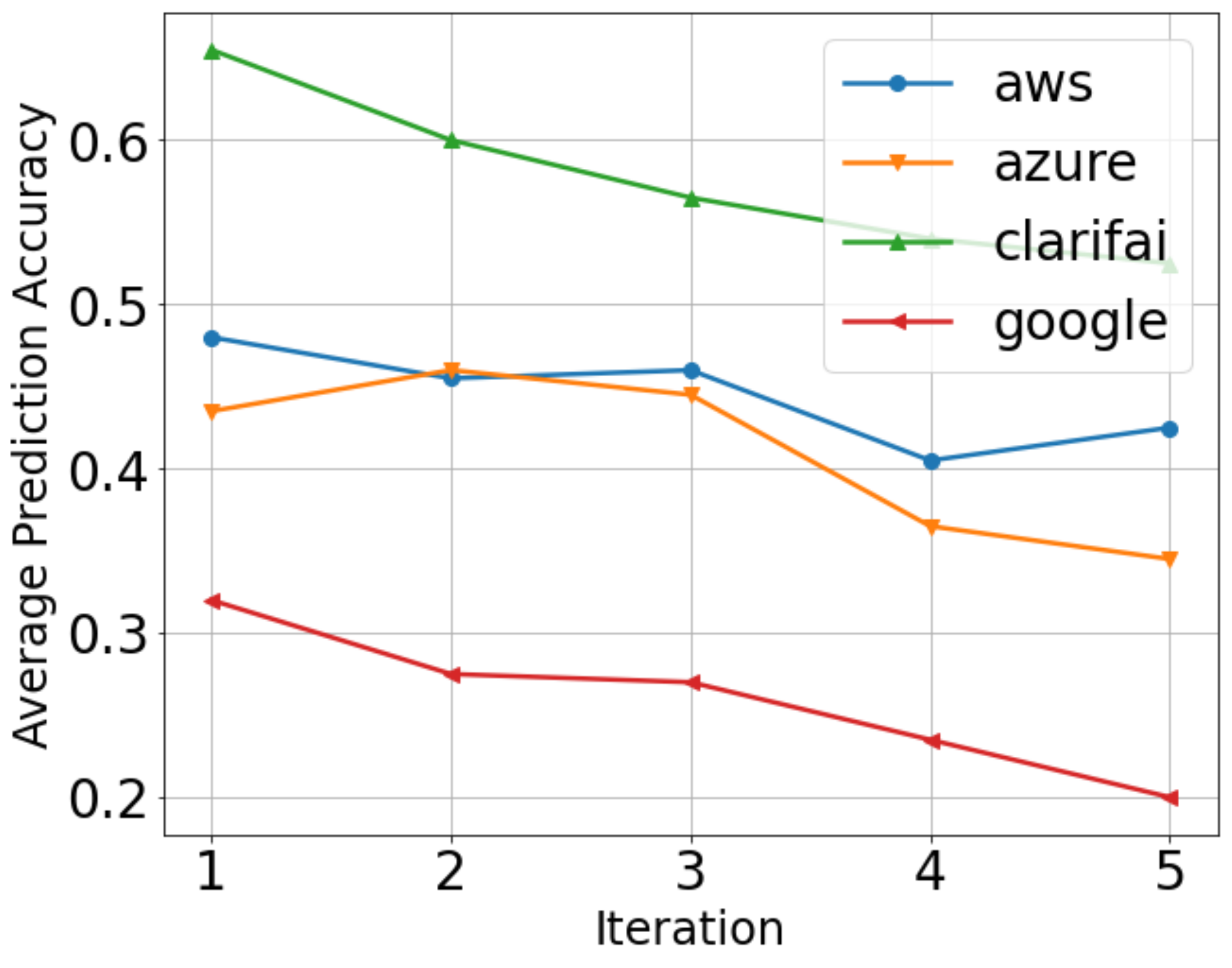}}
\subfigure[Experimental results of NCIS]
{\includegraphics[width=0.32\linewidth]{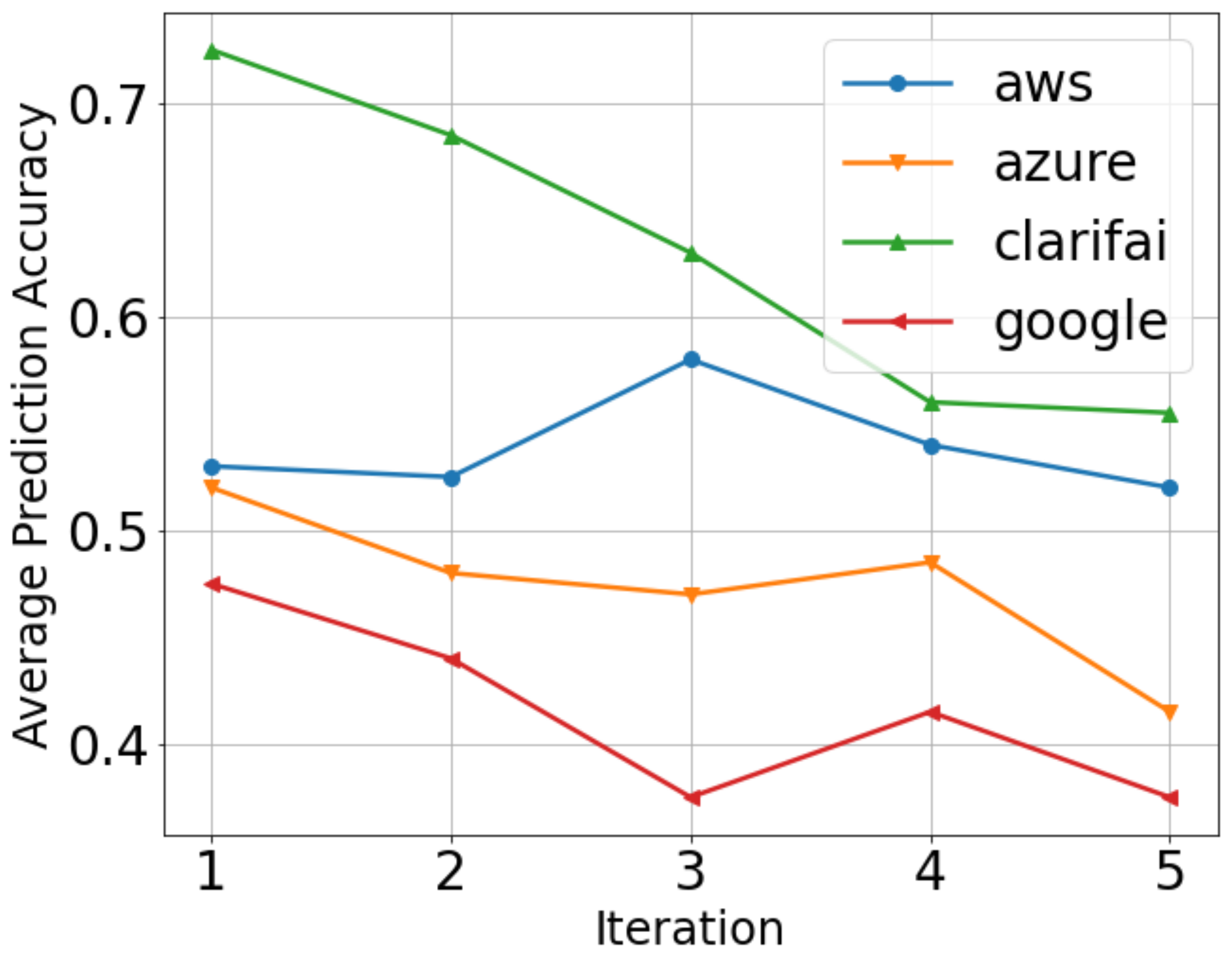}}
\caption{Experiments for hyperparameter selection of GS and NCIS.}
\label{figure:api_hyperparmeter}
 \end{figure*}

\begin{table}[h]
\caption{Experimental results of hyperparameter selection.}
\vspace{-.1in}
\centering
\smallskip\noindent
\resizebox{.5\linewidth}{!}{
\begin{tabular}{|c||c|c|c|c|}
\hline
\begin{tabular}[c]{@{}c@{}}Prediction\\ Accuracy\end{tabular} & AWS          & Azure        & Clarifai     & Google       \\ \hline \hline
GS                                                            & 0.48 ($i = 1$) & 0.46 ($i = 2$) & 0.65 ($i = 1$) & 0.32 ($i = 1$) \\ \hline
NCIS                                                          & 0.58 ($i = 3$) & 0.51 ($i = 1$) & 0.73 ($i = 1$) & 0.48 ($i = 1$) \\ \hline
\end{tabular}
}
\label{table:selection_i_api}
\end{table}

For GS and NCIS, we found the number of iterations $i$ for each API by using the generated validation dataset. 
As a criterion, we only consider highest average Prediction Accuracy of a clean and adversarial example to select best $i$.
Figure \ref{figure:api_hyperparmeter} shows the experimental results and Table \ref{table:selection_i_api} selected best $i$ for each dataset.
Note that all selected $i$ are used for the experiments for APIs in the manuscript.

\newpage

\subsection{Additional qualitative results}

\begin{figure*}[h]
\centering 
\subfigure[Amazon AWS]
{\includegraphics[width=0.98\linewidth]{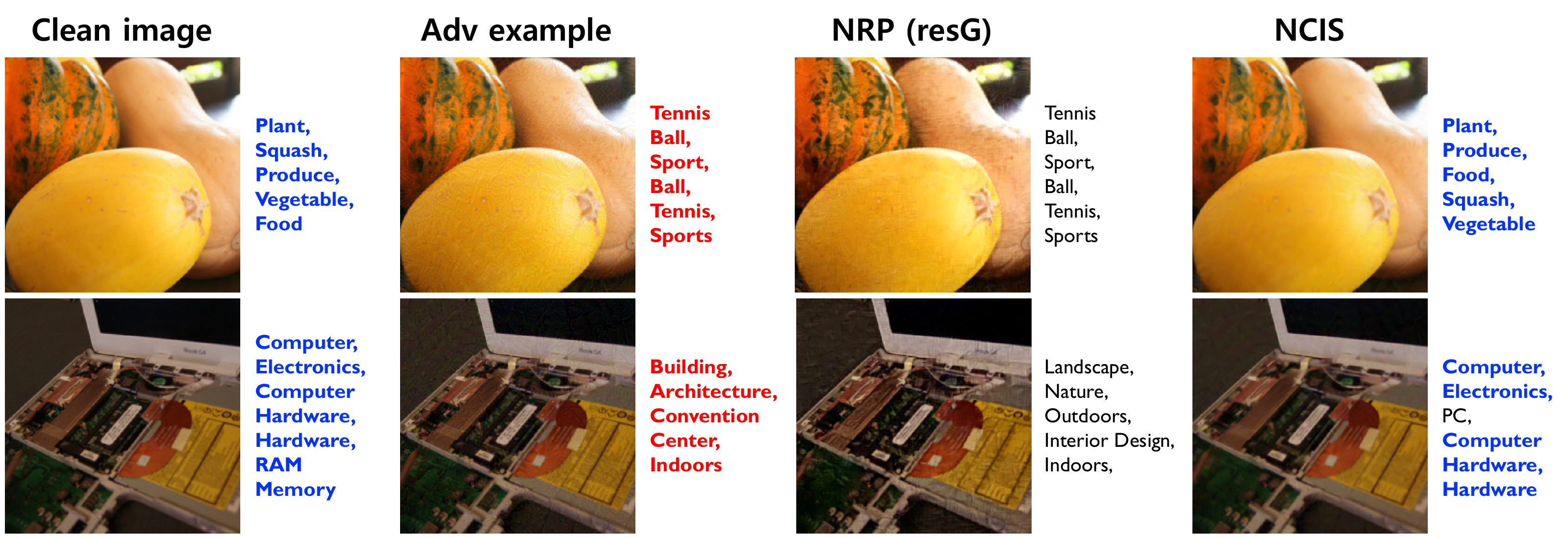}}
\subfigure[Microsoft Azure]
{\includegraphics[width=0.98\linewidth]{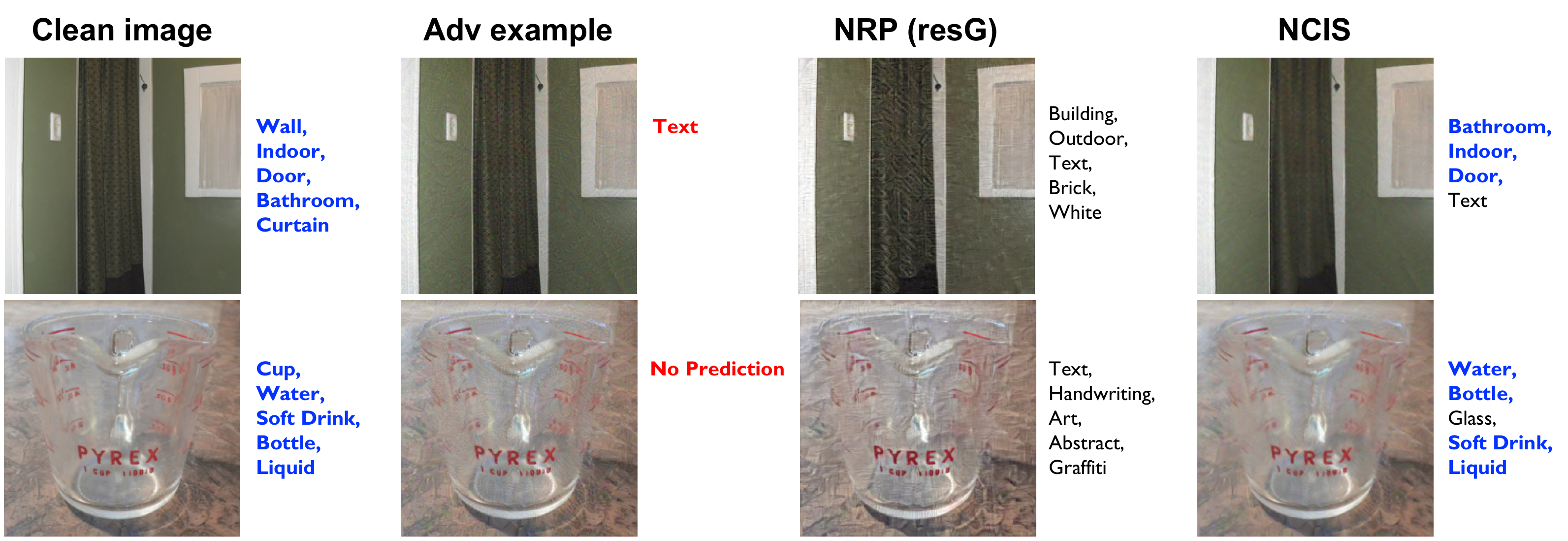}}
\subfigure[Google]
{\includegraphics[width=0.98\linewidth]{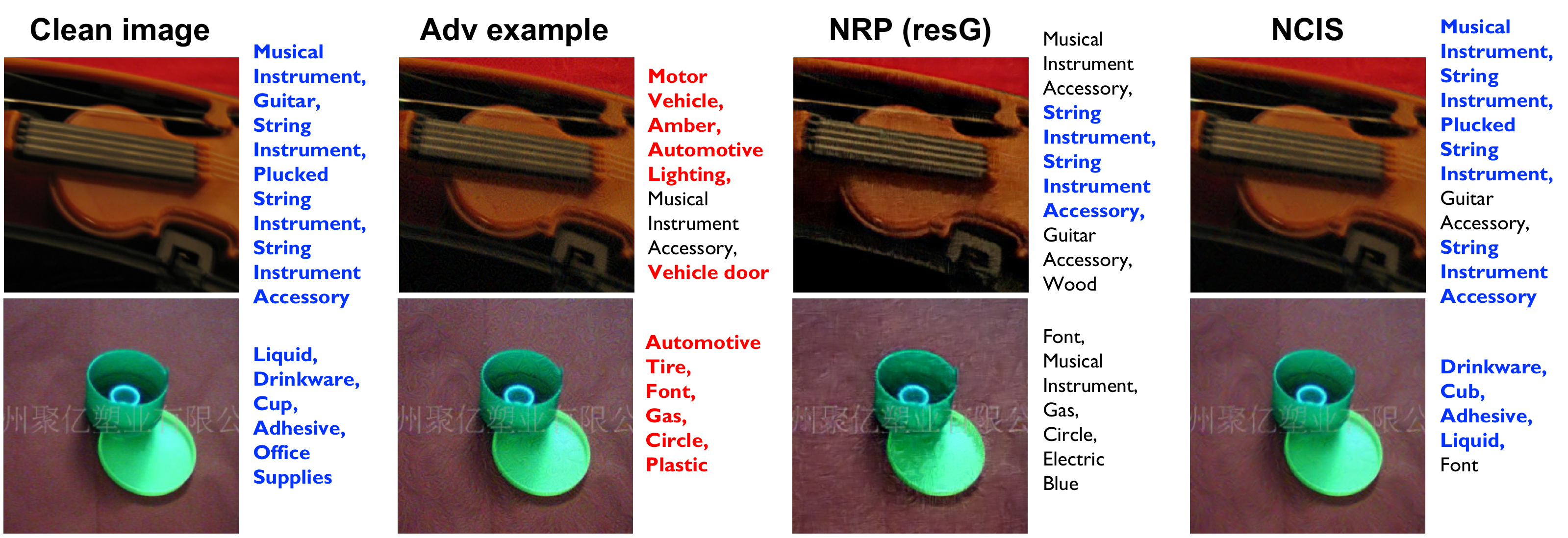}}
\vspace{-2mm}
\caption{Visualization examples of defending commercial vision APIs. The APIs predict correct top-5 predictions of the original clean images (first column), and when completely fooled by the adversarial examples (second column). The right two columns show the prediction results when two purifiers, NRP (resG) \cite{(NRP)naseer2020self} and our NCIS, are applied to the adversarial examples. }\vspace{-.2in}
 \end{figure*}
 
\newpage

\section{Limitations of our work}
Even though we proposed a novel and efficient method for adversarial purification, we think that our work also has some limitations as follows.
First, our NCIS is effective for purifying adversarial examples generated by optimization-based attacks, which has an almost zero mean and symmetric distribution in a patch, but shows the limitation to purify adversarial examples generated by score-based black-box attacks, which have structured adversarial noises.
Second, despite many improvements in inference time and computational cost, NCIS is still slow and requires expensive cost compared to traditional input transformation-based methods.
We believe that future studies to overcome these limitations will provide a new direction for achieving adversarial robustness of classification models in a more practical way.

{\small
\bibliographystyle{ieee_fullname}
\bibliography{egbib}
}